\documentclass[10pt,journal,compsoc]{IEEEtran}
\usepackage[T1]{fontenc}
\ifCLASSOPTIONcompsoc

\usepackage{cite}

\usepackage{bm}
\usepackage{amsmath}

\usepackage[table]{xcolor}
\definecolor{verylightgray}{rgb}{0.85, 0.85, 0.85}

\usepackage{amsthm}
\usepackage{float}
\usepackage{setspace}
\usepackage{amssymb}
\usepackage{stfloats}
\usepackage{cite}
\usepackage{ragged2e}
\usepackage{colortbl}
\usepackage[ruled,vlined,linesnumbered]{algorithm2e}
\usepackage{algorithmic}
\usepackage{amsfonts}
\usepackage{mathrsfs}
\usepackage{amsmath,amsthm}
\usepackage{array,booktabs}
\usepackage{subfigure}
\usepackage{multirow}
\usepackage{cuted}
\usepackage{multicol}
\usepackage{graphicx}
\usepackage{subfigure}
\usepackage{graphicx,xcolor,bm}
\usepackage{hyperref}
\usepackage{threeparttable}
\usepackage{dcolumn}
\usepackage{setspace}
\usepackage{makecell}
\usepackage{lipsum}
\usepackage{enumerate}
\usepackage{hhline}

\usepackage{tabularx}

\usepackage{cite,bm}
\graphicspath{{figures/}}
\usepackage{bm}

\usepackage[protrusion=true,expansion=true]{microtype}
\usepackage{caption}
\usepackage[font=footnotesize,labelfont=bf]{caption}

\makeatletter
\renewcommand\paragraph{\@startsection{paragraph}{4}{0em}
	{1ex \@plus1ex \@minus.2ex}
	{0.3em}
	{\normalfont\normalsize\itshape}}
\makeatother
\AfterEndEnvironment{paragraph}{\noindent\ignorespaces}
\begin{document}
	\title{Towards Seamless Hierarchical Federated Learning under Intermittent Client Participation: A Stagewise Decision-Making Methodology}
	\author{\vspace{-1.5mm}Minghong Wu, Minghui Liwang, \IEEEmembership{Member}, \IEEEmembership{IEEE}, Yuhan Su, Li Li, \IEEEmembership{Member}, \IEEEmembership{IEEE}, Seyyedali~Hosseinalipour, \IEEEmembership{Member}, \IEEEmembership{IEEE}, Xianbin Wang, \IEEEmembership{Fellow}, \IEEEmembership{IEEE}, Huaiyu Dai, \IEEEmembership{Fellow}, \IEEEmembership{IEEE}, Zhenzhen Jiao
		
		\thanks{This work was supported in part by the National Natural Science Foundation of China under Grant nos. 62271424, 62401486, 62088101, 72171172; Shanghai Pujiang Programme under Grant no. 24PJD117; the Shanghai Municipal Science and Technology Major Project under Grant no. 2021SHZDZX0100; the Aeronautical Science Foundation of China under Grant no. 2023Z066038001; the Chinese Academy of Engineering, Strategic Research and Consulting Program under Grant no. 2023-XZ-65.  
        M. Wu (minghongwu@stu.xmu.edu.cn) is with the School of Informatics, Xiamen University, Fujian, China. M. Liwang (minghuiliwang@tongji.edu.cn) and L. Li (lili@tongji.edu.cn) are with the Department of Control Science and Engineering, and also with the Shanghai Research Institute for Intelligent Autonomous Systems, Tongji University,	Shanghai, China. Yuhan Su (ysu@xmu.edu.cn) is with the School of Electronic Science and Engineering, Xiamen University, Xiamen 361005, China.  S. Hosseinalipour (alipour@buffalo.edu) is with Department of Electrical Engineering, University at Buffalo--SUNY, NY, USA. X. Wang (xianbin.wang@uwo.ca) is with the Department of Electrical and Computer Engineering, Western University, Ontario, Canada. H. Dai (hdai@ncsu.edu) is with the Department of Electrical and Computer Engineering, North Carolina State University, Raleigh, USA. Z. Jiao (jiaozhenzhen@teleinfo.cn) is with the iF-Labs, Beijing Teleinfo Technology Co., Ltd., CAICT, China. Corresponding author: Minghui Liwang (minghuiliwang@tongji.edu.cn)
			
	}
}

	\IEEEtitleabstractindextext{
		\begin{abstract}
			\justifying
			Federated Learning (FL) offers a pioneering distributed learning paradigm that enables devices/clients to build a shared global model that can be obtained through frequent model transmissions between clients and a central server, causing high latency, energy consumption, and congestion over backhaul links. To overcome these drawbacks, Hierarchical Federated Learning (HFL) has emerged, which organizes clients into multiple clusters and utilizes edge nodes (e.g., edge servers) for intermediate model aggregations between clients and the central server. Current research on HFL mainly focus on enhancing model accuracy, latency, and energy consumption in scenarios with a stable/fixed set of clients. However, addressing the dynamic availability of clients -- a critical aspect of real-world scenarios -- remains underexplored. This study delves into optimizing client selection and client-to-edge associations in HFL under intermittent client participation so as to minimize overall system costs (i.e., delay and energy), while achieving fast model convergence. We unveil that achieving this goal involves solving a complex NP-hard problem. To tackle this, we propose a stagewise methodology that splits the solution into two stages, referred to as Plan A and Plan B. Plan A focuses on identifying long-term clients with high chance of participation in subsequent model training rounds. Plan B serves as a backup, selecting alternative clients when long-term clients are unavailable during model training rounds. This stagewise methodology offers a fresh perspective on client selection that can enhance both HFL and conventional FL via enabling low-overhead decision-making processes. Through evaluations on diverse datasets, we show that our methodology outperforms existing benchmarks on crucial factors such as model accuracy and system costs.
		\end{abstract}
		
	}
	
\maketitle
\IEEEdisplaynontitleabstractindextext
%
\IEEEpeerreviewmaketitle

\section{Introduction }
\label{sec:introduction}

\IEEEPARstart{T}{he} proliferation of smart devices has led to the generation of massive amounts of data, fueling the rise of artificial intelligence (AI)/machine learning (ML) applications across Internet-of-Things (IoT) ecosystem \cite{ref1}. Conventional AI/ML methods typically involve pooling the data of distributed IoT devices at a central location (e.g., a server) for model training. However, privacy regulations often restrict the transfer of devices' data across the network. This has triggered a paradigm shift from centralized to distributed AI/ML methods, with federated learning (FL) \cite{ref3, ref4, ref43} emerging as a common approach. FL starts with a central server, e.g., a cloud server (CS), broadcasting a global model to clients (e.g., mobile devices). Subsequently, each client first synchronizes its local model with the global model and then trains its local model on its data (e.g., via gradient descent). Each client then uploads its local model parameters back to the server. The server finally aggregates the received models to update the global model (e.g., via weighted averaging). This process of local training and global model aggregations continues until reaching the desired performance \cite{ref2}.

FL naturally relies on frequent model transmissions between clients and the CS, which can incur high overhead (e.g., latency  and energy) \cite{ref5, ref6,ref7}. Hierarchical FL (HFL)  is a well-recognized framework to address this challenge \cite{ref8,ref9}. HFL introduces an intermediate layer to the model training architecture of FL for intermediate model aggregations (a layer of edge servers (ESs)), that synchronize the model parameters of their associated clients frequently, and exchange model updates with the CS with a lower frequency \cite{ref10}.

\subsection{Motivation and Research Direction}\label{subsec:motivation}
Although HFL has tremendous potentials in enhancing the time and energy efficiency of FL, its implementation in large-scale, dynamic networks (e.g., with mobile devices acting as clients and multiple ESs) faces unexplored challenges. We identify and study three major research questions: {\emph{{\romannumeral1})}} \emph{How to actively select/recruit the dynamic and resource-constrained clients to achieve cost-efficiency (e.g., reducing the training delay and energy consumption)?} Due to limited bandwidth availability, only a subset of clients can be selected/recruited for interactions with ESs \cite{ref12}. The intermittent availability and heterogeneity of clients (i.e., varying computational and communication capabilities), as well as differences in their data qualities can further complicate the client selection \cite{ref11}. {\emph{{\romannumeral2})}} \emph{How to associate the selected clients to proper ESs to achieve resource-efficiency and acceleration of model convergence?} Given a set of selected clients, different client-to-edge (C2E) association strategies can significantly influence the model convergence, due to non-independent and identically distributed (non-IID) nature of data across clients \cite{ref13}. We hypothesize that optimal convergence occurs when the C2E association results in clusters of devices at each ES having a collective data distribution closely resembling the global dataset spread across all clients. This aligns with the intuition that effective device clustering should mitigate clients' local model biases during local model aggregations with ESs. This not only accelerates convergence but also diminishes the reliance on resource-intensive edge-to-cloud model aggregations, echoing observations noted in prior research \cite{ref8, ref10, ref14}. {\emph{{\romannumeral3})}} \emph{How to expedite the overall decision-making process required for determining client selection and C2E association (i.e., reduce the decision-making overhead)?} The dynamics of client availability, varying local workloads, and individual selfishness necessitate frequent decision-making at the network operator level, such as at each global model training round in HFL. The requirement for constant decision-making regarding the client recruitment and C2E association can lead to significant overhead (i.e., increased latency and energy consumption). Thus, such frequent decision-making can disrupt the training process and complicate the necessary synchronizations 
within the training period. 

\subsection{Overview and Summary of Contributions}\label{subsec:contribution}
We study HFL over the device-edge-cloud continuum with intermittent client participation. Due to client availability dynamics, client selection and C2E association must be frequently updated. We address the underexplored challenge of reducing decision-making overhead by formulating a joint client selection and C2E association problem and proposing a low-overhead stagewise decision-making methodology. In essence, our approach shifts most decision-making from real time, as commonly done in FL and HFL, to long-term planning, thus accelerating training. To remain adaptive, it further incorporates lightweight real-time refinements. Our main contributions are:

\noindent$\bullet$ We are interested in addressing the challenge of joint client selection and C2E association for HFL under intermittent client participation, by formulating an optimization aimed at
minimizing system costs, while maintaining a tolerable level of data heterogeneity among clients connected to each ES. We
reveal that this problem is NP-hard, rendering its solution
intractable, especially for large-scale networks.

\noindent$\bullet$ To manage the complexity, we implement
a stagewise methodology that breaks the problem into two
subproblems, solved at different points in the timeline: Plan A, deployed in advance to model training, and Plan
B, implemented concurrently with model training. These
stages function as asynchronous yet complementary. Plan A
identifies clients who are likely to participate in future model
training, while Plan B ensures seamless model training if
clients selected under Plan A become unavailable.

\noindent$\bullet$ In Plan A, we introduce the concept of ``client continuity'' and derive closed-form constraints on tolerable data heterogeneity by modeling them as risks. To identify suitable long-term clients, we propose a strategy that combines gain-of-cost minimization-promoted C2E association and local iteration-based determination. This stage provides a novel pre-decision-making process where pre-selected clients can directly participate in each global training round as long as they are online, thus reducing the burden of real-time decision-making. In Plan B, we refine the decisions from Plan A in cases some long-term clients are unavailable (either dropped out or are offline) during a practical model
training round. To address this, we introduce a cluster-based client update algorithm for the rapid exploration of suitable alternative clients, achieving an approximate optimal solution for real-time client recruitment.

\noindent$\bullet$ We conduct extensive experiments on various datasets
(i.e., MNIST, Fashion-MNIST, CIFAR-10 and CIFAR-100, {\color{red}{and ALI}}), verifying our superior performance as comparing with benchmark methods in
terms of model accuracy, system cost, and time efficiency.

\section{Literature Investigation} \label{sec:related work}
Numerous studies investigated balancing the model training performance (e.g., model accuracy) and cost-effectiveness for FL. For example, \cite{ref15,ref16,ref17} study model compression, gradient quantization, and coding techniques to reduce the model aggregation overhead. Also, \cite{ref18,ref19,ref20,ref61} conduct communication/computation resource allocation for FL. Further, \cite{ref21,ref22} aim to reduce the required global aggregation rounds. These works study FL with direct client-to-cloud communications, which can cause increased latency, energy consumption, and congestion over the backhaul links.

To address the above discussed issues, HFL was proposed by \textit{Liu et al.} \cite{ref8}, introducing an intermediate aggregation layer (e.g., formed by ESs) to reduce the frequency of direct communications between clients and the remote cloud. \textit {Feng et al.} \cite{ref24} further optimized subcarrier assignment, transmit power of clients, and computation resources in HFL. \textit {Lin et al.} \cite{ref25} proposed control algorithms for delay-aware HFL, obtaining policies to mitigate communication latency in HFL. 

Considering the constraints of limited resources in wireless networks, developing the joint client selection and the C2E association strategy for HFL emerges as a viable approach to improve model training efficiency, which has attracted a great deal of research attention. For example, \textit {Luo et al.} \cite{ref5} conducted resource allocation and C2E association for HFL, enabling great potentials in low latency and energy-efficient FL. \textit{Deng et al.} \cite{ref10} minimized the total communication cost required for model learning with a target accuracy, by making decisions on edge aggregator selection and node-edge associations under the HFL framework. \textit{Su et al.} \cite{ref26} developed an online learning-based client selection algorithm for ESs, based on empirical learning results, reducing the cumulative delay on computation and communication. \textit{Qu et al.} \cite{ref27} observed the context of local computing and transmission of client-ES pairs, making client selection decisions to maximize the network operator’s utility, under a given budget. 

Besides HFL, several novel architectures have been proposed to address the high communication overhead caused by frequent interactions with the remote server. For example, \textit{Guo et al.} \cite{ref53} proposed a hybrid local stochastic gradient descent (HL-SGD) algorithm, which leverages the availability of fast D2D links to accelerate convergence and reduce training time under non-IID data distributions. However, similar to traditional FL and HFL, the central entity in HL-SGD may still become a bottleneck and suffer from a single point of failure, raising concerns about fault tolerance and scalability. To tackle this issue, a cooperative federated edge learning (CFEL) \cite{ref54,ref55} framework was introduced, which exploits multiple aggregators and eliminates single points of failure, making it more scalable than prior FL frameworks. \textit{Castiglia et al.} \cite{ref54} proposed multi-level local SGD (MLL-SGD) in a two-tier communication network with heterogeneous workers with IID data distributions. \textit{Zhang et al.} \cite{ref55} designed a novel federated optimization method called cooperative edge-based federated averaging (CE-FedAvg), which efficiently learns a shared global model over the collective dataset with non-IID data of all edge devices, under the orchestration of a distributed network of cooperative edge servers.

Existing studies have advanced model training by improving client participation and optimizing network orchestration, laying a solid foundation for efficient FL. However, most assume persistent client availability or overlook the decision-making overhead of client recruitment and association with ESs, namely, factors critical in dynamic, resource-constrained environments. To this end, we explore an underexamined research angle and propose stagewise decision-making for HFL under intermittent client availability. Our method employs two complementary plans for client selection and C2E association at different timescales, ensuring seamless and efficient model training.

\section{System Model}\label{sec:system model}
\begin{figure}
    \centering
	\includegraphics[trim=2cm 22cm 145cm 35cm, clip, width=\columnwidth]{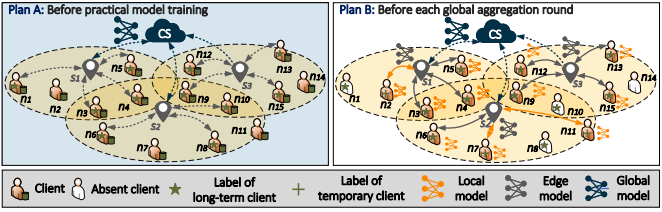}
	\caption{A schematic of stagewise desicion-making for HFL over dynamic device-edge-cloud continuum with intermittent client participation.}
\label{fig:framework}
\end{figure}

As shown in Fig. \ref{fig:framework}, we consider hierarchical federated learning (HFL) over a three-layer network hierarchy including {\emph{{\romannumeral1})}} a cloud server (CS) acting as the centralized aggregator, {\emph{{\romannumeral2})}} several edge servers (ESs), denoted by $\mathcal{S}=\left\{s_{1}, \ldots, s_{j}, \ldots, s_{|\mathcal{S}|}\right\}$, acting as intermediate aggregators, and {\emph{{\romannumeral3})}} a set of mobile devices (clients) $\mathcal{N}=\left\{\bm{n}_{1}, \ldots, \bm{n}_{i}, \ldots, \bm{n}_{|\mathcal{N}|}\right\}$. Each client $\bm{n}_{i} \in \mathcal{N}$ is described via a 4-tuple $\bm{n}_{i}=[\mathcal{D}_{i}, v_{i}, q_{i}, \xi_{i}]$, where $\mathcal{D}_{i}=\left\{\left(\boldsymbol{x}_{1}, y_{1}\right), \ldots,\left(\boldsymbol{x}_{k}, y_{k}\right), \ldots,\left(\boldsymbol{x}_{\left|\mathcal{D}_{i}\right|}, y_{\left|\mathcal{D}_{i}\right|}\right)\right\}$ denotes its local dataset and $\boldsymbol{x}_{k}$ and $y_{k}$ denote the input feature and label of the $k^{\text{th}}$ datapoint, respectively. Let ${\mathcal{D}=\left\{\mathcal{D}_{1}, \ldots, \mathcal{D}_{i}, \ldots, \mathcal{D}_{|\mathcal{N}|}\right\}}$ encapsulate the dataset profile of clients. Also, let $v_{i}$ denote the computation capability of $\bm{n}_i$ (i.e., its CPU frequency), and $q_{i}$ represent the transmit power of $\bm{n}_{i}$.  Learning takes place through a sequence of global and local model aggregations, where the index of an arbitrary global model aggregation is denoted by $g$ and the index of an arbitrary local aggregation is denoted by $\ell$. We capture the intermittent client availability across global aggregations via $\xi_i^{(g)}$, which is an indicator function for describing the participation of client $\bm{n}_i \in \mathcal{N}$. In particular, $\xi_{i}^{(g)}=1$ indicates that $\bm{n}_{i}$ takes part in the $g^{\text {th}}$ global aggregation, and $\xi_{i}^{(g)}=0$, otherwise ($\xi_{i}^{(g)}$ is detailed in Sec. \ref{subsec:client participation}).

Each global aggregation\footnote{``Global aggregation'' and ``global iteration'' are exchangeably used.} of HFL starts with broadcasting the global model from CS to clients, with ESs acting as relays. Afterwards, clients synchronize their local models with the global model, train their local models on their datasets, and then upload their model parameters to their assigned ESs. After aggregating the received models from their clients (i.e., conducting local model aggregation), ESs have two options: {\emph{{\romannumeral1})}} broadcasting the aggregated edge model back to their clients for another round of local model training (i.e., conclusion of one local/edge aggregation round); {\emph{{\romannumeral2})}} sending the aggregated edge model to the CS for global aggregation (i.e., conclusion of one global aggregation round). This procedure calls for addressing both client recruitment/selection (i.e., determining proper clients for training) and client-to-edge (C2E) association (i.e., assigning clients to  proper ESs), which given their combinatorial natures are NP-Hard (discussed in Sec. \ref{sec:stagewiseHFL}). Thus, the overhead of solving these problems can become unacceptable in large-scale networks, affecting the sustainability and practicality of HFL. 

Subsequently, our primary goal is \emph{to recruit clients and assign them to appropriate ESs at each global aggregation round, while ensuring {{\romannumeral1})} the reliability and performance of the system, meaning seamless execution of  HFL with an adequate number of clients and fast convergence of the trained global model; {{\romannumeral2})} low decision-making overhead, which involves minimizing the time spent on real-time decision-making to maximize the time available for model training; {{\romannumeral3})} reduced costs (i.e., energy consumption and latency) of model training.} We achieve this goal via a stagewise decision-making process, consisting of Plan A and Plan B, where in Plan A, a set of long-term clients (often those with higher probabilities of participation in model training) are selected for future global model aggregations in a pre-decision manner. Also, Plan B involves real-time recruitment and orchestration of short-term clients to compensate for long-term ones that are unavailable at each practical global aggregation round.
For instance, as illustrated in Fig. \ref{fig:framework}, consider the case where client $\bm{n}_1$ is pre-selected by Plan~A as a long-term participant. Due to dynamism, $\bm{n}_1$ may become unavailable before the actual training starts, which may result in a partial loss of training data. To mitigate this, Plan B activates a replacement strategy by selecting client $\bm{n}_2$ as a temporary participant, thereby filling the data gap and preserving both the seamlessness and efficiency of HFL progress.
Key notations are summarized in Appx. \ref{appx:notations}.
\subsection{Modeling of HFL Procedure of Our Interest}\label{subsec:HFL procedure}
We introduce $\boldsymbol{A}^{(g)}=\left\{a_{i, j}^{(g)} \mid \bm{n}_{i} \in \mathcal{N}, s_{j} \in \mathcal{S}\right\}$, where $a_{i, j}^{(g)}=1$ if
client $\bm{n}_{i}$ is \textit{selected/recruited and  associated with ES $s_{j}$}; otherwise $a_{i, j}^{(g)}=0$. We next formalize different HFL steps.

\noindent $\bullet$ \emph{\textbf{Step 1}}: Let $\bm{\omega}_{i}^{(g;\ell, t)}$ denote the local model of client $\bm{n}_{i}$, at edge aggregation round $\ell$ ($0\leq \ell \leq \mathcal{L}-1$), where $t$ is the index for local updates. After receiving the global model $\bm{\omega}^{(g)}$ broadcasted by CS at global aggregation $g$, each client $\bm{n}_{i}$ first synchronizes its local model as $\bm{\omega}_i^{(g;0,0)} \leftarrow \bm{\omega}^{(g)}$.  Then, 	$\bm{\omega}_{i}^{(g; \ell, t)}$ is obtained  via stochastic gradient descent (SGD) as
\begin{equation}
	\label{deqn_ex1a}
	\bm{\omega}_{i}^{(g; \ell, t)}=\bm{\omega}_{i}^{(g;\ell, t-1)}-\eta \widetilde{\nabla} F_{i}(\bm{\omega}_{i}^{(g;\ell, t-1)}),~~1\leq t\leq \mathcal{T},
\end{equation}
where $\eta$ is the learning rate and $\widetilde{\nabla}$ denotes the stochastic gradient obtained over a mini-batch of datapoints selected uniformly at random. Also, $F_i$ is the local loss function at $\bm{n}_i$ defined as $
	F_{i}\left(\boldsymbol{\omega}\right)=\frac{1}{d_{i}}{\sum_{k=1}^{d_{i}}f_{i}\left(\boldsymbol{x}_{k},y_{k};\boldsymbol{\omega}\right)},$
where $d_{i}=|\mathcal{D}_{i}|$ represents the data size of client $\bm{n}_{i}$, and $f_i$ is the loss function of the machine learning model (e.g., cross-entropy loss).

\noindent $\bullet$ \emph{\textbf{Step 2}}: Clients will upload their local model parameters to ESs after $\mathcal{T}$ local iterations, followed by each ES $s_j$ aggregating these models via federated averaging \cite{ref2} as
\begin{equation}
	\label{deqn_ex1a}
	\boldsymbol{\omega}_{j}^{(g;\ell+1)}={\sum_{i=1}^{|\mathcal{N}
			|}a_{i,j}^{(g)}\xi_{i}^{(g)}d_{i}\boldsymbol{\omega}_{i}^{(g;\ell,\mathcal{T})}}\big/D_{j}^{(g)},
\end{equation}
where ${D_{j}^{(g)} =\sum_{i=1}^{|\mathcal{N}|}a_{i,j}^{(g)}\xi_{i}^{(g)}d_{i}}$ captures the  dataset size of all clients associated with  ES $s_{j}$. The aggregated edge model $\boldsymbol{\omega}_{j}^{(g;\ell)}$ is sent to the CS if $\ell+1= \mathcal{L}$; otherwise it is returned to associated clients for the next local update. 

\noindent $\bullet$ \emph{\textbf{Step 3}}:  The CS aggregates the model parameters sent from the ESs to form a global model as 
\begin{equation}
	\label{deqn_ex1a}
	\boldsymbol{\omega}^{(g+1)}={\sum_{j=1}^{|\mathcal{S}|}D_{j}^{(g)}\boldsymbol{\omega}_{j}^{(g;\mathcal{L})}}\big/{D}^{(g)},
\end{equation}
where $D^{(g)} =\sum_{j=1}^{|\mathcal{S}|}D_{j}^{(g)}$, which is broadcasted back to clients  to initiate the next round of local model training.
\subsection{Modeling of Computation and Communication}\label{subsec:computation and communication}
The latency and energy consumption at each client $\bm{n}_{i} $ for conducting one SGD update are $t_i^{\textrm {cmp}}=\frac{c_{i}\zeta_i d_{i}}{v_{i}}$ and  $e_i^{\textrm {cmp}}=\alpha_{i}v_{i}^2 c_{i}\zeta_i d_{i}$, respectively, where $\zeta_i \in (0,1]$ is the fraction of local dataset in  SGD mini-batch. Here, $c_{i}$ is the number of cycles of processor (e.g., CPUs and GPUs) required for processing one datapoint, $\alpha_{i}$ is the chipset capacitance of 
$\bm{n}_{i}$, and $v_{i}$ is $\bm{n}_{i}$’s of processor frequency\footnote{The  processor frequency $v_i$ is a virtualized concept designed to capture different processor capabilities. For example, a larger $v_i$ may correspond to a GPU, while a smaller $v_i$ reflects a CPU-only device, which naturally expresses client heterogeneity. Similarly, it is common in FL studies \cite{ref10, ref26, ref50} to use such abstract parameters to represent client computing capacity rather than specifying processor types explicitly, thereby capturing device heterogeneity without binding to specific hardware.}  \cite{ref23}. 

We assume that each ES $s_j$ allocates a fixed bandwidth $B_j$ to each of its  clients\footnote{We assume that orthogonal frequency-division multiple access (OFDMA) is used for clients to upload their models to ESs, allowing us to ignore the interference (similar to \cite{ref30}). The C2E channels are assumed to be stable during each global aggregation duration \cite{ref31,ref32,ref48}.}, and the available bandwidth can support $M_j$ clients concurrently \cite{ref10}. Accordingly, uplink data rate from client $\bm{n}_i$ to ES $s_j$ can be computed as
$
	r_{i,j}^{(g)}=B_{j}\operatorname{log}_{2}\left(1+\frac{q_{i} h_{i,j}^{(g)}}{N_{0}B_{j}}\right)
$, where $h_{i,j}^{(g)}$ is the channel gain between $\bm{n}_i$ and $s_j$ \cite{ref28} during global aggregation $g$, and $N_0$ is the noise power spectral density. Let $\Omega$ denote the size (in bits) of the model parameters of clients \cite{ref8}. The transmission delay and energy consumption between $\bm{n}_i$ and $s_j$ during an edge aggregation in global aggregation round $g$ are given by $t_{i,j}^{(g), \textrm{com}}=\Omega/r_{i,j}^{(g)}$ and $e_{i,j}^{(g),\textrm{com}}=q_{i}t_{i,j}^{(g), \textrm{com}}$, respectively.  Subsequently, the overall delay and energy consumption of the interactions between $\bm{n}_i$ and $s_j$ to conduct an edge aggregation during global aggregation $g$ are
$
	\label{deqn_ex1a}
	T_{i,j}^{(g)}=\mathcal{T} t_{i}^{\textrm{cmp}}+t_{i,j}^{(g),\textrm{com}},
$ and
$
	\label{deqn_ex1a}
	E_{i,j}^{(g)}=\mathcal{T} e_{i}^{\textrm{cmp}}+e_{i,j}^{(g),\textrm{com}}
$, respectively.

We make a few standard assumptions to facilitate analysis. First, the computing time and energy consumption of ESs and CS for model aggregation are ignored (supported by \cite{ref24}). Second, for the wired transmissions between ESs and the CS, the transmission delay $T_j^{(g),\textrm{com}}$ and energy consumption $E_j^{(g),\textrm{com}}$ for ES $s_j$ are assumed to be known (as supported by \cite{ref29}, relying on the real-world simulations). Third, due to the large downlink bandwidth and transmit power of ESs, we neglect the downlink latency and energy consumption \cite{ref24}. 


Consequently, the overall training delay $\mathbb{T}^{(g)}$, and energy consumption $\mathbb{E}^{(g)}$ of global aggregation $g$ are given by
\begin{equation}
	\label{Time}
	\mathbb{T}^{(g)}=\max_{s_{j}\in\mathcal{S}}\left\{\mathcal{L} \max_{\bm{n}_{i}\in\mathcal{N}}\left\{a_{i,j}^{(g)}\xi_{i}^{(g)}T_{i,j}^{(g)}\right\}+T_{j}^{(g),\textrm{com}}\right\},
\end{equation}
\begin{equation}
	\label{Energy}
	\mathbb{E}^{(g)}=\sum_{s_{j}\in\mathcal{S}}\left(\mathcal{L} \sum_{\bm{n}_{i}\in\mathcal{N}}a_{i,j}^{(g)}\xi_{i}^{(g)}E_{i,j}^{(g)}+E_{j}^{(g),\textrm{com}}\right).
\end{equation}
\subsection{Modeling of Client Participation}\label{subsec:client participation}
Given the stochastic nature of client availability, the status indicator of client $\bm{n}_i$ in the $g^{\text{th}}$ global aggregation $\xi_i^{(g)}$ is probabilistic. To reflect this, we model it as a Bernoulli random variable (e.g., $\xi_i^{(g)}\sim\mathcal{B}\left(p_i\right)$), where $E[\xi_i^{(g)}]=p_i$.
Note that $p_i$ can be obtained via inspecting the client $\bm{n}_i$'s historical behavior\footnote{Probabilistic methods (e.g., Poisson point processes) and channel models (e.g., Gauss-Markov) are widely used to characterize network behavior from historical data \cite{ref45,ref46}. Their popularity stems from the fact that network conditions often evolve gradually, allowing a finite history to reliably guide future decisions. Leveraging such models enables effective capture of network dynamics, leading to more accurate forecasting and informed decision-making.}. In particular, let $\tau$ indicate the length of observation rolling window and $K$ denote the total number of windows. We estimate each $p_i$ according to the historical statistics of client $n_i$ in executing model training tasks over the previous $K\tau$ global iterations. Formally, at the end of the $K^\text{th}$ observation rolling window, we calculate the estimated online probability $p_i$ for client $n_i$ by combining these weighted observations as follows:
\begin{equation}
	\label{eq:window}
	\widetilde{p_i}=\sum_{\kappa=1}^{K}\frac{2\kappa}{K\left(K+1\right)}\sum_{g=\left(\kappa-1\right)\tau+1}^{\kappa\tau}\frac{\textbf{X}_i^{(g)}}{\tau},
\end{equation}
where $\textbf{X}_i^{(g)}$ is the observed value of $\xi_i^{(g)}$ at the $g^\text{th}$ time instant of window $\kappa$. Fresher historical records (e.g., the $(K-1)^\text{th}$ window is 
closer to the present time than the $(K-2)^\text{th}$) are assigned with larger weights (i.e., $\tfrac{\kappa}{\sum_{\kappa=1}^{K}\kappa}=\tfrac{2\kappa}{K\left(K+1\right)}$
) in~\eqref{eq:window}. 

During the training process, the online status of clients can be continuously updated and incorporated into the historical data, enabling real-time improvements in predicting their future online behaviors. Consequently, the estimation of a client's online probability and the implementation of Plan A are not limited to the initial phase of training but can also be performed repeatedly throughout the training period (an example is shown in Fig. \ref{fig:performance with multiple plan a}, Appx. \ref{appx:evaluation I}). Also, please note that in the main text, for analytical simplicity, we assume that Plan A is implemented for once before the training process, namely, before the first global iteration starts. 

\section{Stagewise Client Selection and C2E Association over  Device-Edge-Cloud Hierarchy}
\label{sec:stagewiseHFL}
\vspace{2mm}
This section elaborates on our stagewise HFL mechanism, where our major goal is \emph{to determine appropriate clients for each ES to accelerate the model training, while minimizing the training delay (i.e., the time spent on transmission and data processing) along with the energy costs (i.e., the energy consumption of clients induced by model parameter transfer and local training)}.

\vspace{2mm}
\subsection{Primary Optimization Formulation}
\label{subsec:problem formulation}
\vspace{2mm}
\subsubsection{Key criteria in constraint design: reaching adequate and balanced data for model training}
\label{subsubsec:dmin and kldmax}
\vspace{2mm}
To ensure a desired convergence rate for the global model, it is essential to engage enough clients (corresponding to enough training data\cite{ref20}) while keeping the heterogeneity of data of clients associated with each ES within a reasonable range, to avoid local model bias. To this end, we begin by introducing Kullback-Leibler divergence (KLD) \cite{ref10,ref47} as a measure to quantify the difference between the data distribution covered by each ES and a reference data distribution (e.g., the uniform distribution), attempting to assess the degree of data heterogeneity across ESs without exposing their data. Considering a supervised learning task where the input data is divided into $|\mathcal{Z}|$ classes based on the label set ${\mathcal{Z}=\left\{z_{1}, \ldots, z_{h}, \ldots, z_{|\mathcal{Z}|}\right\}}$. We define the set of datapoints in  client $\bm{n}_i$'s dataset with label $z_h$ as   $\bm{y}_{i}(h)=\left\{\left(\boldsymbol{x}_{k}, y_{k}\right) \in \mathcal{D}_i \mid y_{k}=z_{h}\right\}$. Accordingly, the KLD of edge data at ES $s_j$ can be defined as follows:
\begin{equation}
	\label{kld}
	\operatorname{KLD}\left(\boldsymbol{P}^{(g)}_j||\boldsymbol{Q}\right)=\sum_{h=1}^{|\bm{\mathcal{Z}}|}\boldsymbol{P}_{j}^{(g)}(h)\operatorname{log}\frac{\boldsymbol{P}_j^{(g)}\left(h\right)}{\boldsymbol{Q}\left(h\right)},\forall s_{j}\in \mathcal{S},
\end{equation}
where 
\begin{equation}
	\label{distribution}
	\boldsymbol{P}_j^{(g)}\left(h\right)=\frac{\sum_{i=1}^{|\mathcal{N}|}a_{i, j}^{(g)}\xi_{i}^{(g)}\boldsymbol{y}_{i}(h)}{D_{j}^{(g)}},\forall s_{j}\in\mathcal{S},\forall z_{h}\in\mathcal{Z}.
\end{equation}
Specifically, $\boldsymbol{P}^{(g)}_j$ represents the data distribution of ES $s_j$ at global aggregation $g$, while $\boldsymbol{Q}$ describes the reference data distribution shared to all clients \cite{ref33}. Upon constraining the KLD acorss ESs, we can impose a high level of similarity across the associated clients, and thus a faster model convergence due to a lower model bias (i.e., constraint \eqref{p0_9a} in the later formulation).
Also, since having sufficient training data plays a crucial role in the model convergence, we consider a tolerable minimum data size $D_{\textrm {min}}$ to be covered by each ES (i.e., constraint \eqref{p0_9b} in the later formulation).

\vspace{2mm}
\subsubsection{Optimization problem}
\label{subsubsec:p0 formulation}
We next formulate the joint problem of client selection and C2E association for each global aggregation $g$ as $\bm{\mathcal{P}}^{(g)}_0$, with the goal of minimizing the weighted sum of delay $\mathbb{T}^{(g)}$ and energy cost $\mathbb{E}^{(g)}$, under KLD and data size constraints: 
\begin{equation}
	\bm{\mathcal{P}}_0^{(g)}:~~\underset{\boldsymbol{A}^{(g)}}{\min}~~ \lambda_{t}\mathbb{T}^{(g)}+\lambda_{e}\mathbb{E}^{(g)}
\end{equation}
\setcounter{equation}{8}
\vspace{-7mm} 
\begin{subequations}\label{p0}{
		
		\begin{align}
			\text{s.t.}~~~~&
			\operatorname{KLD}\left(\boldsymbol{P}_j^{(g)}||\boldsymbol{Q}\right)\le {\it{KLD}}_{\textrm {max}}, \forall s_{j}\in\mathcal{S}\label{p0_9a}\\
			&D_j^{(g)}\ge D_{\textrm {min}}, \forall s_{j}\in\mathcal{S}\label{p0_9b}\\
			&a_{i,j}^{(g)}\in\left\{0,1\right\},\forall \bm{n}_{i}\in\mathcal{N},\forall s_{j}\in\mathcal{S}\label{p0_9c}\\
			&\sum_{j=1}^{|\mathcal{S}|}a_{i,j}^{(g)}\le 1,\forall \bm{n}_{i}\in\mathcal{N}\label{p0_9d}\\
			&\sum_{i=1}^{|\mathcal{N}|}a_{i,j}^{(g)}\le M_j,\forall s_{j}\in\mathcal{S}\label{p0_9e}\\
			&a_{i,j}^{(g)}=0,\text{if}\,\,\xi_{i}^{(g)}=0,\forall \bm{n}_{i}\in\mathcal{N}, \forall s_{j}\in\mathcal{S}\label{p0_9f}
	\end{align}}
\end{subequations}

\noindent
where $\lambda_t$ and $\lambda_e$ are weighting coefficients, signifying the importance of energy consumption and delay. Constraint \eqref{p0_9a} restricts the KLD of the data covered by each ES below a maximum threshold, denoted as ${\it{KLD}}_{\textrm {max}}$, to ensure balanced training data across the ESs (which reflects in low model bias). Constraint \eqref{p0_9b} ensures sufficient training data contributed by clients associated with each ES. Constraint \eqref{p0_9c} forces binary values for $a_{i,j}^{(g)}$, constraint \eqref{p0_9d} ensures that each client is linked to at most one ES, while constraint \eqref{p0_9e} limits the number of clients that are covered by each ES $s_j$. Lastly, constraint \eqref{p0_9f} ensures that only available/participating clients are assigned to the ESs.

Optimization $\bm{\mathcal{P}}^{(g)}_0$ is a 0-1 integer programming (01-IP), making it NP-Hard \cite{ref36}. Specifically, as the number of clients increases (i.e., as $\xi_i^{(g)}$ grows), identifying and associating clients with proper ESs can become increasingly complex, which is why many studies in FL do not prioritize optimal solutions \cite{ref50,ref51,ref52}. To mitigate the excessive solution overhead that could disrupt the continuity of model training, we reformulate $\bm{\mathcal{P}}^{(g)}_0$ into two subproblems, representing two asynchronous yet complementary stages of decision-making\footnote{We solve problem $\bm{\mathcal{P}}^{(g)}_0$ from a unique angle by segmenting the into two distinct stages, diverging from the conventional approach of simplifying a complex problem, which often involves two easier subproblems. Consequently, while the subproblems in Plan A and Plan B may still be complex, they possess a reduced problem size.}. These two subproblems are addressed via two plans. {\emph{{\romannumeral1})}} \emph{Plan A} entails a pre-decision-making process in advance to practical model training. It conducts long-term decision-making by relying on historical statistics of client participation. {\emph{{\romannumeral2})}} \emph{Plan B} is a real-time decision-making process that takes place at the beginning of each global iteration based on the present clients (i.e., those with $ \xi_i^{(g)}=1$). Specifically, Plan A identifies long-term clients who are likely to participate in subsequent global training rounds shifting the overhead of real-time decision-making to the long-term phase. Since solely relying on Plan A\footnote{Plan A is flexible and can be executed multiple times alongside model training, allowing pre-decisions to adapt to varying channel conditions, client capacities, and thus online probabilities. For simplicity, we first analyze the case where these factors remain constant, so Plan A runs once before training. In Appx. \ref{appx:evaluation I}, we extend to fluctuating scenarios, where Plan A runs repeatedly, for example, once before training and then synchronized with subsequent training rounds, with each new global iteration reflecting updated client settings.} is impractical (e.g., due to the unpredictable absence of long-term works), Plan B serves as a contingency strategy, identifying necessary clients during each global aggregation to ensure a seamless model training. The overall workflow of our stagewise decision-making mechanism is summarized in Alg. \ref{algorithm1} (provided in Appx. \ref{appx:pseudocode}), which explicitly illustrates how Plan A (pre-decision-making via Algs. \ref{algorithm2}–\ref{algorithm3} for solving $\bm{\mathcal{P}}_1^{(g)}$) and Plan B (real-time decision-making via Alg. \ref{algorithm4} for solving $\bm{\mathcal{P}}_2^{(g)}$) are coordinated throughout the training process.

\subsection{Problem Transformation}
\label{subsec:problem transformation}
We next tackle $\bm{\mathcal{P}}^{(g)}_0$ by decoupling it into the two subproblems $\bm{\mathcal{P}}^{(g)}_1$ and $\bm{\mathcal{P}}^{(g)}_2$,  addressed by Plan A and Plan B. In this section, we provide a practical and intuitive view for tackling the 0-1 integer programming problem, by leveraging clients’ historical information through a two-stage segmentation, to accelerate model training while balancing accuracy and system cost.

\subsubsection{Design of Plan A}
\label{subsubsec:plan a}
Since Plan A requires preselecting some clients and assign them to the appropriate ESs before the start of model training, we begin by obtaining its intended subproblem $\bm{\mathcal{P}}^{(g)}_1$. This subproblem offers an unique long-term perspective, with the aim of minimizing the  system cost by selecting relatively ``stable'' clients. In other words, these clients are more likely to participate in the subsequent global training rounds, thereby reducing the workload of Plan B, i.e., reducing the problem size and its search space through decreasing the number of clients that need to be considered.

To develop Plan A, one naive approach to reach its long-term objective is to minimize the expectation of the overall system cost given by \eqref{p0} under the distribution of clients participation (i.e., the expected value of $\lambda_t\mathbb{T}^{(g)} + \lambda_e\mathbb{E}^{(g)}$). Nevertheless, implementing this approach can inadvertently result in choosing clients with lower participation probabilities. Although this may reduce transmission delay and energy costs, it does not align with another key objective of Plan A, which is ensuring consistent and reliable client participation to reduce the overhead of real-time decision-making in Plan B. To cope with this, we first introduce a unique consideration by setting $\xi_i^{(g)}=1$ ($\forall \bm{n}_i \in \mathcal{N}$) in both $\mathbb{T}^{(g)}$ and $\mathbb{E}^{(g)}$, updated by $\widehat{\mathbb{T}}^{(g)}$ and $\widehat{\mathbb{E}}^{(g)}$, indicating that all the clients are supposed to be online in Plan A. We define a matrix $\widehat{\boldsymbol{A}}^{(g)}=\left\{\widehat{a}^{(g)}_{i, j} \mid \bm{n}_{i} \in \mathcal{N}, s_{j} \in \mathcal{S}\right\}$ to differentiate from the C2E solution of $\bm{\mathcal{P}}^{(g)}_0$, where $\widehat{a}_{i,j}^{(g)}=1$ indicates that $\bm{n}_i$ is selected and associated with $s_j$ as a long-term client in Plan A. If, $\widehat{a}_{i,j}^{(g)}=1$, client $\bm{n}_i$ will join each practical global iteration proactively as long as $\xi_i^{(g)}=1$. To guarantee that clients with a higher chance of participation in the training process are identified in Plan A, we develop the concept of ``client continuity''. This concept is measured through the following metric, ensuring the consistency of participation among the selected clients in Plan A:
\begin{equation}
	\label{continuity}
	\mathbb{C}^{(g)}=\left(\prod_{i: \sum_{j=1}^{|\mathcal{S}|}\widehat{a}_{i, j}^{(g)}=1} p_{i}\right)^{\frac{1}{\sum_{j=1}^{|\mathcal{S}|}\sum_{i=1}^{|\mathcal{N}|} \widehat{a}_{i, j}^{(g)}}}.
\end{equation}
In \eqref{continuity}, $\mathbb{C}^{(g)}$ describes the geometric mean of participating probabilities of selected clients for global aggregation $g$. Specifically, the index term $i : \sum_{j=1}^{|\mathcal{S}|}\widehat{a}_{i, j}^{(g)}=1$ counts the  clients who have been selected and associated with ESs. 

Apparently, Plan A employs pre-decisions, including either pre-decision steps made prior to the actual model training (standing for one primary focus of our paper) or additional adjustments implemented concurrently with the training process (e.g., estimation of online probability of clients). In doing so, directly obtaining the original constraints \eqref{p0_9a} and \eqref{p0_9b} presents a noteworthy challenge. For instance, the practical values of $\xi_i^{(g)}$  in each actual global aggregation round may be unknown. Consequently, we reformulate these constraints into probabilistic expressions, and express them as follows:
\begin{equation}
	\label{p1_11a}
	\operatorname{Pr}\left(\operatorname{KLD}\Big(\widehat{\boldsymbol{P}}_j^{(g)}||\boldsymbol{Q}\Big)>{\it{KLD}}_{\textrm {max}}-\Delta_k\right)\le\delta, \forall s_{j}\in\mathcal{S},\tag{11a}
\end{equation}
\begin{equation}
	\label{p1_11b}
	\operatorname{Pr}\left(\widehat{D}_j^{(g)}<D_{\textrm {min}}+\Delta_{d}\right)\le\varepsilon, \forall s_{j}\in\mathcal{S}.\tag{11b}
\end{equation}
where $\widehat{\boldsymbol{P}}_j^{(g)}$ and $\widehat{D}_j^{(g)}$ are the updates after replacing $a_{i,j}^{(g)}$ in $\boldsymbol{P}_j^{(g)}$ and $D_j^{(g)}$ with $\widehat{a}_{i,j}^{(g)}$, respectively; while $\operatorname{Pr}(\cdot)$ represents the probability. In these updated constraints, we have introduced two constants $\Delta_k$ ($\Delta_k>0$) and $\Delta_d$ ($\Delta_d>0$) to tune the likelihood of the satisfaction of the original ones (i.e., \eqref{p0_9a} and \eqref{p0_9b}), aiming to mitigate the risk of obtaining unsatisfactory KLD and data size during practical model training, by ensuring these values remain below thresholds $\delta$ and $\varepsilon$\footnote{Although these probabilistic constraints might result in suboptimal client selection, they enhance the robustness and effectiveness of the preliminary decisions in Plan A by taking a long-term perspective on constraints related to KLD and data size. Therefore, these assumptions are more likely to hold in practice under dynamic conditions, thereby accelerating the convergence of the global model.}. Accordingly, the optimization subproblem $\bm{\mathcal{P}}^{(g)}_1$ , which is intended to be solved in Plan A is given by
\begin{equation}
	\bm{\mathcal{P}}^{(g)}_1:~~\underset{\widehat{\boldsymbol{A}}^{(g)}}{\min}~~ \lambda_{t} \widehat{\mathbb{T}}^{(g)}+\lambda_{e}\widehat{\mathbb{E}}^{(g)}-\lambda_{c}\mathbb{C}^{(g)}
\end{equation}
\setcounter{equation}{10}
\vspace{-3mm} 
\begin{subequations}\label{p1}{	
		\setcounter{equation}{2}
		\begin{align}
			\text{s.t.}~~~~&
			\text{(11a), (11b)},\notag\\[-.2em]
			&\widehat{a}_{i,j}^{(g)}\in\left\{0,1\right\},\forall \bm{n}_{i}\in\mathcal{N},\forall s_{j}\in\mathcal{S}\label{p1_11c}\\
			&\sum_{j=1}^{|\mathcal{S}|}\widehat{a}_{i,j}^{(g)}\le 1,\forall \bm{n}_{i}\in\mathcal{N}\label{p1_11d}\\
			&\sum_{i=1}^{|\mathcal{N}|}\widehat{a}_{i,j}^{(g)}\le M_j,\forall s_{j}\in\mathcal{S}\label{p1_11e}
	\end{align}}
\end{subequations}

\noindent
Constraints \eqref{p1_11a} and \eqref{p1_11b} in $\bm{\mathcal{P}}^{(g)}_1$ aim to mitigate the risk of obtaining unsatisfactory KLD and data size during practical model training, by ensuring these values remain below thresholds $\delta$, and $\varepsilon$. Constraint \eqref{p1_11d} guarantees that each client can be connected to no more than one ES, while constraint \eqref{p1_11e} caps the total number of clients that can be simultaneously connected to each ES during every edge aggregation. Note that obtaining the close form of \eqref{p1_11a} and \eqref{p1_11b} for arbitrary distribution of client participation is highly challenging. To tackle this challenge, we use the Markov inequality \cite{ref44} to convert them into tractable forms, which are given by \eqref{p1_11f} and \eqref{p1_11g} below (details of derivations of these transformations can be found in Appx. \ref{appx:derivation of 11f} and Appx. \ref{appx:derivation of 11g}). In Appx. \ref{appx:evaluation II} and Appx. \ref{appx:evaluation III}, we have included more extended experimental evaluations, to further substantiate the theoretical soundness and practical efficacy of our proposed constraints.
\begin{multline}
	\label{p1_11f}
	\prod_{i : \widehat{a}_{i,j}^{(g)}=1}\left(1-p_i\right)+\Big(1-\prod_{i : \widehat{a}_{i,j}^{(g)}=1} \left(1-p_i\right)\Big)\\[-.1em]
	\times\sum_{h=1}^{|\mathcal{Z}|}\frac{\bm{G}_{j}^{(g)}(h)}{{\it{KLD}}_{\textrm {max}}-\Delta_k}\le \delta, \forall s_{j}\in\mathcal{S}\tag{11f}
\end{multline}
\begin{equation}
	\label{p1_11g}
	\sum_{i=1}^{|\mathcal{N}|}\widehat{a}_{i,j}^{(g)}p_{i}d_{i}\ge\left(D_{\textrm {min}}+\Delta_d\right)\left(1-\varepsilon\right), \forall s_{j}\in\mathcal{S}\tag{11g}
\end{equation}
\noindent where $\bm{G}_{j}^{(g)}(h)$ is a piecewise function related to the upper bound of  $\operatorname{KLD}(\widehat{\boldsymbol{P}}_j^{(g)}||\boldsymbol{Q})$. It can be construed that with replacing the above two constrains instead of \eqref{p1_11a} and \eqref{p1_11b} in $\bm{\mathcal{P}}^{(g)}_1$, $\bm{\mathcal{P}}^{(g)}_1$ will be a non-linear 01-IP, rendering the attainment of its optimal solution  challenging  \cite{ref10,ref37}. Also, the constraints on risks, specifically \eqref{p1_11f} and \eqref{p1_11g}, add another layer of complexity. To accelerate reaching an approximate solution for $\bm{\mathcal{P}}^{(g)}_1$, we delve into designing two lightweight algorithms, which are \underline{g}ain-\underline{o}f-\underline{c}ost \underline{min}imization-promoted \underline{C2E} \underline{a}ssociation (GoCMinC2EA) for addressing the C2E association problem, and \underline{l}ocal \underline{i}teration-based \underline{long}-term \underline{client} \underline{d}etermination (LILongClientD) for solving the long-term client selection problem. These two algorithms work iteratively until reaching the near optimum $\bm{\mathcal{P}}^{(g)}_1$. For brevity, we use $\widehat{F}^{(g)}=\lambda_{t} \widehat{\mathbb{T}}^{(g)}+\lambda_{e} \widehat{\mathbb{E}}^{(g)}-\lambda_{c}\mathbb{C}^{(g)}$ to denote the cost function in Plan A -- to distinguish it from $F^{(g)}=\lambda_t\mathbb{T}^{(g)}+\lambda_e\mathbb{E}^{(g)}$.

\paragraph*{\small A. Gain-of-cost minimization-promoted C2E Association (GoCMinC2EA) in Plan A }

\noindent We outline our approach for addressing the C2E association problem associated with $\bm{\mathcal{P}}_1^{(g)}$, which can be described as: \textit{how to assign clients within a specified long-term set, $\mathcal{N}^s$, to a ESs to minimize system latency and energy costs, while guaranteeing service continuity within an acceptable risk threshold?} To tackle this, we propose GoCMinC2EA detailed in Alg. \ref{algorithm2}.

Alg. \ref{algorithm2} considers a set of long-term clients $\mathcal{N}^s$ (obtained later by Alg. \ref{algorithm3}). It first allocates clients exclusively covered by a single ES to that ES (lines 2-3). For clients covered by multiple ESs, we employ a greedy strategy to link each client $\bm{n}_i$ with one of its accessible ESs from the set $\mathcal{S}_i$ (lines 5-10), aiming to reduce the value of $\widehat{F}^{(g)}$. For unmatched clients, GoCMinC2EA checks the variation denoted by $\Delta \widehat{F}^{(g)}_{i,j}$: the difference of the value of $\widehat{F}^{(g)}$ before (i.e., $\widehat{F}^{(g)}_{(-)}$) and after (i.e., $\widehat{F}^{(g)}_{(+)}$) assigning client $\bm{n}_i \in \mathcal{N}^o$ to ES $s_j \in \mathcal{S}_i$ (line 6). By evaluating all potential mappings, we allocate a client to an ES that yields the minimum change in $\widehat{F}^{(g)}$, denoted as $\Delta \widehat{F}_{i^*,j^*}^{(g)}$, thereby optimizing the system's performance. Following this assignment, we update the set of not-associated clients $\mathcal{N}^o$ (lines 7-9). We repeat this procedure until every client in $\mathcal{N}^o$ has been assigned to an ES.

Constraints \eqref{p1_11e}–\eqref{p1_11g} are checked after all clients are associated with ESs. If the conditions are satisfied, we set $\widehat{F}{\textrm {min}}^{(g)}=\widehat{F}^{(g)}{(+)}$ to indicate a feasible association; otherwise, the association is marked infeasible, and a backtracking algorithm is used to refine the greedy solution. In this process, clients are reconsidered one by one in reverse order. For instance, if assigning client $n_1$ to ES $s_1$ violates a constraint, we remove this mapping and try alternative ESs (e.g., $s_2$ or $s_3$). If no feasible option exists for $n_1$, we step back to the previous client (e.g., $n_2$), cancel its mapping, and attempt to reassign it to a different ES, until reaching a feasible solution.
\setcounter{algocf}{1}
\begin{algorithm}[h]
		\footnotesize
	\caption{Gain-of-cost minimization-promoted C2E association (GoCMinC2EA)}\label{algorithm2}
	\SetKwInOut{Input}{Input}\SetKwInOut{Output}{Output}
	\Input{the set of ES $\mathcal{S}_i$ that client $\bm{n}_i$ can associate with; a given long-term client set $\mathcal{N}^s$}
	\Output{the C2E association result $\widehat{\boldsymbol{A}}^{(g)}=\left\{\widehat{a}_{i, j}^{(g)} \mid \bm{n}_{i} \in \mathcal{N}, s_{j} \in \mathcal{S}\right\}$; the minimum value $\widehat{F}^{(g)}_{\textrm {min}}$ of cost function $\widehat{F}^{(g)}$}
	Initialization: $\widehat{\boldsymbol{A}}^{(g)}\leftarrow\boldsymbol{0}$, $\Delta\boldsymbol{F}^{(g)}\leftarrow\boldsymbol{0}$, $\mathcal{N}^o \leftarrow \emptyset$, $\widehat{F}^{(g)}_{\textrm {min}}\leftarrow \it{INF}$\footnotemark;
	
	$\mathcal{N}^o \leftarrow \mathcal{N}^o \cup \left\{\bm{n}_i \mid~ |\mathcal{S}_i|>1\right\}$;
	
	Associate each client $\bm{n}_i \in \mathcal{N}^s \textbackslash{} \mathcal{N}^o$ with its available ES and update $\widehat{\bm{A}}^{(g)}$;
	
	\Repeat{$\widehat{F}^{(g)}_{\text{\textnormal{min}} }<\it{INF}$ \textnormal{or no feasible association can be found}}{
		\Repeat{$
			\mathcal{N}^o=\emptyset$}{
			Calculate all feasible cost function variations $\Delta \widehat{F}_{i,j}^{(g)}=\widehat{F}^{(g)}_{(+)}-\widehat{F}^{(g)}_{(-)}$ for associating client $\bm{n}_i \in \mathcal{N}^o$ with each ES $s_j \in \mathcal{S}_i$, and record them in $\Delta \bm{F}^{(g)}$;
			
			$\Delta \widehat{F}^{(g)}_{i^*,j^*} \hspace{-.5mm} \leftarrow \hspace{-.5mm} \operatorname{min}{\left\{\Delta \widehat{F}^{(g)}_{i,j} \mid 
				\Delta \widehat{F}^{(g)}_{i,j}{\in} \Delta \bm{F}^{(g)},
				\Delta \widehat{F}^{(g)}_{i,j}{>}0\right\}}$;		
			
			$\widehat{a}_{i^*,j^* }^{(g)}\leftarrow 1$;
			
			$\mathcal{N}^o \leftarrow \mathcal{N}^o \textbackslash{} \left\{\bm{n}_{i^*}\right\}$;
		}
		\If{\textnormal{constraints \eqref{p1_11e}, \eqref{p1_11f} and \eqref{p1_11g} are satisfied}}{
			$\widehat{F}_{\textrm {min}}^{(g)} \leftarrow \widehat{F}_{(+)}^{(g)}$;}
		\Else{
			Disassociate and mark these associations as ``infeasible association'';
		}
	}
	
	{\bf{return}} $\widehat{\boldsymbol{A}}^{(g)}$, $\widehat{F}^{(g)}_{\textrm {min}}$;
\end{algorithm}
\footnotetext{We define $\textit{INF}$ as a large number, ensuring that any value of $\widehat{F}^{(g)}_{\textrm {min}}$ satisfying all constraints will be less than $\it{INF}$.}

If the solution obtained from GoCMinC2EA either  satisfies all the constraints or can be modified into a feasible solution with minimal backtracking, the computational complexity of Alg. \ref{algorithm2} could closely approximate that of its greedy component (the best case), i.e., $\mathcal{O}(|\mathcal{N}^o|)$. This implies that the additional steps required for verification and adjustment may not increase the computations, especially if the need for backtracking is limited. However, a worst case may occur where extensive backtracking is necessary to identify a feasible solution, in which case the complexity of GoCMinC2EA will approximate that of the backtracking component. This suggests that, in situations requiring significant adjustments to the initial mappings, the computations and time needed to reach a feasible solution could increase, aligning  with the complexity of the backtracking, e.g., $\mathcal{O}(\prod_{i=1}^{|\mathcal{N}^o|}|\mathcal{S}_i|)$.

\paragraph*{\small B. Local Iteration-based Long-term Client Determination (LILongClientD) in Plan B}
\noindent We next delve into LILongClientD, which aims to recruit long-term clients  that induce a low cost of model training. Since there are $2^{|\mathcal{N}|}$ combinations of clients, considering all combinations is impractical. To this end, we design a lightweight algorithm called LILongClientD (presented in Alg. \ref{algorithm3}) that updates the selection of clients in an iterative manner. This method optimizes the client selection by focusing on incremental performance gains rather than evaluating every possible combination. We also adopt the introduced GoCMinC2EA algorithm to determine the appropriate C2E associations, aiming to minimize the cost function $\widehat{F}^{(g)}$. In a nutshell, for a given set of clients, $\mathcal{N}^s$, Alg. \ref{algorithm3} evaluates whether an alternative set of clients with a better solution than the present one exists. Specifically, in LILongClientD, we design three key operations: \emph{Add} (lines 4-11),  \emph{Remove} (lines 12-19) and \emph{Exchange} (lines 20-28), for refining the selected clients, with details moved to Appx. \ref{appx:operations}.
\begin{algorithm}[h]
	\footnotesize
	\caption{Local iteration-based long-term client determination (LILongClientD)}\label{algorithm3}
	\SetKwInOut{Input}{Input}\SetKwInOut{Output}{Output}
	\Input{the set of ES $\mathcal{S}$; the set of clients $\mathcal{N}$}
	\Output{the C2E association result $\widehat{\boldsymbol{A}}^{(g)}=\left\{\widehat{a}^{(g)}_{i, j} \mid \bm{n}_{i} \in \mathcal{N}, s_{j} \in \mathcal{S}\right\}$; the minimum value $\widehat{F}^{(g)}_{\textrm {min}}$ of the cost function $\widehat{F}^{(g)}$}
	Initialization: a feasible clients selection solution $\mathcal{N}^s$;
	
	Obtain $\widehat{\boldsymbol{A}}^{(g)}$, $\widehat{F}^{(g)}_{\textrm {min}}$ from GoCMinC2EA ($\mathcal{N}^s$);
	
	\Repeat{\textnormal{no adjustment can reduce $\widehat{F}^{(g)}_{\textrm {min}}$}}{%
		
		//\textbf{Operation 1. Add}
		
		$\mathcal{N}^s_0 \leftarrow \mathcal{N}^s$;
		
		\For{$\bm{n}_i \in \mathcal{N} \textbackslash{} \mathcal{N}^s_0$}{
			Obtain $\widehat{\boldsymbol{A}}^{(g)}_{\cup \left\{\bm{n}_i\right\}}$, $\widehat{F}^{(g)}_{\cup \left\{\bm{n}_i\right\}}$ from GoCMinC2EA ($\mathcal{N}^s_0 \cup \left\{\bm{n}_i\right\}$) in Alg. \ref{algorithm2};
			
			\If{$\widehat{F}^{(g)}_{\cup \left\{\bm{n}_i\right\}}<\widehat{F}^{(g)}_{\textnormal {min}}$}{
				$\widehat{F}^{(g)}_{\textrm {min}} \leftarrow \widehat{F}^{(g)}_{\cup \left\{\bm{n}_i\right\}}$;
				
				$\mathcal{N}^s \leftarrow \mathcal{N}^s_{0}\cup \left\{\bm{n}_i\right\}$;
				
				$\widehat{\boldsymbol{A}}^{(g)}\leftarrow \widehat{\boldsymbol{A}}^{(g)}_{\cup \left\{\bm{n}_i\right\}}$;
			}
		}
		//\textbf{Operation 2. Remove}
		
		$\mathcal{N}^s_0 \leftarrow \mathcal{N}^s$;
		
		\For{$\bm{n}_i \in \mathcal{N}^s_0$}{
			Obtain $\widehat{\boldsymbol{A}}^{(g)}_{\textbackslash{} \left\{\bm{n}_i\right\}}$, $\widehat{F}^{(g)}_{\textbackslash{} \left\{\bm{n}_i\right\}}$ from GoCMinC2EA ($\mathcal{N}^s_0 \textbackslash{} \left\{\bm{n}_i\right\}$) in Alg. \ref{algorithm2};
			
			\If{$\widehat{F}^{(g)}_{\textbackslash{} \left\{\bm{n}_i\right\}}<\widehat{F}^{(g)}_{\textnormal {min}}$}{
				$\widehat{F}^{(g)}_{\textrm {min}} \leftarrow \widehat{F}^{(g)}_{\textbackslash{}\left\{\bm{n}_i\right\}}$;
				
				$\mathcal{N}^s \leftarrow \mathcal{N}^s_0 \textbackslash{} \left\{\bm{n}_i\right\}$;
				
				$\widehat{\boldsymbol{A}}^{(g)} \leftarrow \widehat{\boldsymbol{A}}^{(g)}_{\textbackslash{}\left\{\bm{n}_i\right\}}$;
			}
		}
		//\textbf{Operation 3. Exchange}
		
		$\mathcal{N}^s_0 \leftarrow \mathcal{N}^s$;
		
		\For{$\bm{n}_i \in \mathcal{N}^s_0$}{
			\For{$\bm{n}_j \in \mathcal{N}-\mathcal{N}^s_0$}{
				Obtain $\widehat{\boldsymbol{A}}^{(g)}_{\textbackslash{}\left\{\bm{n}_i\right\}\cup \left\{\bm{n}_j\right\}}$, $\widehat{F}^{(g)}_{\textbackslash{}\left\{\bm{n}_i\right\}\cup \left\{\bm{n}_j\right\}}$ from GoCMinC2EA ($\mathcal{N}^s_0 \textbackslash{} \left\{\bm{n}_i\right\} \cup \left\{\bm{n}_j\right\}$) in Alg. \ref{algorithm2};
				
				\If{$\widehat{F}^{(g)}_{\textbackslash{}\left\{\bm{n}_i\right\}\cup \left\{\bm{n}_j\right\}}<\widehat{F}^{(g)}_{\textnormal {min}}$}{
					$\widehat{F}^{(g)}_{\textrm {min}} \leftarrow \widehat{F}_{\textbackslash{}\left\{\bm{n}_i\right\}\cup \left\{\bm{n}_j\right\}}^{(g)}$;
					
					$\mathcal{N}^s \leftarrow \mathcal{N}^s_0 \textbackslash{} \left\{\bm{n}_i\right\} \cup \left\{\bm{n}_j\right\}$;
					
					$\widehat{\boldsymbol{A}}^{(g)} \leftarrow \widehat{\boldsymbol{A}}^{(g)}_{\textbackslash{}\left\{\bm{n}_i\right\}\cup \left\{\bm{n}_j\right\}}$;
				}
			}
		}
	}
	{\bf{return}} $\widehat{\boldsymbol{A}}^{(g)}$, $\widehat{F}^{(g)}_{\textrm {min}}$;
\end{algorithm}

Considering Alg. \ref{algorithm3}, by initializing a randomly selected client set, LILongClientD repeats the above three operations until no adjustments reduce $F_{\textrm {min}}^\prime$. According to the number of iterations within the internal loop on \emph{Add}, \emph{Remove} and \emph{Exchange} and the complexity of Alg. \ref{algorithm2}, denoted by $\mathcal{O}(V)$ (where $\mathcal{O}(|\mathcal{N}^o|)\le \mathcal{O}(V)\le\mathcal{O}(\prod_{i=1}^{|\mathcal{N}^o|}|\mathcal{S}_i|)$),  their computational complexities are $\mathcal{O}(V|\mathcal{N}\textbackslash{} \mathcal{N}^s|)$, $\mathcal{O}(V|\mathcal{N}^s|$ and $\mathcal{O}(V|\mathcal{N}||\mathcal{N}^s|$, respectively. Also, the complexity of Alg. \ref{algorithm3} depends heavily on the number of repetitions\footnote{Here, ``repetition'' refers to the executions in lines 4-28, Alg. \ref{algorithm3}, i.e., each repetition involves executing each of the three operations once.} assumed to be $\beta$. In particular, the  computation complexity of Alg. \ref{algorithm3} is $\mathcal{O}(\beta V|\mathcal{N}||\mathcal{N}^s|)$.

Unlike prior exhaustive or greedy search methods \cite{ref5,ref10,ref37}, which suffer from heavy backtracking and poor scalability, we design a novel two-stage framework: Plan A conducts offline optimization to handle complex constraints, while Plan B performs lightweight online refinement during training. This design ensures that even as $|\mathcal{N}|$ grows, the real-time decision-making remains highly efficient, effectively mitigating the runtime bottleneck encountered in conventional single-stage decision-making approach.

\subsubsection{Design of Plan B}
\label{subsubsec:plan b}
Apparently, implementing Plan A represents a coexistence of both risks and opportunities, as it replies on the estimation of uncertain online behaviors of clients. Although we have made certain efforts to control potential risks that can leave negative impacts on model training (e.g., by using \eqref{p1_11a} and \eqref{p1_11b}), one case may raise, in which some of the long-term clients are absent from the training process, further degrading the performance of Plan A. Thus, when long-term decision-making becomes ineffective due to the potential absence, we switch to real-time decision-making (Plan B), which is crucial to ensure a desired training performance. In Plan B, we  recruit  short-term clients at each global aggregation who have not been selected in Plan A with the aim to ensure a seamless training process through obtaining the solution for $\bm{\mathcal P}^{(g)}_2$ given by
\begin{equation}
	\bm{\mathcal P}^{(g)}_2:~~\underset{\widetilde{\boldsymbol{A}}^{(g)}}{\min}~~ \lambda_{t} {\widetilde{\mathbb{T}}}^{(g)}+\lambda_{e}\widetilde{\mathbb{E}}^{(g)}
\end{equation}
\vspace{-2mm} 
\setcounter{equation}{11}
\begin{subequations}\label{p2}{	
		\begin{align}
			\text{s.t.}~~~~&
			\operatorname{KLD}\left(\widetilde{\boldsymbol{P}}_j^{(g)}||\boldsymbol{Q}\right)\le {\it{KLD}}_{\textrm {max}}, \forall s_{j}\in\mathcal{S}\label{p2_12a}\\[-.1em]
			&\widetilde{D}_j^{(g)}\ge D_{\textrm {min}}, \forall s_{j}\in\mathcal{S}\label{p2_12b}\\[-.1em]
			&\widetilde{a}_{i,j}^{(g)}\in\left\{0,1\right\},\forall \bm{n}_{i}\in\mathcal{N},\forall s_{j}\in\mathcal{S}\label{p2_12c}\\[-.1em]
			&\widetilde{a}_{i,j}^{(g)}=1,\text{if}\,\, \widehat{a}_{i,j}^{(g)}\xi_{i}^{(g)}=1,\forall \bm{n}_{i}\in\mathcal{N},\forall s_{j}\in\mathcal{S}\label{p2_12d}\\[-.1em]
			&\sum_{j=1}^{|\mathcal{S}|}\widetilde{a}_{i,j}^{(g)}\le 1,\forall \bm{n}_{i}\in\mathcal{N}\label{p2_12e}\\[-.1em]
			&\sum_{i=1}^{|\mathcal{N}|}\widetilde{a}_{i,j}^{(g)}\le M_j,\forall s_{j}\in\mathcal{S}\label{p2_12f}
	\end{align}}
\end{subequations}

Note that $\widetilde{a}_{i,j}^{(g)}$ denotes a binary indicator as shown by constraint \eqref{p2_12c}, describing the selection of  proper online clients that are not long-term ones in Plan A. In particular, we use a set $\widetilde{\bm{A}}^{(g)}$ to collect $\widetilde{a}_{i,j}^{(g)}$, distinguishing them from the previous discussed $\bm{A}^{(g)}$ and $\widehat{\bm{A}}^{(g)}$. More importantly, when $\widehat{a}_{i,j}^{(g)}\xi_{i}^{(g)}=1$, we have $\widetilde{a}_{i,j}^{(g)}=1$ to involve all the online long-term clients into the training, as depicted by constraint \eqref{p2_12d}. Similar to $\widehat{\boldsymbol{P}}^{(g)}_j$ and $\widehat{D}_j^{(g)}$, $\widetilde{\boldsymbol{P}}^{(g)}_j$ and $\widetilde{D}_j^{(g)}$ are the results after replacing $a^{(g)}_{i,j}$ in $\boldsymbol{P}^{(g)}_j$ and $D^{(g)}_j$ with $\widetilde{a}_{i,j}^{(g)}$, respectively. More importantly, in Plan B, during the actual training process, the constraints on KLD and data size are treated as standard inequality constraints (i.e., \eqref{p2_12a} and \eqref{p2_12b}), rather than in Plan A (they are formulated probabilistically with risk-control measures). This adjustment reflects the actual client participation status and facilitates more deterministic and reliable decisions for client selection and C2E association. Together, these complementary plans accelerate convergence by balancing the uncertainty managed in Plan A with the in-situ optimization provided by Plan B.

\begin{algorithm}[h]
		\footnotesize
	\caption{ Cluster-based Client Update (CCU)}\label{algorithm4}
	\SetKwInOut{Input}{Input}\SetKwInOut{Output}{Output}
	\Input{the clients set $\mathcal{N}_j$ covered by each ES $s_j$; the C2E association result of Plan A $\widehat{\boldsymbol{A}}^{(g)}=\left\{\widehat{a}^{(g)}_{i, j} \mid \bm{n}_{i} \in \mathcal{N}, s_{j} \in \mathcal{S}\right\}$; the clients' participation indicator $\bm{\xi}^{(g)}=\left\{\xi_1^{(g)}, \ldots, \xi_{|\mathcal{N}|}^{(g)}\right\}$}
	\Output{the C2E association result $\widetilde{\boldsymbol{A}}^{(g)}=\left\{\widetilde{a}^{(g)}_{i, j} \mid n_{i} \in \mathcal{N}, s_{j} \in \mathcal{S}\right\}$; the minimum value of $\widetilde{F}^{(g)}_{\textrm {min}}$ of $\lambda_{t} \widetilde{\mathbb{T}}^{(g)}+\lambda_{e} \widetilde{\mathbb{E}}^{(g)}$ }
	
	Obtain the similarity matrix $\bm{\Psi}^{(g)}=\left\{{\psi}^{(g)}_{i, i^{\prime}}\mid \bm{n}_{i}, \bm{n}_{i^\prime} \in \boldsymbol{\mathcal{N}}\right\}$ via ${\psi}^{(g)}_{i, i^{\prime}}={\boldsymbol{u}^{(g)}_{i} \cdot \boldsymbol{u}^{(g)}_{i^{\prime}}}\big/\big({\big\|\boldsymbol{u}^{(g)}_{i}\big\| \times\big\|\boldsymbol{u}^{(g)}_{i^{\prime}}\big\|}\big)$;
	
	Use DBSCAN algorithm\cite{ref38} to cluster $\mathcal{N}_j$ into clusters $\mathcal{C}_j$;
	
	\For {$n_i \in \mathcal{N}$}{
		$\widetilde{a}_{i,j}^{(g)} \leftarrow 1,\text{if}\,\, \widehat{a}^{(g)}_{i,j}=1, \forall s_{j}\in\mathcal{S}$
	}
	\If {\textnormal{constraint \eqref{p2_12a} and \eqref{p2_12b} are not satisfied}}{
		Attempt to replace the selected offline clients with the online ones in the same cluster until constraint \eqref{p2_12a} and \eqref{p2_12b} are satisfied;
		
		\If {\textnormal{someone belonging to the noise dropouts or the backup cannot be found to meet the constraints}}{
			Apply Algs. \ref{algorithm2} and \ref{algorithm3} to these clients for iterative optimization;
		}
	} 
	Calculate $\widetilde{F}_{\textrm {min}}^{(g)}$ via \eqref{Time}, \eqref{Energy} and \eqref{p2};
	
	{\bf{retrun}} $\widetilde{\boldsymbol{A}}^{(g)}$, $\widetilde{F}^{(g)}_{\textrm {min}}$;
\end{algorithm}

To tackle $\bm{\mathcal{P}}^{(g)}_2$, we introduce a \underline{c}luster-based \underline{c}lient \underline{u}pdate (CCU) algorithm, referred to as Plan B, for timely on-site decision-making. This method is outlined in Alg. \ref{algorithm4}. In a nutshell, the method starts with obtaining the similarity evaluation function {\footnotesize ${\psi}^{(g)}_{i, i^{\prime}}=\frac{\boldsymbol{u}^{(g)}_{i} \cdot \boldsymbol{u}^{(g)}_{i^{\prime}}}{\left\|\boldsymbol{u}^{(g)}_{i}\right\| \times\left\|\boldsymbol{u}^{(g)}_{i^{\prime}}\right\|}$} to determine the similarity matrix $\bm{\Psi}^{(g)}=\left\{{\psi}^{(g)}_{i, i^{\prime}}\mid \bm{n}_{i}, \bm{n}_{i^\prime} \in \mathcal{N}\right\}$, where ${\psi}^{(g)}_{i, i^{\prime}}$ is the cosine similarity between $\boldsymbol{u}^{(g)}_{i}$ and $\boldsymbol{u}^{(g)}_{i^{\prime}}$, and $\boldsymbol{u}^{(g)}_{i}=\left[d_{i}, T_{i, j}^{(g)}, E_{i, j}^{(g)}\right]$ is the vector characterized by data size, delay and energy consumption of client $\bm{n}_i$ when associated with ES $s_j$. 
Given a similarity threshold ${\psi}_{\textrm {min}}$ of a \emph{neighborhood} (clients $\bm{n}_{i^{\prime}}$ is considered as one of the \emph{neighbors} within the \emph{neighborhood} of client $\bm{n}_{i}$ if ${\psi}^{(g)}_{i, i^{\prime}} \ge {\psi}_{\textrm {min}}$) and a density threshold $P_{\textrm {min}}$ for the \emph{neighborhood} (there are at least $P_{\textrm {min}}$ points/clients in the \emph{neighborhood} of a \textit{core point}\footnote{For more details about DBSCAN algorithm, please refer to \cite{ref38}.}), we use the density-based spatial clustering of applications with noise (DBSCAN)\cite{ref38} to cluster $\mathcal{N}_j$ into clusters $\mathcal{C}_j$. While DBSCAN-based clustering is generally effective, it may occasionally fail to form a valid cluster if all candidate clients under an ES significantly differ from the offline long-term client in terms of data characteristics, latency, or energy consumption.

As shown in \eqref{p2_12d}, we copy $\widehat{\boldsymbol{A}}^{(g)}$ derived from Plan A to $\widetilde{\boldsymbol{A}}^{(g)}$ (lines 3-4). According to the similarity matrix $\bm{\Psi}^{(g)}$, the clients in the neighborhood of each offline long-term client are sorted in an increasing order of similarity ${\psi}^{(g)}_{i, i^{\prime}}$. For each ES, it first selects the same number of offline long-term clients according to  $\bm{\Psi}^{(g)}$ in the corresponding cluster (line 6). Should the constraints remain unmet or if any participant falling into the category of \textit{noise point} dropouts thus no valid cluster can be formed, a fallback mechanism is triggered. Specifically, we first attempt to randomly select an equivalent number of clients as backups to replace the offline long-term clients for each ES. Following this, Algs. \ref{algorithm2} and \ref{algorithm3} are applied to these backup clients for local iterative optimization (lines 7-8 of Alg. \ref{algorithm4})\footnote{In extreme scenarios where a large number of long-term clients dropout under the same ES, even the fallback may fail to fully satisfy data constraints. Possible remedies include discarding the ES result, reweighting its contribution in global aggregation, or temporarily reassigning its long-term clients to other ESs. We leave these strategies as potential extensions for future work.}.

\section{Evaluations}
\label{sec:evaluations}
This section presents experiments on both real-world data, EUA dataset\footnote{https://github.com/swinedge/eua-dataset} (see Sec. \ref{subsec:real-world experiments}), and numerical simulation data  (outlined in Sec. \ref{subsec:numerical experiments}) to evaluate our proposed StagewiseHFL framework. Simulations were conducted using MATLAB R2021a and Python 3.9 with PyTorch 1.13 on a workstation equipped with an Intel Core i9-13900K @ 3.00 GHz processor, 32 GB memory, and an NVIDIA GeForce RTX 4080 GPU. More details can be found in Appx. \ref{appx:testbed}. We focus on three metrics: \emph{\romannumeral1)} model test accuracy, \emph{\romannumeral2)} the overall cost of reaching a target learning accuracy, calculated by the cost function $\sum_g{F^{(g)}}=\sum_g(\lambda_{t}\mathbb{T}^{(g)}+\lambda_{e}\mathbb{E}^{(g)})$ (the sum of the objective of $\bm{\mathcal{P}}^{(g)}_0$), and \emph{\romannumeral3)} running time, reflecting the overhead caused by decision-making, i.e., the time spent on making decisions of C2E association and client selection. The considered benchmarks inspired by \cite{ref5, ref10} are detailed below.

\noindent $\bullet$ \textit{OrigProbSolver}: This method directly solves the original problem $\bm{\mathcal{P}}^{(g)}_0$ by iteratively using Algs. \ref{algorithm2} and \ref{algorithm3}, before the start of each global iteration.

\noindent $\bullet$ \textit{KLDMinimization}: This method minimizes the averaged KLD of data between ESs via Algs. \ref{algorithm2} and \ref{algorithm3}, before each global iteration, i.e., to reach the most approximately edge-IID data.

\noindent $\bullet$ \textit{ClientSelOnly}: This method iteratively optimizes the selected clients to solve $\bm{\mathcal{P}}^{(g)}_0$ via Alg. \ref{algorithm3}, while associating them randomly to ESs, before the start of each global iteration.

\noindent $\bullet$ \textit{C2EAssocOnly}: This method randomly selects a set of clients \cite{ref2} and optimizes C2E associations to solve $\bm{\mathcal{P}}^{(g)}_0$ via Alg. \ref{algorithm2}, before the start of each global iteration. 

\noindent $\bullet$ \textit{C2EGreedyAssoc}: This method iteratively optimizes the selected clients to solve $\bm{\mathcal{P}}^{(g)}_0$ via Alg. \ref{algorithm3}, while greedily associate each client $\bm{n}_i$ to an ES $s_j$  with the minimum transmission latency $t_{i,j}^{com}$, before each global iteration.

\noindent $\bullet$ \textit{FedCS} \cite{ref56}: This method selects as many clients as possible for each ES via Alg. \ref{algorithm3}, before the start of each global iteration.

Note that, methods ClientSelOnly, C2EAssocOnly, C2EGreedyAssoc, and FedCS, which focus solely on optimizing either the C2E association or the client selection problem, face challenges in simultaneously meeting the constraints related to KLD and data size. Thus, we exclude the KLD as a constraint for these benchmark methods in our evaluation.
\subsection{Real-world EUA Dataset-driven Experiments}
\label{subsec:real-world experiments}

We first use the real-world EUA dataset, which has been widely employed in edge computing and FL environments. This dataset includes geographic locations of 1,464 base stations (BSs) and 174,305 end-users within the Melbourne metropolitan area, Australia. We selected 93 users as clients and 4 BSs as ESs with a 500m $\times$ 500m region. The dataset provides latitude and longitude of these entities, enabling us to calculate the distances between clients and ESs accordingly.

We evaluate on five datasets and models: \emph{\romannumeral1)} MNIST \cite{ref39} with CNN, \emph{\romannumeral2)} Fashion-MNIST \cite{ref41} with LeNet5 \cite{ref39}, \emph{\romannumeral3)} CIFAR-10 \cite{ref40} with CNN, \emph{\romannumeral4)} CIFAR-10 with ResNet18 \cite{ref42}, and \emph{\romannumeral5)} ALI \cite{ref65, ref66} with VGG \cite{ref67}. Due to space limitation, we report details for the first four in the main document, with extended analysis and results in Appx. \ref{appx:dataset and model} and Appx. \ref{appx:evaluation V}. While existing methodologies fields such as computer vision have attained high test accuracies for these datasets\cite{ref57,ref58,ref59,ref60}, our primary concern in this paper is to assess the performance of various methods in terms of time and energy efficiency against a predetermined target accuracy, rather than striving for the highest possible accuracy. This also aligns with the rationale of several HFL studies such as \cite{ref10,ref11,ref24}.

\begin{table}[H]
	\footnotesize
	\caption{	
		 Simulation setting \cite{ref5,ref24}}
         \label{tab:simulation setting}
	\centering
	\begin{tabular}{|>{\centering\arraybackslash}m{5.4cm}|@{\hskip 3pt}|c|}
		\hline
		\rowcolor{verylightgray}
		\bf{Parameter} & \bf{Value}\\
		\hline
		\hline
		Online probability, $p_i$ & [0.5,1)\\
		
		\hline
		\rowcolor{verylightgray}
		Capacitance coefficient, $\alpha_i$ & $10^{-28}$\\
		\hline
		Number of CPU cycles required for processing one sample data, $c_i$ & [30,100] cycles/bit\\
		
		\hline
		\rowcolor{verylightgray}
		Clients' CPU frequency, $f_i$ & [1,10] GHz\\
		\hline
		Allocated bandwidth for clients, $B_j$ & 1 MHz\\
		
		\hline
		\rowcolor{verylightgray}
		Maximum number of clients that can access to ESs, $M_j$ & [8,12]\\
		\hline
		Clients' transmit power, $q_i$ & [200,800] mW\\
		
		\hline
		\rowcolor{verylightgray}
		Noise power spectral density, $N_0$ & -174 dBm/Hz\\
		\hline
		Transmission delay for ESs uploading the edge model to the CS, $T_j^{\it com}$ & [160,200] ms\\
		
		\hline
	\end{tabular}
\end{table}
To emulate non-IID data, we allocate datapoints from only 1 to 3 labels (out of 10) to each client for MNIST, Fashion-MNIST and CIFAR-10, while 10 to 30 labels (out of 100) for CIFAR-100. Each client holds a data quantity within the range of [255, 1013] for MNIST and Fashion-MNIST, and [206, 863] for CIFAR-10 and CIFAR-100. The participation of client $n_i$ follows a Bernoulli distribution $\mathcal{B}\left(1,p_i\right)$, where $p_i$ is randomly chosen from the interval [0.5,1). The minimum tolerable data size $D_{\textrm {min}}$ in OrigProbSolver and StagewiseHFL is set as 2500, while  ${\it{KLD}}_{\textrm {max}}$ is set by 0.2. Other parameters are summarized in Table \ref{tab:simulation setting}.

\subsubsection{Evaluation of  test accuracy}
\label{subsubsec:evaluation on acc}

\begin{figure}[htbp]

	\centering
	
	\subfigure[MNIST]{
		\label{subfig:acc on mnist}
        \includegraphics[trim=0cm 0cm 1.0cm 0.5cm, clip, width=0.47\columnwidth]{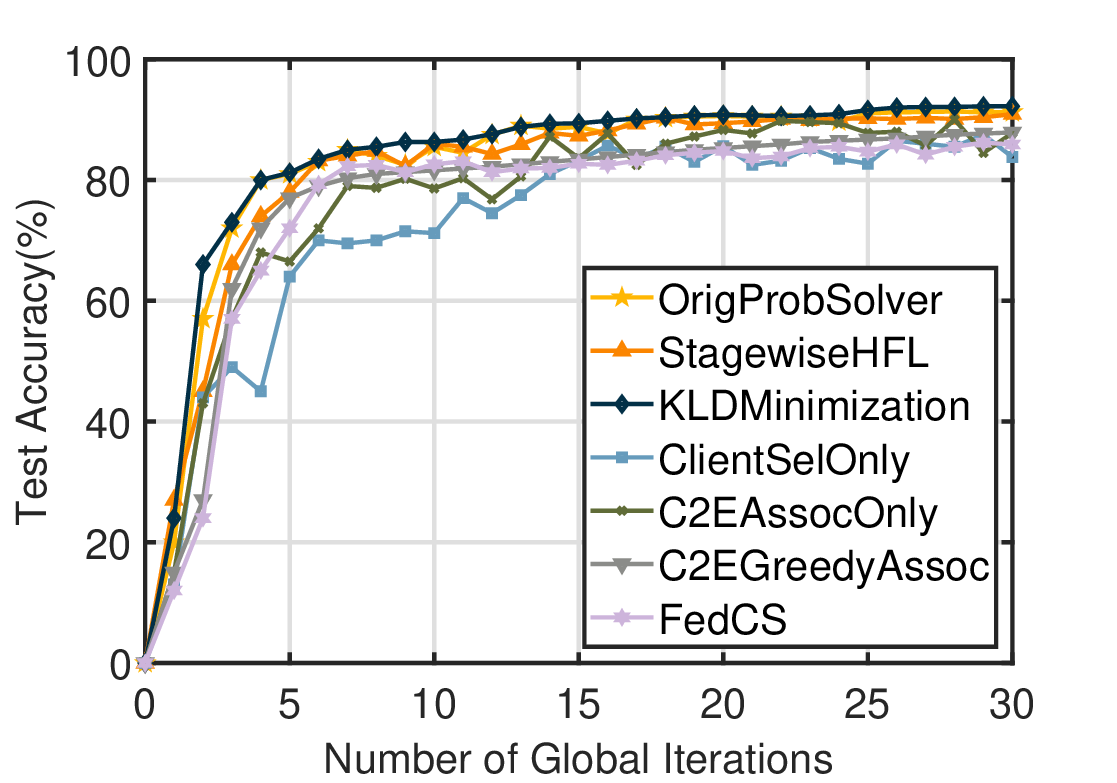}
	}
	\hspace{-10pt}
	\subfigure[Fashion-MNIST]{
        \label{subfig:acc on fmnist}
		\includegraphics[trim=0cm 0cm 1.0cm 0.5cm, clip, width=0.47\columnwidth]{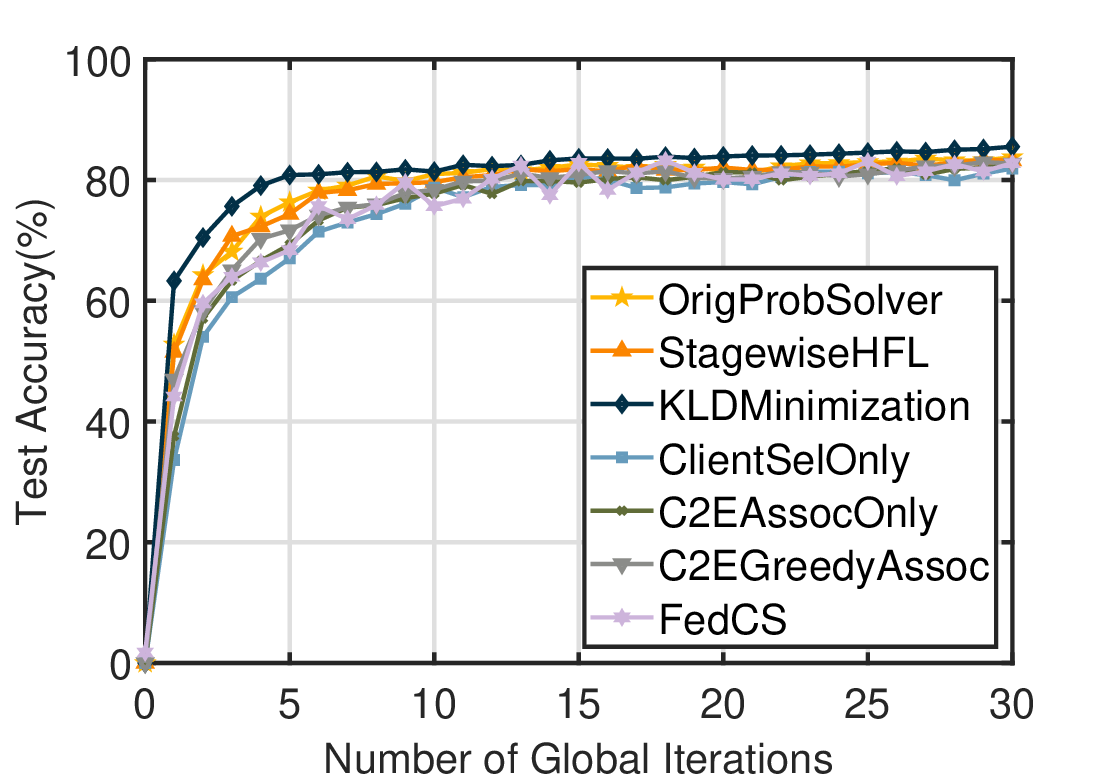}
	}
	
	\subfigure[CIFAR-10]{
		\label{subfig:acc on cifar-10}
        \includegraphics[trim=0cm 0cm 1.0cm 0.5cm, clip, width=0.47\columnwidth]{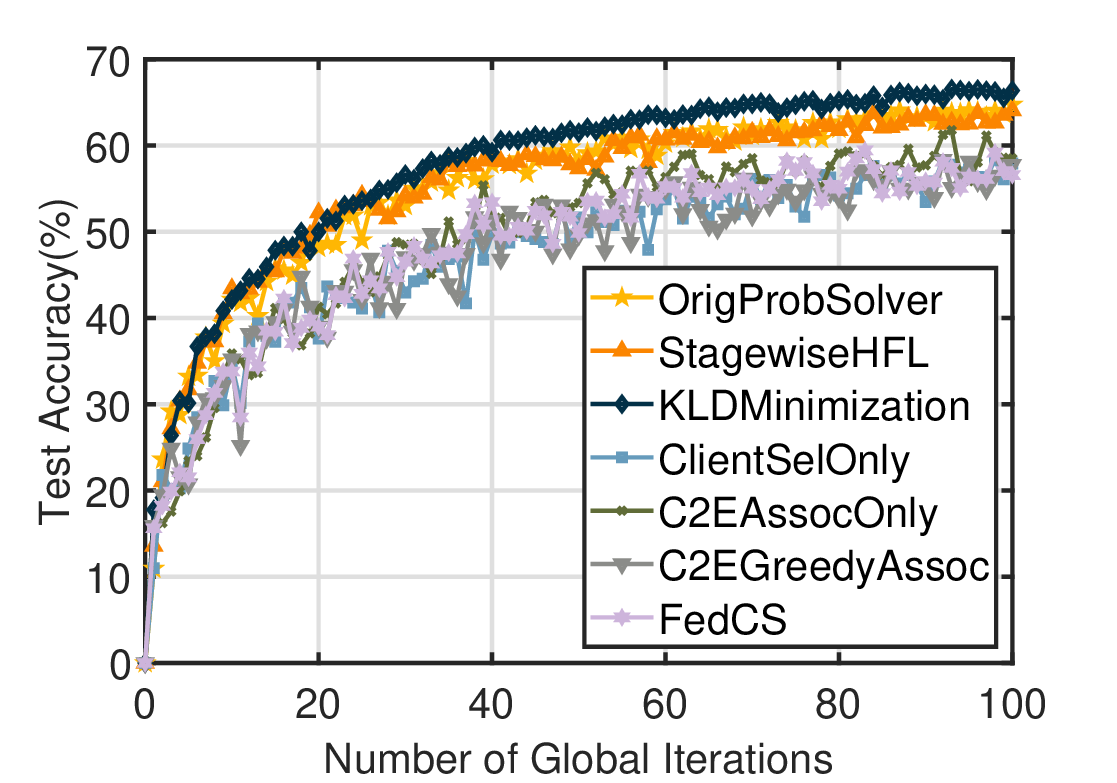}
	}
	\subfigure[CIFAR-100]{
        \label{subfig:acc on cifar-100}
        \includegraphics[trim=0cm 0cm 1.0cm 0.5cm, clip, width=0.47\columnwidth]{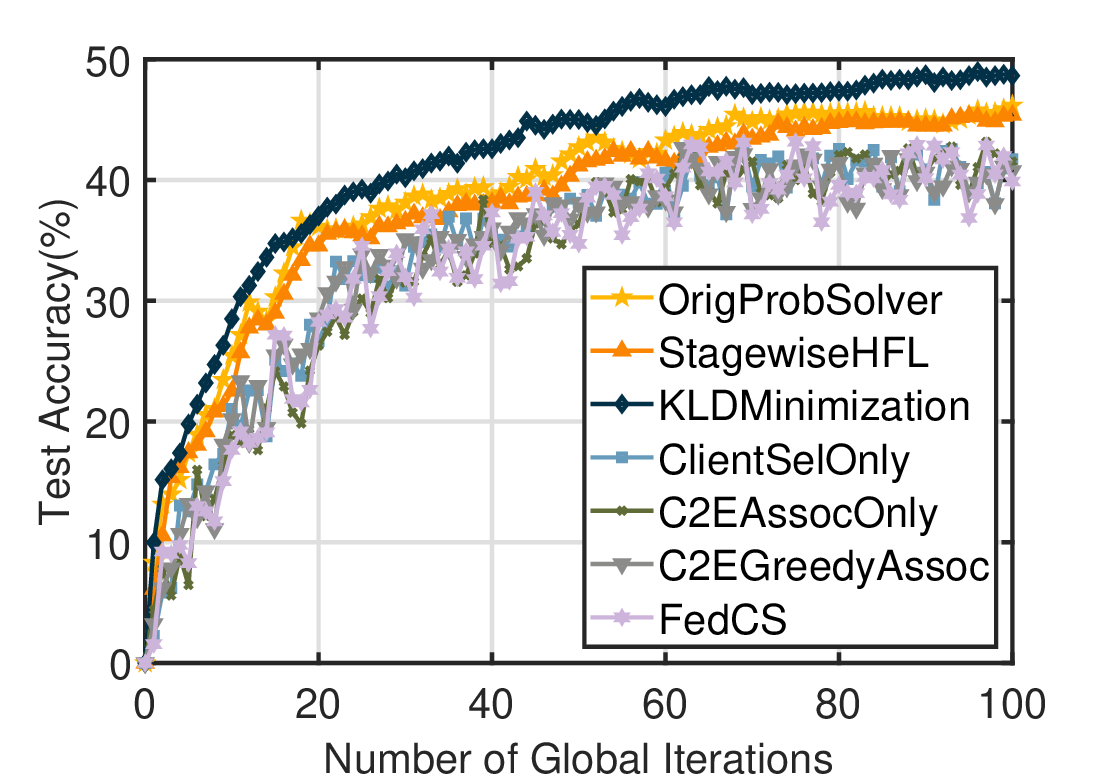}
	}
	\caption{Performance comparison in terms of test accuracy.}
    \label{fig:acc on 4}
\end{figure}



Fig. \ref{fig:acc on 4} shows the performance in terms of test  accuracy on MNIST, Fashion-MNIST, CIFAR-10 and CIFAR-100, where $\mathcal{T}$ and $\mathcal{L}$ are set to 5 and 3, respectively\cite{ref10,ref49}. As shown in Fig. \ref{subfig:acc on mnist}, our  StagewiseHFL outperforms ClientSelOnly, C2EAssocOnly, C2EGreedyAssoc and FedCS by achieving higher test accuracy by 3.12\%, 7.11\%, 3.03\%, and 5.03\%, respectively, within 30 global iterations on MNIST. Comparing to OrigProbSolver, our StagewiseHFL obtains a slightly lower test accuracy (around 0.34\% lower). This is because our proposed Plan A relies on historical information, where the gap between the estimation of historical records and the actual network condition (i.e., actual attendance of clients) can lead to performance degradation. We will later show that this slight performance edge of OrigProbSolver comes with the drawback of significant running time, rendering it inapplicable for large scale systems (later depicted in Fig. \ref{fig:performace on running time}). Similarly, our StagewiseHFL exhibits 1.31\% lower test accuracy than KLDMinimization method. This is raised from that  KLDMinimization solely focuses achieving data homogeneity (i.e., IID data) across the ESs, while ignoring the cost (i.e., latency and energy consumption) induced by such C2E assignment. As a result, as later shown, KLDMinimization suffers from a prohibitively high cost  $F$ and running time. Notably, KLDMinimization, which directly reduces the average KLD across ESs, achieves higher accuracy and faster convergence than competing baselines, while OrigProbSolver and StagewiseHFL, both incorporating KLD constraints into their optimization, can consistently outperform other methods that lack explicit heterogeneity control. Together, the superior performance of KLDMinimization, OrigProbSolver, and StagewiseHFL underscores a key insight: as the distribution of data across ESs becomes more uniform, the convergence of the model can be greatly improved as reflected in the reduced number of global rounds to reach the target accuracy.

We can observe a similar trend, where our proposed StagewiseHFL demonstrates outstanding performance in terms of model accuracy and convergence speed, as shown in Figs. \ref{subfig:acc on fmnist}-\ref{subfig:acc on cifar-100}. For example, considering CIFAR-10, our StagewiseHFL outperforms ClientSelOnly, C2EAssocOnly, C2EGreedyAssoc and FedCS on CIFAR-10 by 6.12\%, 5.52\%, 6.25\% and 7.47\% in test accuracy, while being only 0.71\% lower than OrigProbSolver and 2.33\% lower than KLDMinimization after 100  global iterations, as shown in  Fig. \ref{subfig:acc on cifar-10}. We next show that our method, while showing slightly lower prediction performance than the best benchmark, exhibits a notable performance gap when other metrics are considered. 

\subsubsection{Evaluation of  delay, energy cost, and running time}
\label{subsubsec:evaluation on delay, energy cost, and running time}
\begin{figure}[htbp]

	\centering
	\includegraphics[trim=1.5cm 0.5cm 3.0cm 0.5cm, clip, width=\columnwidth]{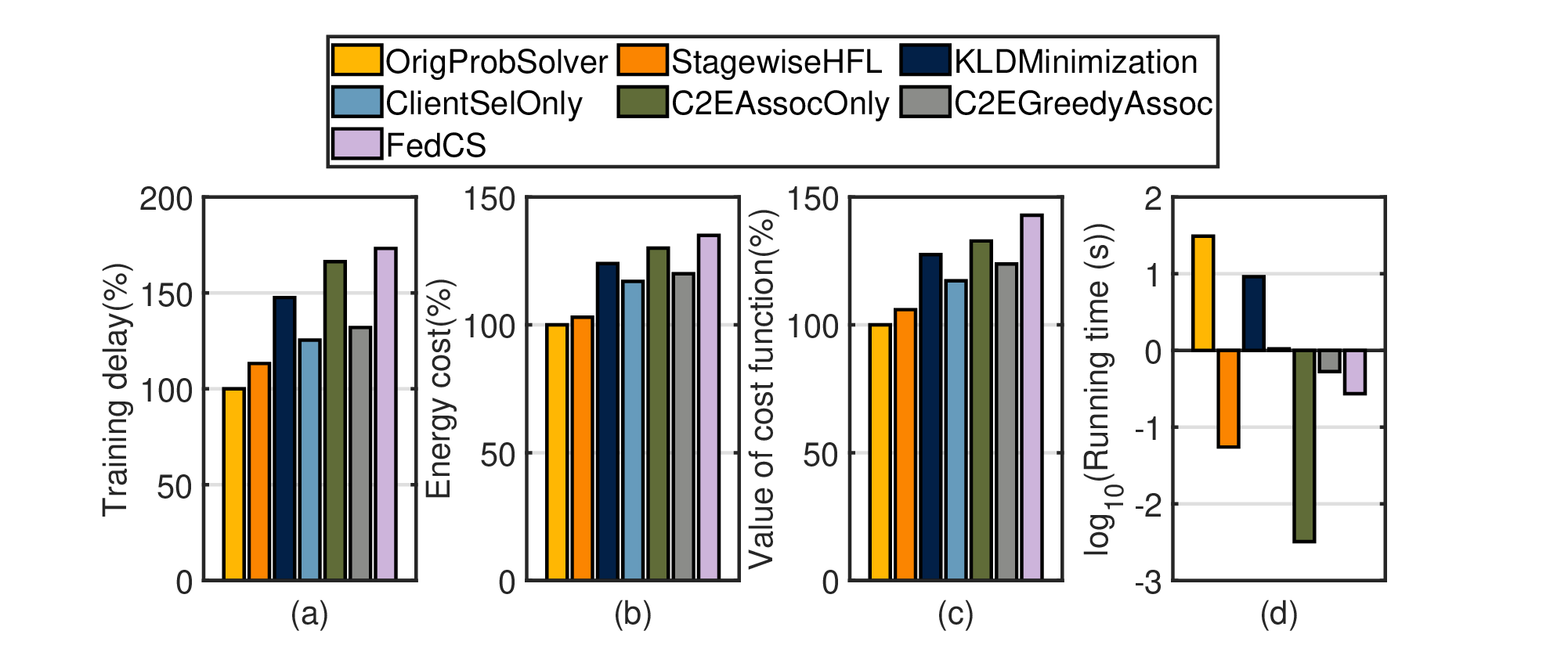}
	\caption{Performance on training delay, energy consumption, overall cost, and running time upon having real-world dataset.}
    \label{fig:delay, energy cost, and running time}
	
\end{figure}


Fig. \ref{fig:delay, energy cost, and running time} illustrates the training delay, energy consumption, the overall cost (i.e.,  $\sum_g{F^{(g)}}$), and the time consumed for inference decision, to reach over 90\% test accuracy on MNIST. We set $\mathcal{T}=5$, $\mathcal{L}=3$, and regard OrigProbSolver as the reference line, to better show the gap among different methods in Figs. \ref{fig:delay, energy cost, and running time}(a)-\ref{fig:delay, energy cost, and running time}(d). In Fig. \ref{fig:delay, energy cost, and running time}(a), the weighting factors $\lambda_t$ (for delay) and $\lambda_e$ (for energy) are assigned values of 1 and 0, respectively. This indicates that we put full emphasis on minimizing the training delay. Comparing to OrigProbSolver which optimizes problem $\bm{\mathcal{P}}^{(g)}_0$ at the beginning of each global iteration, different benchmark methods can introduce various levels of increase on the training delay due to the complications embedded in their designs: 13.2\%, 47.6\%, 25.4\%, 66.3\%, 32.0\% and 73.2\% increase for StagewiseHFL, KLDMinimization, ClientSelOnly, C2EAssocOnly,  C2EGreedyAssoc, FedCS, respectively.  Our StagewiseHFL stands out as the superior choice among the evaluated methods. In Fig. \ref{fig:delay, energy cost, and running time}(b), we showcase the performance metrics related to energy consumption, by setting $\lambda_t=0$ and $\lambda_e=1$. Our StagewiseHFL, in comparison to OrigProbSolver, incurs only a marginal performance loss of 3.0\%. This slight decrease demonstrates StagewiseHFL's superior efficiency, highlighting its capability in balancing energy conservation and performance\footnote{Reported energy values are normalized, as commonly done in FL \cite{ref5, ref56}, to avoid hardware dependence and ensure reproducibility. Since all methods run under identical conditions, these values still offer a fair and meaningful comparison.}. Without loss of generality, when it comes to randomly assigning weighting factors of training delay and energy consumption such that $\lambda_t+\lambda_e=1$ ($\lambda_t$, $\lambda_e\in\left(0,1\right)$) in Fig. \ref{fig:delay, energy cost, and running time}(c), our StagewiseHFL only incurs a 5.9\% loss compared to OrigProbSolver, while achieving performance improvements of up to 21.6\%, 11.3\%, 26.9\%, and 17.9\%, and 36.9\% in comparison with KLDMinimization, ClientSelOnly, C2EAssocOnly, C2EGreedyAssoc, and FedCS, respectively. We next show that the above-discussed close performance of our method to the best baseline comes with a notable gap in terms of running time.

To illustrate the real-time decision-making overhead of different methods, i.e., the time spent on selecting clients and associating them with ESs, Fig. \ref{fig:delay, energy cost, and running time}(d) employs a logarithmic scale to represent the average decision-making time per global training round. Our StagewiseHFL shows the closest performance to OrigProbSolver in terms of cost savings (see Figs. \ref{fig:delay, energy cost, and running time}(a)-\ref{fig:delay, energy cost, and running time}(c)), while offering an average inference time of less than 100ms, which is order of magnitude lower than that of OrigProbSolver. This verifies the effectiveness of our approach in terms of achieving a reasonable value of resource consumption (i.e., energy and delay) under a notably low decision-making overhead.

\subsection{Numerical Data-driven Experiments}
\label{subsec:numerical experiments}
To further evaluate our approach's decision-making overhead and model convergence, we conduct a complementary set of experiments. These experiments aim to investigate the impact of various factors, such as the number of clients (i.e., $|\mathcal{N}|$) and ESs (i.e., $|\mathcal{S}|$), the minimum acceptable data size for model training (i.e., $D_{\textrm {min}}$), and the likelihood of client participation (i.e., $p_i$), on the performance. 

\begin{table}[H]
	\small
	\captionsetup{justification=centering}
	\setlength{\tabcolsep}{0.1pt}
	\begin{center}
		\caption{	
			Parameter setting used in Sec. \ref{subsec:numerical experiments}\\[0.5em]
			* $\mathcal{I}_1$: [0.4, 1), $\mathcal{I}_2$: [0.45, 1), $\mathcal{I}_3$: [0.5, 1), $\mathcal{I}_4$: [0.6, 1), $\mathcal{I}_5$: [0.7, 1)	
			}
		\label{tab:4 sets}
		\begin{tabular}{| c |@{\hskip 3pt}| c | c | c | c |}
			\hline
			\rowcolor{verylightgray}
			&$|\mathcal{N}|$&$|\mathcal{S}|$&$D_{\textrm {min}}$&$p_i$\\
			\hline
			\hline
			Set\#1&75, 80, $\ldots$, 120&4&2500&$\mathcal{I}_3$\\
			\hline
			\rowcolor{verylightgray}
			Set\#2&100&3, 4, $\ldots$,7&2500&$\mathcal{I}_3$\\ 
			\hline
			Set\#3&100&4&2000, 2500, $\ldots$, 4000&$\mathcal{I}_3$\\		
			\hline 
			\rowcolor{verylightgray}
			Set\#4&100&4&2500& $\mathcal{I}_1$, $\mathcal{I}_2$, $\ldots$, $\mathcal{I}_5$\\
			\hline
		\end{tabular}
		
	\end{center}
	\medskip 
	~~~~
\end{table}

\subsubsection{Evaluation of  the overall cost}\label{subsubsec:evaluation on cost}
We first evaluate the induced cost\footnote{The reported cost metrics are virtualization-based, derived from analytical models and empirical measurements \cite{ref23,ref28}, and abstracted from specific hardware, ensuring general applicability of our results.} to achieve a 90\% accuracy on MNIST dataset under various parameters: the number of clients $|\mathcal{N}|$, the number of ESs $|\mathcal{S}|$, the tolerable minimum data size $D_{\textrm {min}}$ for each ES, and the participation probability $p_i$ for each client $\bm{n}_i$. For illustrations, we conduct the next experiments under 4 sets of parameters detailed in Table \ref{tab:4 sets}.

\begin{figure*}[htbp]

	\centering
	
	\subfigure[Cost vs. $|\mathcal{N}|$ (Set\#1)]{
    \label{subfig: cost vs n}
		\includegraphics[trim=1cm 0cm 2cm 0.5cm, clip, width=0.47\columnwidth]{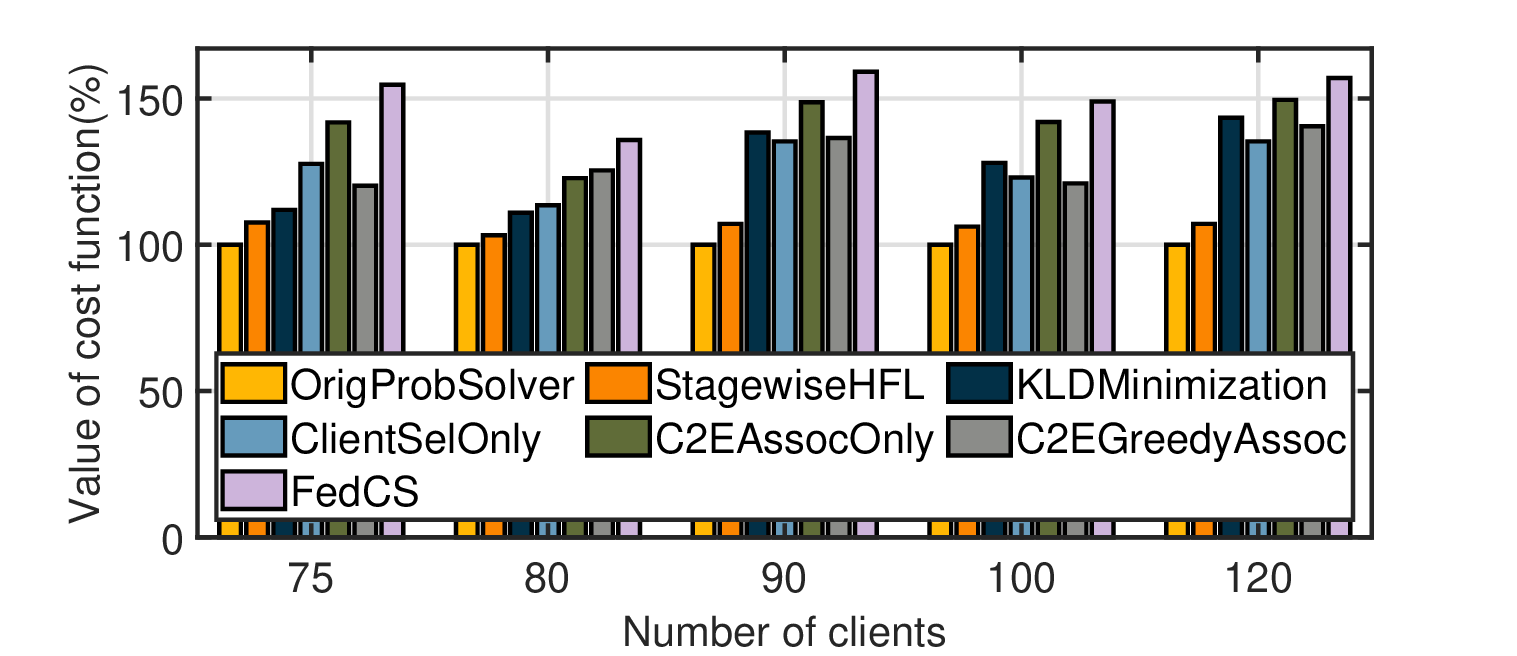}
	}
	\subfigure[Cost vs. $|\mathcal{S}|$ (Set\#2)]{
        \label{subfig: cost vs s}
		\includegraphics[trim=1cm 0cm 2cm 0.5cm, clip, width=0.47\columnwidth]{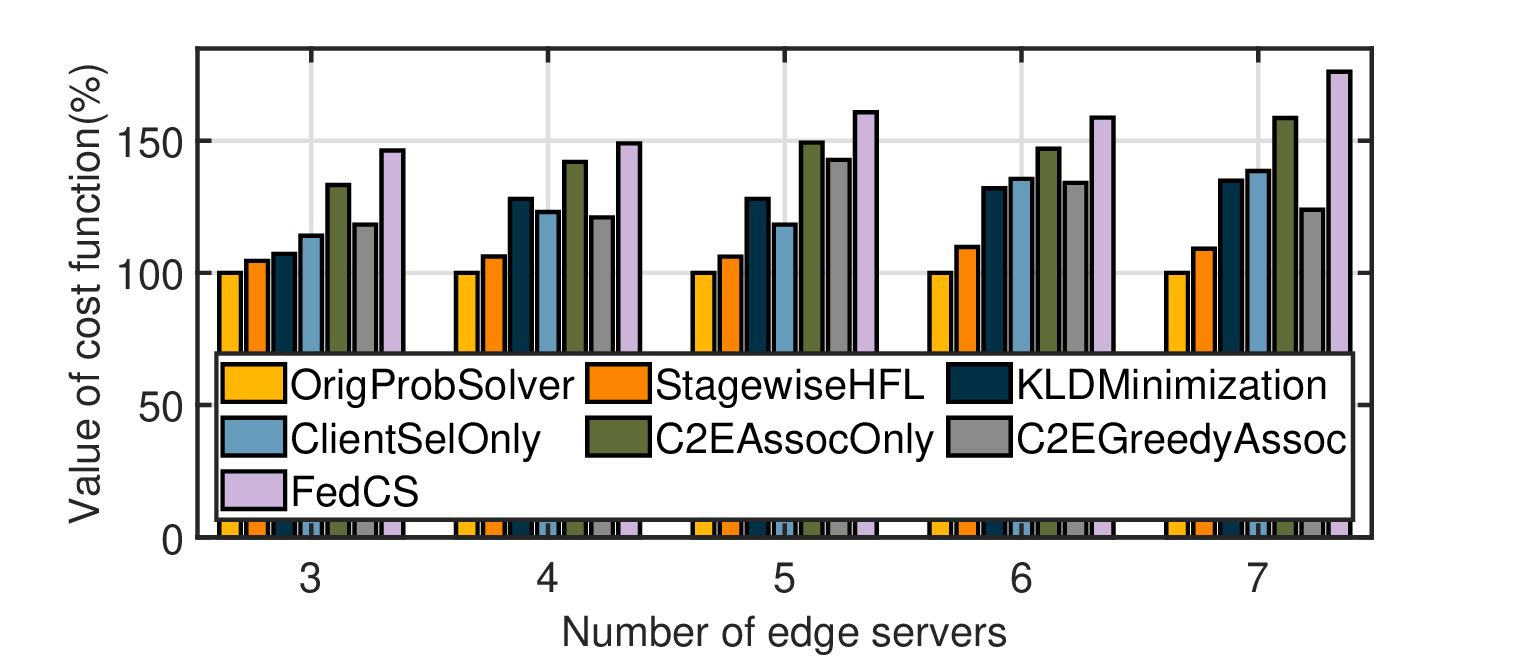}
	}
	\subfigure[Cost vs. $D_{\textrm {min}}$ (Set\#3)]{
    \label{subfig: cost vs dmin}
		\includegraphics[trim=1cm 0cm 2cm 0.5cm, clip, width=0.47\columnwidth]{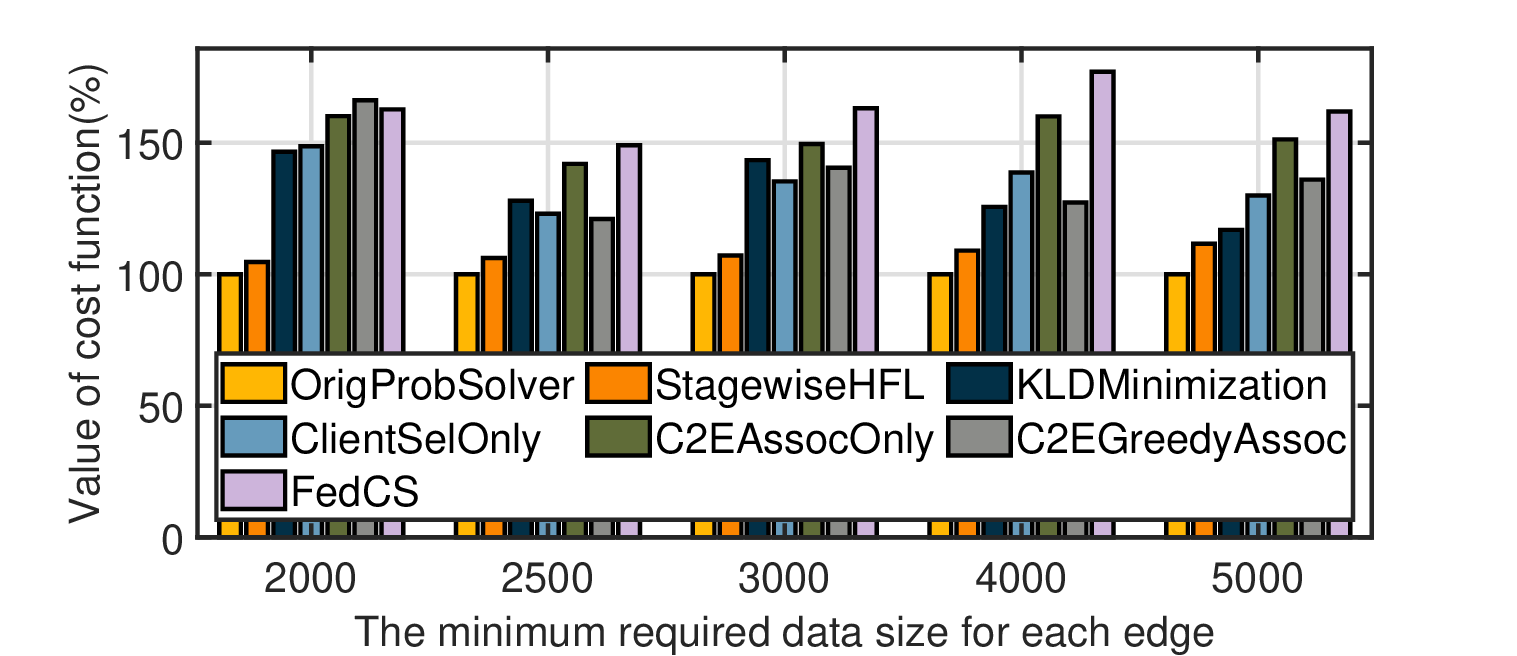}
	}
	\subfigure[Cost vs. $p_i$ (Set\#4)]{
    \label{subfig: cost vs pi}
		\includegraphics[trim=1cm 0cm 2cm 0.5cm, clip, width=0.47\columnwidth]{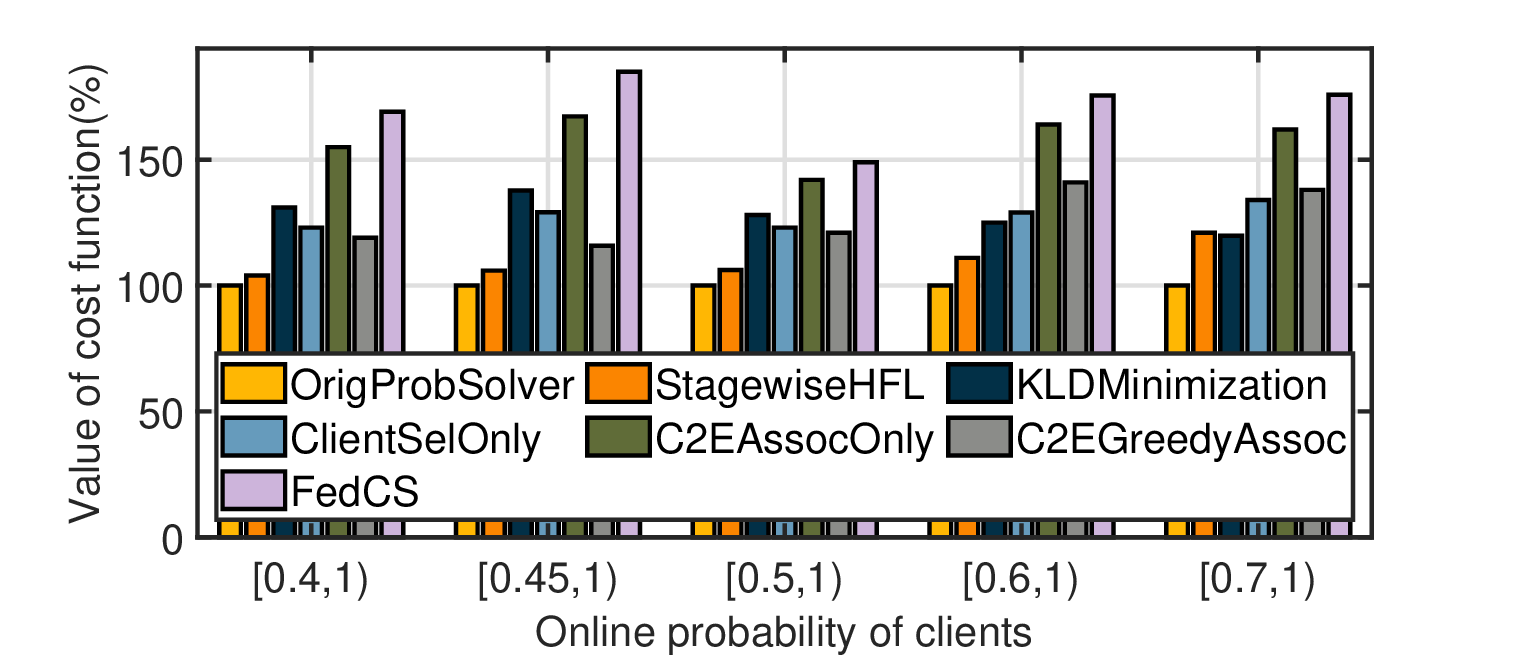}
	}
	\caption{Performance on the overall system cost within numerical dataset.}
    \label{fig:performance on cost}
\end{figure*}

\begin{figure*}[htbp]
	\centering
	\subfigure[Running time vs. $|\mathcal{N}|$ (Set\#1)]{
    \label{subfig:running time vs n}
		\includegraphics[trim=1cm 0cm 2cm 0.5cm, clip, width=0.47\columnwidth]{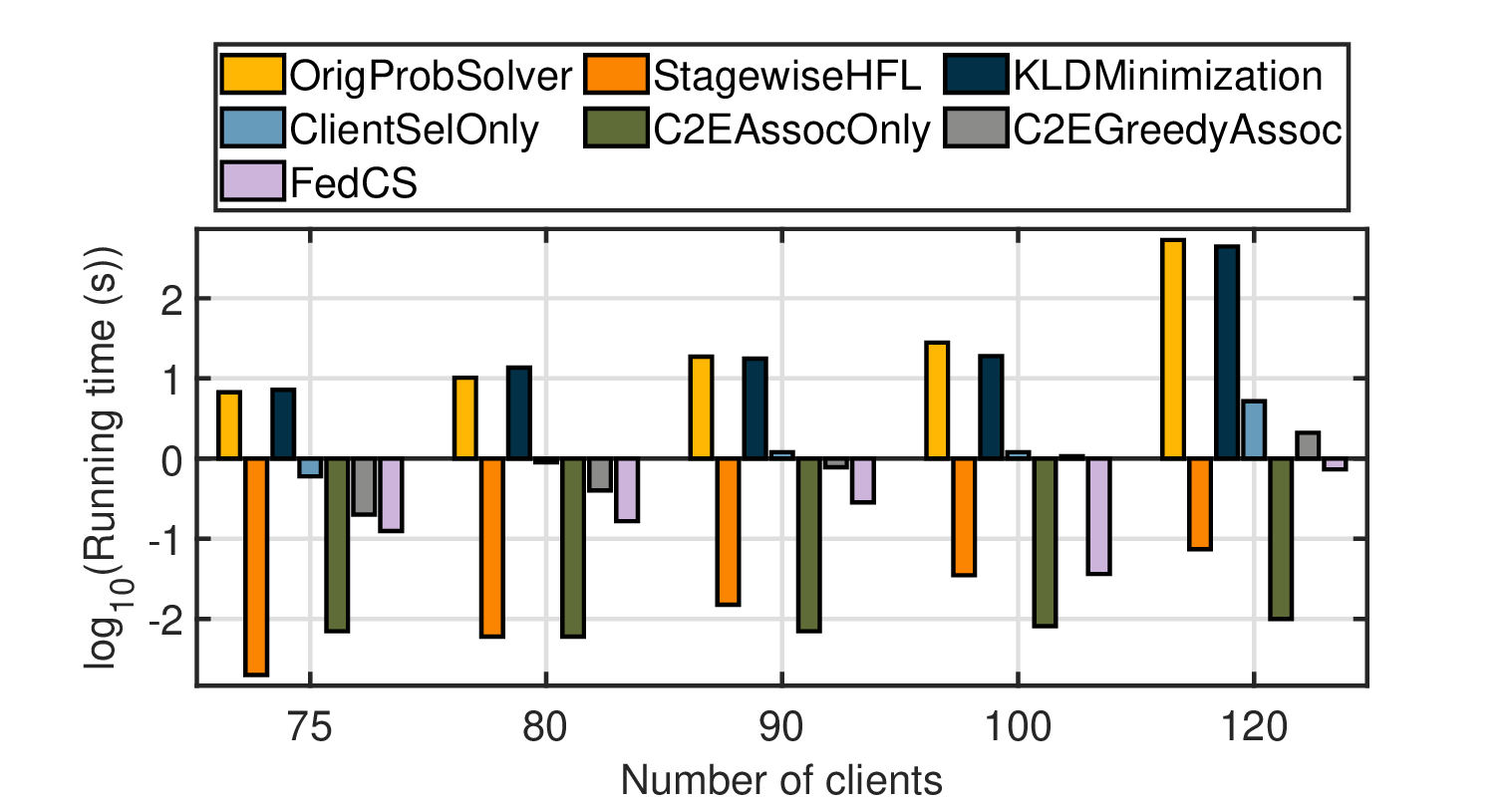}
	}
	\subfigure[Running time vs. $|\mathcal{S}|$ (Set\#2)]{
        \label{subfig:running time vs s}
		\includegraphics[trim=1cm 0cm 2cm 0.5cm, clip, width=0.47\columnwidth]{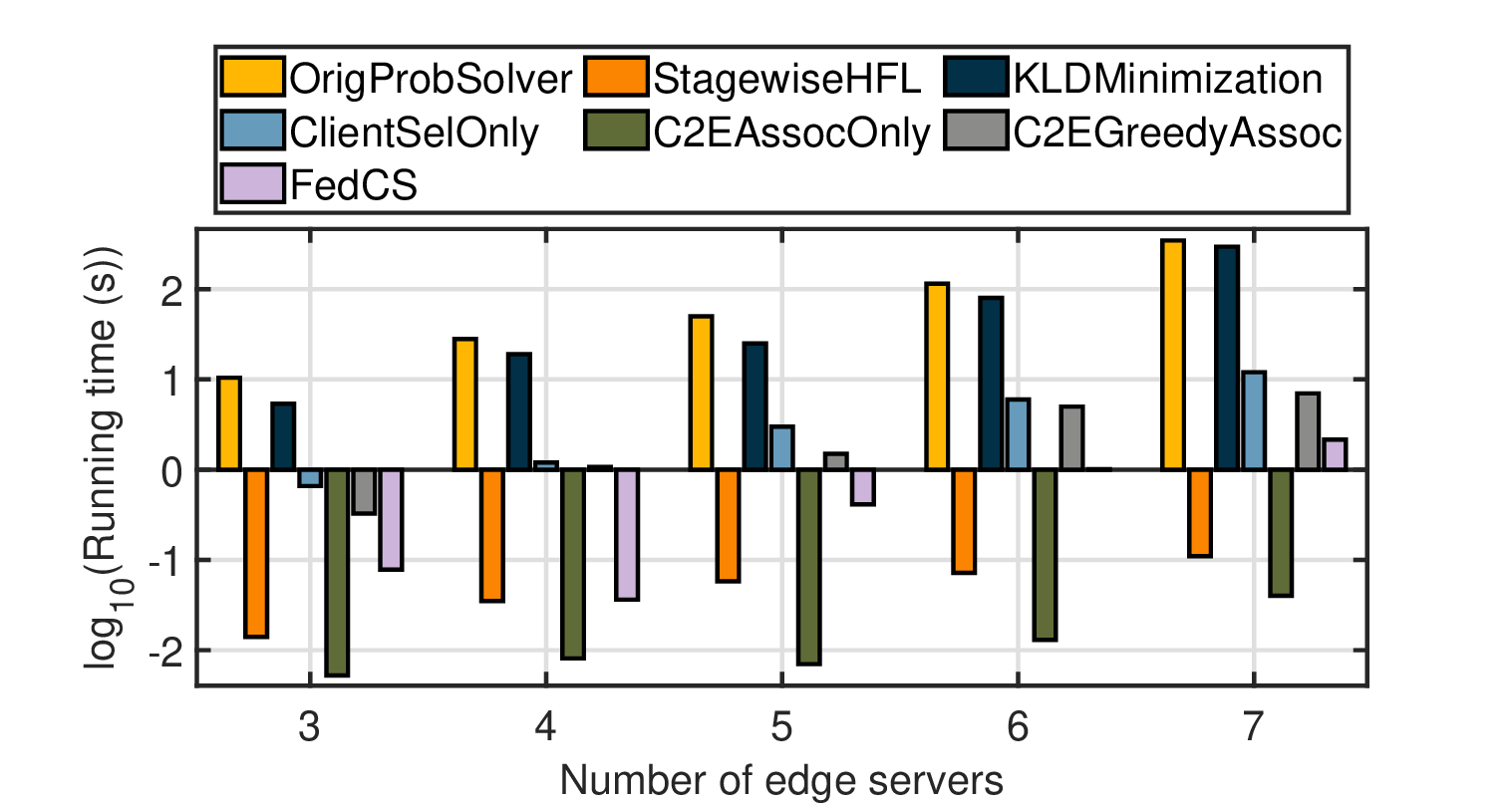}
	}
	\subfigure[Running time vs. $D_{\textrm {min}}$ (Set\#3)]{
    \label{subfig:running time vs dmin}
		\includegraphics[trim=1cm 0cm 2cm 0.5cm, clip, width=0.47\columnwidth]{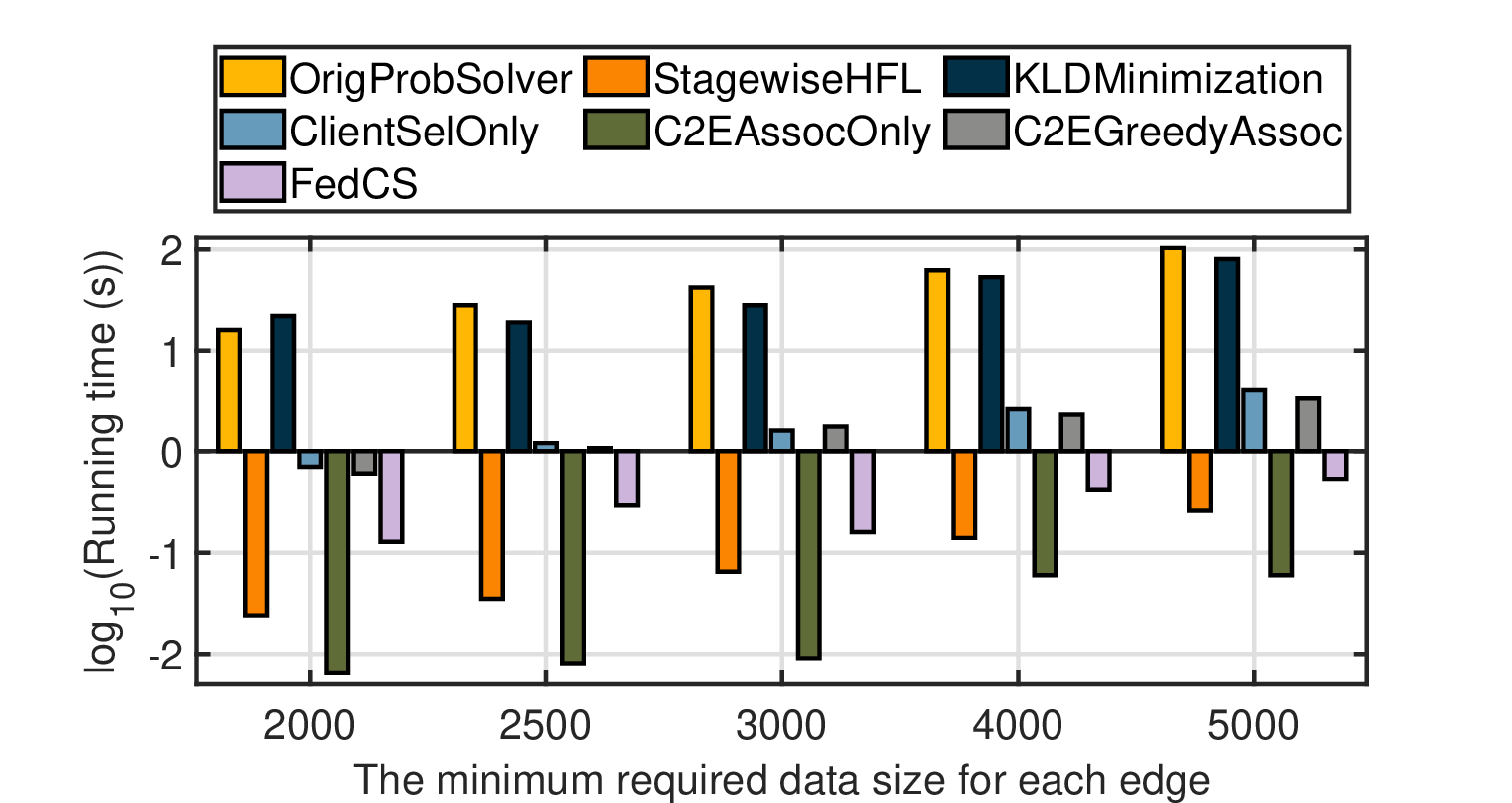}
	}
	\subfigure[Running time vs. $p_i$ (Set\#4)]{
    \label{subfig:running time vs pi}
		\includegraphics[trim=1cm 0cm 2cm 0.5cm, clip, width=0.47\columnwidth]{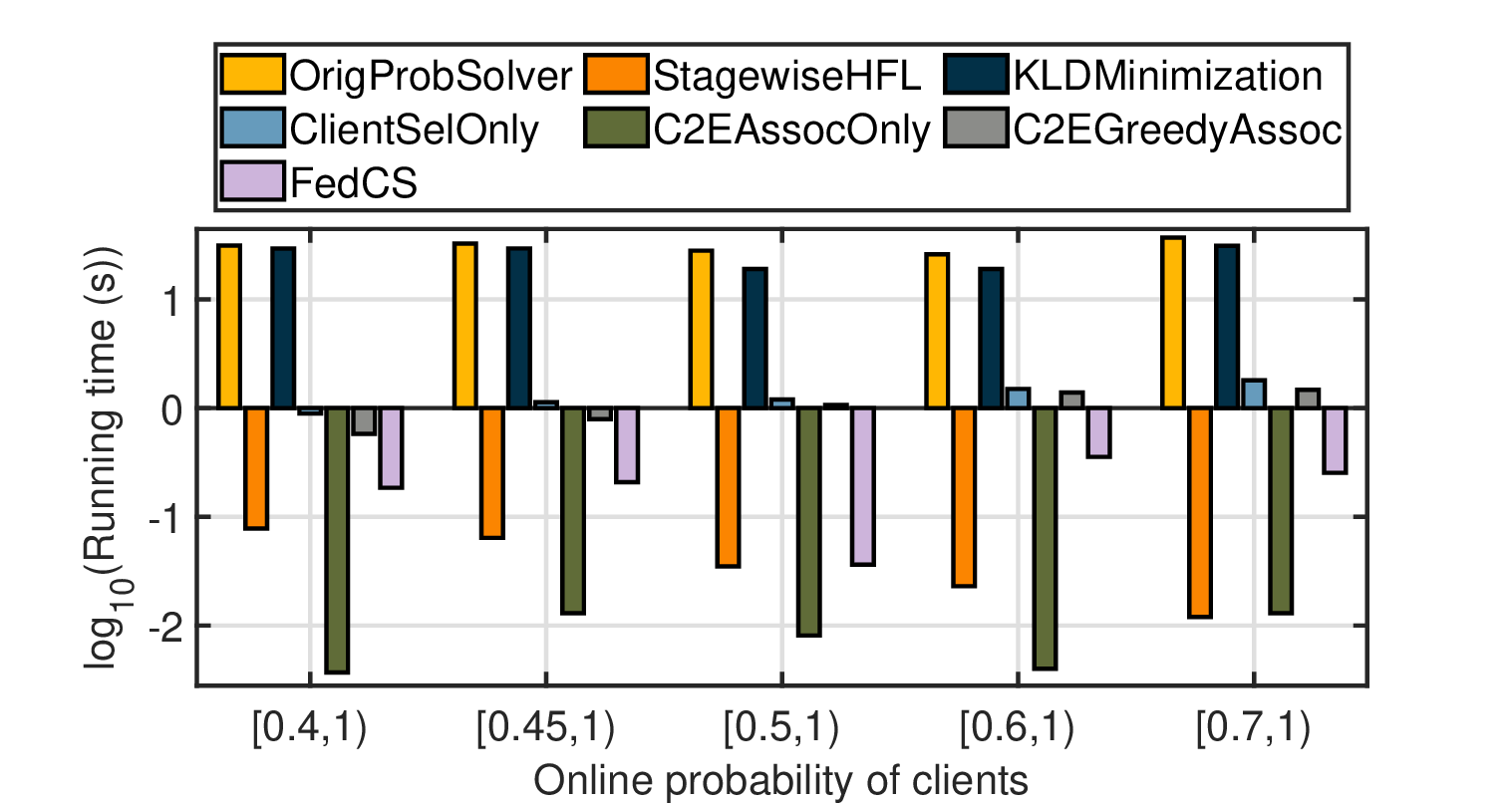}
	}
	\caption{Performance on running time within numerical dataset.}
\label{fig:performace on running time}
\end{figure*}

We present a performance comparison on the value of cost function $F$  in Fig. \ref{subfig: cost vs n}, using OrigProbSolver as the reference line. Taking into account varying numbers of clients (75 to 120), our StagewiseHFL experiences a performance gap of 7.6\%, 3.3\%, 7.1\%, 6.2\%, and 7.2\% as compared to OrigProbSolver. This is because  Plan A in StagewiseHFL relies on historical data, introducing risks of inaccurate estimates in client selection and C2E association (the drawbacks associated with running time of OrigProbSolver will be analyzed in Fig. \ref{fig:performace on running time}). Our StagewiseHFL exhibits cost-effectiveness when $|\mathcal{N}|$ ranges from 75 to 120 compared to KLDMinimization, ClientSelOnly, C2EAssocOnly, C2EGreedyAssoc and FedCS. This is attributed to our optimized decision-making process for both client selection and C2E association. For instance, upon having $|\mathcal{N}|=80$ in Fig. \ref{subfig: cost vs n}, StagewiseHFL achieves a reduction in system cost by 7.7\%, 10.26\%, 19.5\%, 22.2\% and 32.6\% as compared to KLDMinimization, ClientSelOnly, C2EAssocOnly, C2EGreedyAssoc, and FedCS, respectively. A similar trend, where our StagewiseHFL demonstrates cost-saving performance that ranks second with a small margin only to OrigProbSolver, is seen in Figs. \ref{subfig: cost vs s}-\ref{subfig: cost vs pi}. 

\subsubsection{Evaluation of  running time}
\label{subsubsec:evaluation on running time}
Fig. \ref{fig:performace on running time} evaluates the performance on time efficiency reflected by running time, upon having different values of $|\mathcal{N}|$, $|\mathcal{S}|$, $D_{\textrm {min}}$, and $p_i$. Our StagewiseHFL significantly improves time efficiency compared to the benchmark methods under consideration, with the exception of C2EAssocOnly in some certain scenarios. This is because C2EAssocOnly does not prioritize optimizing client selection, thereby streamlining the decision-making process (however, it suffers from a large cost shown in Fig. \ref{fig:performance on cost}).  

Inspecting Figs. \ref{subfig:running time vs n}-\ref{subfig:running time vs dmin}, varying $|\mathcal{N}|$, $|\mathcal{S}|$ and $D_{\textrm {min}}$ changes the problem scale, which in turn affects the complexity of algorithms used for client selection and C2E association. For instance, in Fig. \ref{subfig:running time vs n}, the delay on decision-making of ClientSelOnly increases from 600ms to 9.8s with $|\mathcal{N}|$ rises from 70 to 120. Note that although OrigprobSolver outperforms our StagewiseHFL in terms of cost savings (see Fig. \ref{fig:performance on cost}), its inference delay becomes prohibitively high as the problem size increases (with the rise of $|\mathcal{N}|$, $|\mathcal{S}|$ and $D_{\textrm {min}}$). For example, its decision-making delay varies from 10.44s to 346s as $|\mathcal{S}|$ increases from 3 to 7, shown in Fig. \ref{subfig:running time vs s}. While the time cost does escalate with an increase in problem size, our approach continues to be viable across various environments compared to all benchmarks. This superior performance is primarily due to our strategically designed Plan A, which pre-selects a specific number of long-term clients. This significantly mitigates the problem size encountered in Plan B, thereby accelerating  the real-time decision-making process for client selection and C2E associations. A similar trend is found in Fig. \ref{subfig:running time vs dmin}, where rising tolerable minimum data size $D_{\textrm {min}}$ results in a growth of inference time, e.g., the inference time increases from 22s to 80s for KLDMinimization while from 24ms to  260ms for our StagewiseHFL when $D_{\textrm {min}}$ varies from 2000 to 5000. Also, as the client participation probability rises, our StagewiseHFL increasingly relies less on Plan B during actual training (since more selected long-term clients will participate in training), resulting in reduced complexity and further enhancing inference speeds. As illustrated in Fig. \ref{subfig:running time vs pi}, having the client online probability range of [0.7, 1), StagewiseHFL achieves an average of only 12ms per global aggregation round, which is even faster than C2EAssocOnly.

In summary, Fig. \ref{fig:performace on running time} reveals that the running time of StagewiseHFL consistently remains below 500ms across various problem scales, which offers a commendable reference for future low-overhead HFL design in dynamic environments.

\subsection{Analysis of the Impact of Key Parameters}
\label{subsec:analysis on key paras}
Since improper parameter settings can adversely model training performance, we further evaluate various combinations of ${\it{KLD}}_{\textrm {max}}$ and $D_{\textrm {min}}$, as well as $\mathcal{T}$ and $\mathcal{L}$ in this section, to show how they affect the system while identifying the best combinations.
\subsubsection{Analysis of the impact of risk-related parameters}
\label{subsubsec:analysis on T and L}
Our StagewiseHFL introduces distinctive considerations on risk evaluation in Plan A, representing one of our contributions, which sets this work apart from existing studies. As two key parameters that affect the risks are KLD and data size, we first illustrate how ${\it{KLD}}_{\textrm {max}}$ and $D_{\textrm {min}}$ impact the overall system cost and the required global aggregation rounds to reach a target test accuracy in Table \ref{tab:dmin vs kldmax}. To achieve 90\% accuracy on MNIST within $\mathcal{T}=5$ and $\mathcal{L}=3$, Table \ref{tab:dmin vs kldmax} shows that lowering the KLD for edge data can reduce the number of global aggregations. Considering $D_{\textrm {min}}=2500$, setting the value of ${\it{KLD}}_{\textrm {max}}$ from 0.5 to 0.2  reduces global iterations from 29 to 19. However, under non-IID data, setting the value of ${\it{KLD}}_{\textrm {max}}$ too low often leads to the selection of more clients for participation, increasing system costs.  For example, when ${\it{KLD}}_{\textrm {max}}=0.1$, the system costs are higher than those with ${\it{KLD}}_{\textrm {max}}=0.2$ for all $D_{\textrm {min}}$ settings. Extended evaluation in Appx. \ref{appx:evaluation IV} further examine the relationship between ${\it{KLD}}_{\textrm {max}}$ and model accuracy on CIFAR-100 and Tiny-ImageNet, as shown in Tables \ref{tab:acc vs kldmax on cifar100} and \ref{tab:acc vs kldmax on tiny-imagenet}. These results confirm that the positive effect of KLD control on accuracy also extends to more complex datasets.
For the tolerable data size, although raising $D_{\textrm {min}}$ can help improve model accuracy, when $D_{\textrm {min}}$ is set to a higher value, incorporating an excessive amount of data into training may decelerate convergence, thereby necessitating an increased number of global aggregations. For instance, upon having  ${\it{KLD}}_{\textrm {max}}=0.4$, increasing $D_{\textrm {min}}$ from 2000 to 4000 increases the required global aggregation rounds from 27 to 36. In sum, the careful selection of $D_{\textrm {min}}$ and ${\it{KLD}}_{\textrm {max}}$ is crucial for optimizing the overall performance.
\begin{table}[H]
	\small
	\centering
	\caption{	
		Impact of $D_{\textrm {min}}$ vs. ${\it{KLD}}_{\textrm {max}}$ for reaching the accuracy of 90\%}
        \label{tab:dmin vs kldmax}
	\setlength{\tabcolsep}{1.5pt}
	\begin{tabular}{|c|c|@{\hskip 3pt}|c|c|c|c|c|}
		\hline
		\rowcolor{verylightgray}
		\multicolumn{2}{|c|@{\hskip 3pt}|}{\textbf{Cost/}} & \multicolumn{5}{c|}{${\it{KLD}}_{\textrm {max}}$} \\ \cline{3-7}
		  
		\multicolumn{2}{|c|@{\hskip 3pt}|}{\textbf{Aggregations}} & 0.1   & 0.2   & 0.3   & 0.4   & 0.5 \\ 
	
		\hline
		\hline
		\rowcolor{verylightgray}
		& 2000  & 18.09/23 & 11.02/23 & 9.65/22 & 11.29/27 & 11.89/28 \\ 
	\hhline{|~|------|}
		& 2500  & 17.31/22 & \textbf{9.11/19} & 10.91/26 & 13.15/29 & 13.18/29  \\ \hhline{|~|------|}
		\rowcolor{verylightgray}& 3000  & 16.53/21 & 10.09/20 & 14.18/30 & 17.48/35 & 21.75/47 \\ \hhline{|~|------|}
		& 4000  & 17.37/22 & 14.08/24 & 11.80/22 & 18.79/36 & 20.00/36\\ \hhline{|~|------|}
		\rowcolor{verylightgray}\multirow{-5}{*}{$D_{\textrm {min}}$} & 5000  & 16.58/21 & 14.75/22 & 14.36/23 & 14.73/23 & 15.52/25\\ 
		\hline

	\end{tabular}

\end{table}

\subsubsection{Analysis of the impact of training configuration}
\label{subsubsec:analysis on dmin and kldmax}
 We next analyze the impact of $\mathcal{T}$ and $\mathcal{L}$ on the overall system cost. Although $\mathcal{T}$ and $\mathcal{L}$ could ideally be optimized alongside other parameters to minimize overall system costs, our experiments are conducted under constraints including access limitations $M_j$ of ESs, and fixed data volumes. As a result, the number of clients selected in each global iteration remains relatively consistent, enabling us to identify a practical and broadly applicable combination of $\mathcal{T}$ and $\mathcal{L}$. To keep the problem tractable and to ensure fair comparisons across different experimental settings, we follow existing works \cite{ref5,ref8,ref10,ref49,ref61}, which similarly adopt fixed settings of $\mathcal{T}$ and $\mathcal{L}$\footnote{Note that the primary decision variable in our formulation is the client-to-edge (C2E) association matrix $\bm{A}$. Fixing $\mathcal{T}$ and $\mathcal{L}$ allows us to better highlight the core contribution of this work—namely, the establishment of a novel two-stage optimization paradigm for HFL, which has rarely been explored in the literature. While many factors in HFL can indeed be optimized, our focus here is on client association and selection, as they are most aligned with the proposed paradigm. The optimization of $\mathcal{T}$ and $\mathcal{L}$ will be considered as an interesting direction for future research.}. Nevertheless, $\mathcal{T}$ and $\mathcal{L}$ are not arbitrarily chosen.  Through extensive tests over a wide range of candidate values, we verified that the adopted configuration consistently provides competitive performance, making it a reasonable and representative choice across our experiments. As shown in Table \ref{tab:t vs l}, when $\mathcal{T}$ is relatively small (e.g., $\mathcal{T}=5$ and $\mathcal{L}=10$), increasing the value of $\mathcal{L}$ can effectively decrease the number of required global iterations, thereby enhancing the model's convergence speed. At a specific intermediate value (e.g., $\mathcal{L}=3$), we achieve a joint minimum for both the system cost and the required number of global aggregations. However, with larger values of $\mathcal{T}$, the number of global aggregations needed for achieving the target accuracy may not necessarily reduce; in fact, it could result in inferior performance outcomes. Due to the imbalanced data distribution across clients and a lower edge aggregation frequency (that is, executing more local iterations within a single round of edge iteration), the local models may become biased, preventing the global model from convergence. Particularly if $\mathcal{L}=4$ and $\mathcal{L}=5$ when $\mathcal{T}=50$, the global model never reaches the accuracy of 90\% and thus the training cost is denoted by infinity.
 
 Overall, these empirical results indicate that although $\mathcal{T}$ and $\mathcal{L}$ are treated as fixed hyper-parameters, their settings are carefully tuned to ensure that the reported results reflect near-optimal system performance under different scenarios. In fact, the comparisons presented in Table \ref{tab:t vs l} implicitly serve as an ablation study, as varying $\mathcal{T}$ and $\mathcal{L}$ allows us to examine their influence on convergence speed, training cost, and final accuracy.

\subsection{Threats to Validity}
\label{subsec:threats}
Since this work tries the first attempt to having two complementary plans for client selection, this section is dedicated to studying the validity of StagewiseHFL through inspecting factors that can compromise its effectiveness. Our goal is to highlight the benefits of our method from a rational  perspective.
\subsubsection{Threats to internal validity}
\label{subsubsec:internal threats}
\begin{figure}[b]
\label{fig:invalidity}
	\centering
	\subfigure[Cost vs. $p_i$]{
    \label{subfig:invalid cost vs pi}
		\includegraphics[trim=1cm 0cm 1.8cm 0.5cm, clip, width=0.47\columnwidth]{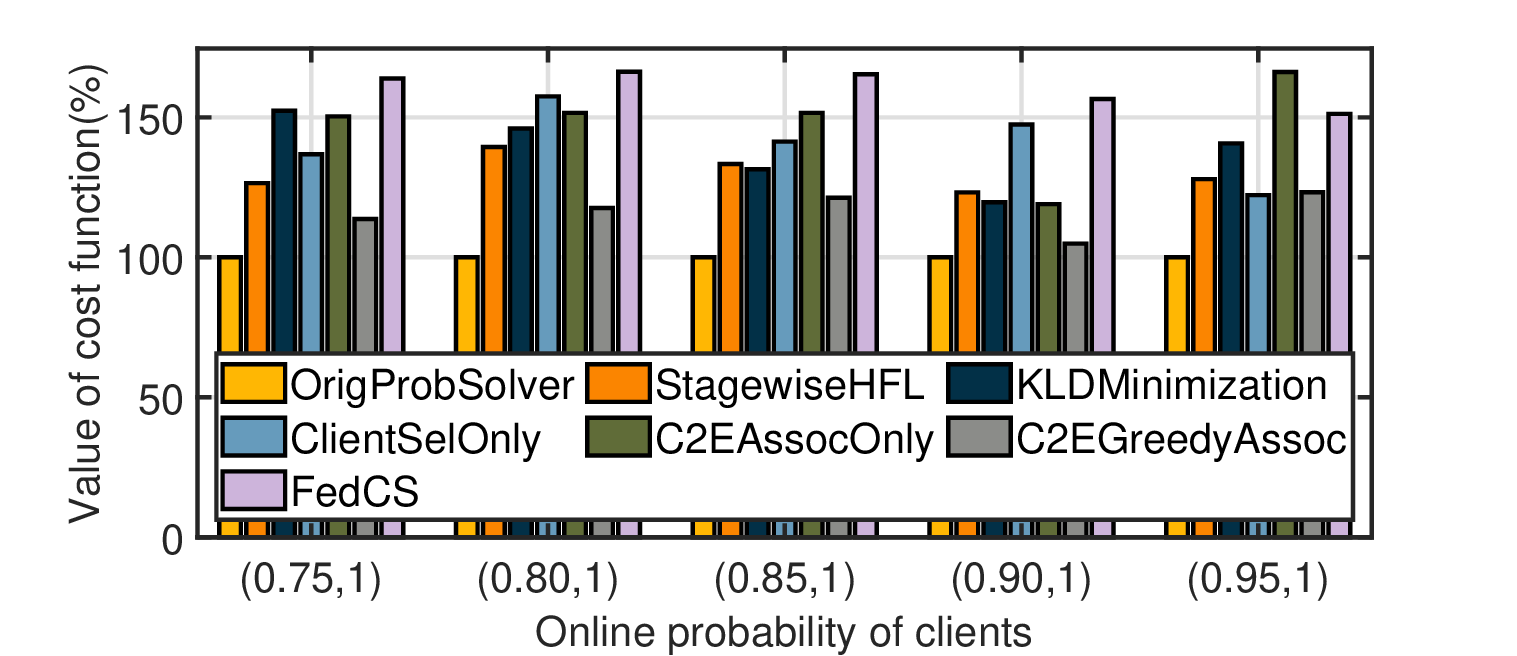}
	}
	\subfigure[Running time vs. $|\mathcal{N}|$ for Plan A in StagewiseHFL]{
    \label{subfig:invalid running time vs n}
		\includegraphics[trim=1cm 0cm 1.8cm 0.5cm, clip, width=0.47\columnwidth]{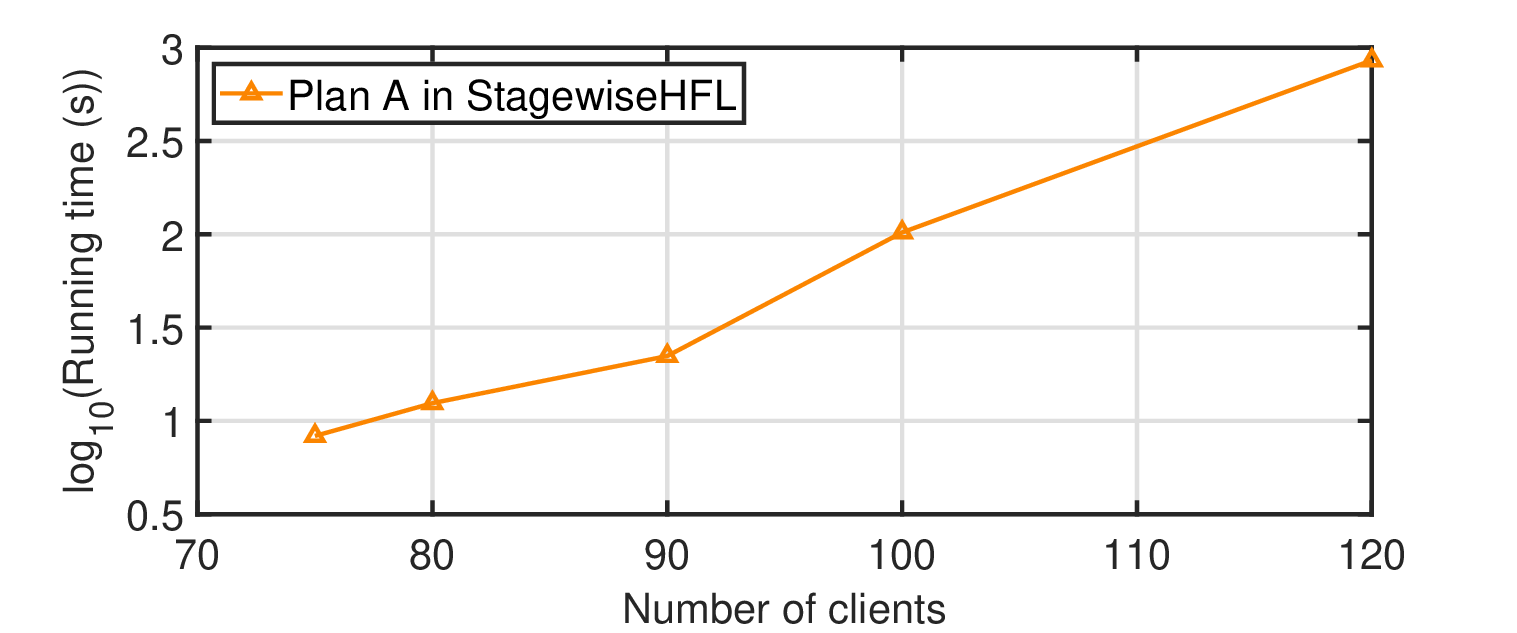}
	}
	\caption{Invalidity of StagewiseHFL.}

\end{figure}

The first threat to the validity of our StagewiseHFL is that the experimental environment may have favored it in the previous results. To fairly compare the performance of different methods, we change the settings of different parameters, e.g., $D_{\textrm {min}}$, and $p_i$, while we also perform 100 experiments for diverse parameter settings (Table \ref{tab:simulation setting}). Note that when clients have a high online probability, e.g., an extreme case in which all the clients will join in each global training rounds ($p_i=1$, $\forall \bm{n}_i \in \mathcal{N}$), our design of pre-selection and pre-association of clients in Plan A may be no longer needed. The reason is that our plan A may result in a high overlap in the selected long-term clients, during each global iteration, potentially decreasing the diversity of training data while slowing down the convergence speed of the global model. As shown in Fig. \ref{subfig:invalid cost vs pi}, when the interval of $p_i$ varies from [0.75, 1) to [0.95, 1), our StagewiseHFL incurs performance losses of 26.5\%, 39.4\%, 33.6\%, 23.2\%, 28.0\% compared to OrigprobSolver in terms of cost-savings. It may also perform worse in comparison with C2EGreedyAssoc, and even worse than the ClientSelOnly and KLDMinimization in some cases. Such observations indicate that our StagewiseHFL should be used in scenarios with intermittent client participation. 
\begin{table}[t]
	\small
	\centering
	\caption{	
		Impact of $\mathcal{T}$ vs. $\mathcal{L}$ for reaching the accuracy of 90\% }
        \label{tab:t vs l}
	\setlength{\tabcolsep}{1.5pt}
	\begin{tabular}{|c|c|@{\hskip 3pt}|c|c|c|c|c|}
		\hline
		\rowcolor{verylightgray}
		\multicolumn{2}{|c|@{\hskip 3pt}|}{\textbf{Cost/}} & \multicolumn{5}{c|}{$\mathcal{L}$} \\ \cline{3-7}  
		\multicolumn{2}{|c|@{\hskip 3pt}|}{\textbf{Aggregations}} & 1   & 2   & 3   & 4   & 5 \\ \hline
		\hline
		\rowcolor{verylightgray} & 5 & 10.32/52 & 11.06/28 & \textbf{9.11/19} & 12.58/16 & 15.69/16 \\ \cline{2-7}
		& 10  & 12.03/41 & 13.45/23 & 14.21/16 & 18.90/16 & 16.23/11
		\\ 	\hhline{|~|------|}
		\rowcolor{verylightgray}& 20  & 13.86/28 & 15.83/16 & 16.30/11 & 21.70/11 & 27.11/11 \\ 	\hhline{|~|------|}
		& 30  & 31.16/23 & 15.25/11 & 22.83/11 & 30.41/11 & 37.99/11\\ 	\hhline{|~|------|}
		\rowcolor{verylightgray}\multirow{-5}{*}{$\mathcal{T}$}& 50  & 17.46/16 & 23.95/11 & 35.88/11 & $\infty$ & $\infty$\\ \hline
	\end{tabular}
\end{table}

\subsubsection{Threats to external validity}
\label{subsubsec:external threats}
We next acknowledge that the scalability of the long-term decision-making component in our method may be affected by the network size. As discussed in Sec. \ref{subsubsec:plan a}, the computational complexity of Plan A grows with the number of clients $|\mathcal{N}|$, potentially leading to long decision-making times when $|\mathcal{N}|$ is large. As illustrated in Fig. \ref{subfig:invalid running time vs n}, when $|\mathcal{N}|$ increases from 75 to 120, the runtime of Plan A rises from 8.3s to 855.8s. This observation highlights an inherent trade-off in StagewiseHFL, where a certain amount of long-term optimization time must be sacrificed to enable smoother subsequent training. 
	
Nevertheless, this limitation does not affect practicality: Plan A runs once before training and later in parallel, so its overhead does not hinder real-time decisions. Results in Sec. \ref{sec:evaluations} show that despite this cost, our two-stage framework significantly improves scalability, efficiency, and deployment feasibility, highlighting the necessity of the hybrid strategy.


\section{Conclusion and Future Work}
\label{sec:future work}
This paper we studied client selection and C2E association for dynamic HFL network, for which we investigate a stagewise decision-making  methodology with two stages, namely Plan A and Plan B. In particular, in Plan A, we introduced the concept of ``continuity'' of clients, strategically determining long-term clients and associating them with appropriate ESs. In Plan B, we propose a method for rapidly identifying backup clients in case those recruited by Plan A were unavailable in practical model training rounds. Comprehensive simulations on different datasets and models demonstrated the commendable performance of our method compared to other benchmark methods. 

This work is the first to systematically explore stagewise decision-making in FL and HFL, showing its potential to accelerate training while reducing overhead. The promising results and clear rationale are expected to inspire further research on its theoretical optimality. Future research includes modeling uncertainties in client computation and communication, optimizing when to trigger Plan A in dynamic environments, and studying the impact of hardware accelerators (e.g., GPUs/TPUs) in heterogeneous settings. Another promising direction is leveraging D2D communications to improve local aggregation under intermittent client availability.



\newpage
\clearpage
\onecolumn
\appendices
\section{List of Notations}
\label{appx:notations}
\begin{table*}[htbp]
	\small
	\caption{List of Notations}
    \label{notation}
	\centering
	\begin{tabular}{|>{\centering\arraybackslash}m{2.2cm}|>{\centering\arraybackslash}m{6.0cm}|@{\hskip 3pt}|>{\centering\arraybackslash}m{2.2cm}|>{\centering\arraybackslash}m{6.0cm}|}
		\hline
		\rowcolor{verylightgray}
		\bf{Notation} & \bf{Definition} & \bf{Notation} & \bf{Definition}\\
		\hline
		\hline
		$\mathcal{N}$/$\bm{n}_{i}$ & Set of clients/the $i^\text{th}$ client &	$\mathcal{S}$/$s_{j}$ & Set of ESs/the $j^\text{th}$ ES\\
		\hline
		\rowcolor{verylightgray}
		$\mathcal{Z}$/$z
		_{h}$ & Label set/the $h^\text{th}$ label & $\mathcal{D}_{i}$ & Dataset of client $\bm{n}_{i}$\\
		\hline
		$d_i^{(g)}$ & Data size of client $\bm{n}_i$ & $D_j^{(g)}$ & Data size of all clients associated with ES $s_j$\\
		\hline
		\rowcolor{verylightgray}
		$D^{(g)}$ & Data size of all clients that take part in the training process & $\boldsymbol{x}_{k}$/${y}_{k}$ & Input vector/labeled output of the $k^\text{th}$ data\\
		\hline
		${v}_{i}$ & CPU frequency of client $\bm{n}_{i}$ & ${q}_{i}$ & Transmission power of client $\bm{n}_{i}$\\
		\hline
		\rowcolor{verylightgray}
		$\xi_{i}^{(g)}$ & Binary indicator describing the participation of a client $\bm{n}_{i}$ in $g^{\text {th}}$ global iteration & 
		$\boldsymbol{A}^{(g)}$ & Association matrix indicating whether client $\bm{n}_i$ is associated with ES $s_j$ or not\\
		\hline
		$\boldsymbol{\omega}_{i}^{(g;\ell, t)}$ & Local model parameter of client $\bm{n}_i$ in $t^\text{th}$ local update during $\ell^\text{th}$ edge aggregation round & $\boldsymbol{\omega}_{j}^{(g;\ell)}$ & Edge model parameter of ES $s_j$ in $\ell^\text{th}$ edge aggregation round\\
		\hline
		\rowcolor{verylightgray}
		$\boldsymbol{\omega}^{(g)}$ & Global model parameter & $\eta$ & Learning rate\\
		\hline
		$\mathcal{T}$/$\mathcal{L}$ & Edge/global aggregation frequency & $c_i$ & Number of CPU cycles required for processing one sample data on $\bm{n}_i$\\
		\hline
		\rowcolor{verylightgray}
		$\alpha_i$ & Effective capacitance coefficient of $\bm{n}_i$’s computing chipset & $B_j$ & Allocated bandwidth for each client associated with $s_j$\\
		\hline
		$M_j$ & Maximum number of clients that can access to $s_j$ & $h_{i,j}^{(g)}$ & Channel gain of between $\bm{n}_i$ and $s_j$\\
		\hline
		\rowcolor{verylightgray}
		$N_0$ & Noise power spectral density & $R_{i,j}^{(g)}$ & Data transmission rate of the uplink from client $\bm{n}_i$ to ES $s_j$\\
		\hline
		$\Omega$ & Data size of the model parameters uploaded by clients & $p_i$ & Participation probability of $\bm{n}_i$\\
		\hline
		\rowcolor{verylightgray}
		$t_i^{\textrm {cmp}}$/$e_i^{\textrm {cmp}}$ & Computation latency/energy consumption for a client $\bm{n}_i$ to perform one local training iteration & $t_{i,j}^{(g),\textrm {com}}$/$e_{i,j}^{(g),\textrm {com}}$ & Transmission delay/energy consumption between $\bm{n}_i$ and $s_j$ during one edge iteration\\
		\hline
		$T_{i,j}^{(g)}$/$E_{i,j}^{(g)}$ & Delay/energy consumption caused by the interaction between $\bm{n}_i$ and $s_j$ during one edge iteration & $T_j^{(g),\textrm {com}}$/$E_j^{(g),\textrm {com}}$ & Transmission delay/energy consumption  for ES $s_j$ uploading the edge model to the CS\\
		\hline
		\rowcolor{verylightgray}
		$\mathbb{T}^{(g)}$/$\mathbb{E}^{(g)}$ & Overall training delay/energy consumption during $g^{\text {th}}$ global iteration & $\mathbb{C}^{(g)}$ & System ``continuity'' describing the geometric mean of online probability of each selected client\\
		\hline
		$\boldsymbol{y}_i$ & Vector representing the data distribution of client $\bm{n}_i$ & $\boldsymbol{P}_j^{(g)}$/$\boldsymbol{Q}$ & Data distribution of ES $s_j$/reference data distribution\\
		\hline
		\rowcolor{verylightgray}
		${\it{KLD}}_{\textrm {max}}$ & Upper bound of KLD the covered data associated with each ES & $D_{\textrm {min}}$ & lower bound of size of the covered data associated with each ES\\
		\hline
	\end{tabular}
\end{table*}

\section{Pseudocode of Alg. \ref{algorithm1}}
\label{appx:pseudocode}
\setcounter{algocf}{0}
\begin{algorithm}[h]
	\footnotesize
	\caption{Stagewise Hierarchical Federated Learning (StagewiseHFL)}\label{algorithm1}
	\SetKwInOut{Input}{Input}\SetKwInOut{Output}{Output}
	\Input{the set of ES $\mathcal{S}$; the set of clients $\mathcal{N}$}
	\Output{the long-term C2E association decision $\widehat{\boldsymbol{A}}^{(g)}$; the real-time C2E association decision $\widetilde{\boldsymbol{A}}^{(g)}$}
	
	\BlankLine
	\%{\bf{Long-term decision-making stage (Plan A)}}
	
	Estimate $p_i$ ($\forall \bm{n}_i \in \mathcal{N}$) before practical model training;
	
	Solve $\bm{\mathcal{P}}_1$ through Algs. \ref{algorithm2}-\ref{algorithm3} and obtain the long-term C2E association decision $\widehat{\boldsymbol{A}}^{(0)}$;
	
	\BlankLine
	\%{\bf{Real-time decision-making stage (Plan B)}}
	
	\For{$g=1$ \textnormal{to} ${g}_{\textnormal {max}}$}{
		$\widehat{\boldsymbol{A}}^{(g)}=\widehat{\boldsymbol{A}}^{(0)}$;
		
		Solve $\bm{\mathcal{P}}_2$ in the $g^{\text{th}}$ global aggregation through Algs. \ref{algorithm4} and obtain the real-time C2E association decision $\widetilde{\boldsymbol{A}}^{(g)}$;}
	
	{\bf{return}} $\widehat{\boldsymbol{A}}^{(g)}$ and $\widetilde{\boldsymbol{A}}^{(g)}$;
\end{algorithm}

\section{Derivation of \eqref{p1_11f}}
\label{appx:derivation of 11f}
By substituting the definition of $\operatorname{KLD}\left(\widehat{\boldsymbol{P}}_j^{(g)}||\boldsymbol{Q}\right)$ from \eqref{kld} and \eqref{distribution} into \eqref{p1_11a}, we can obtain
\begin{equation}
	\label{11f_1}
	\operatorname{Pr}\left(\sum_{h=1}^{|\mathcal{Z}|}\frac{\sum_{i=1}^{|\mathcal{N}|}\widehat{a}_{i, j}^{(g)}\xi_{i}^{(g)}\boldsymbol{y}_{i}(h)}{\sum_{i=1}^{|\mathcal{N}|}\widehat{a}_{i, j}^{(g)}\xi_{i}^{(g)}d_{i}}\operatorname{log}\frac{\sum_{i=1}^{|\mathcal{N}|}\widehat{a}_{i, j}^{(g)}\xi_{i}^{(g)}\boldsymbol{y}_{i}(h)}{\boldsymbol{Q}(h)\widehat{a}_{i, j}^{(g)}\xi_{i}^{(g)}d_{i}}>{\it{KLD}}_{\textrm{max}}-\Delta_k\right)\le \delta,\forall s_{j}\in\mathcal{S}.
\end{equation}
Note that the fractions in the above \eqref{11f_1} have no meaning when $\sum_{i=1}^{|\mathcal{N}|}\widehat{a}_{i, j}^{(g)}\xi_{i}^{(g)}=0$, i.e., all clients selected and associated with ES $s_j$ in plan A are offline, then, we consider $\operatorname{KLD}\left(\widehat{\boldsymbol{P}}_j^{(g)}||\boldsymbol{Q}\right)$ to be $+\infty$ in such a case. Given the law of total probability, we can decompose the left side of \eqref{p1_11a} as:
\begin{align}
	&\operatorname{Pr}\!\left(\operatorname{KLD}\left(\widehat{\boldsymbol{P}}_j^{(g)}||\boldsymbol{Q}\right)\!>\!{\it{KLD}}_{\textrm{max}}\!-\!\Delta_k\right) \!=\!\operatorname{Pr}\left(\sum_{i=1}^{|\mathcal{N}|}\widehat{a}_{i, j}^{(g)}\xi_{i}^{(g)}\!=\!0\right)\!\times\!\operatorname{Pr}\!\left(\operatorname{KLD}\left(\widehat{\boldsymbol{P}}_j^{(g)}||\boldsymbol{Q}\right)\!>\!{\it{KLD}}_{\textrm{max}}\!-\!\Delta_k\Big|\sum_{i=1}^{|\mathcal{N}|}\widehat{a}_{i, j}^{(g)}\xi_{i}^{(g)}\!=\!0\right)\notag\\
	&+ \operatorname{Pr}\left(\sum_{i=1}^{|\mathcal{N}|}\widehat{a}_{i, j}^{(g)}\xi_{i}^{(g)}\neq 0\right)\times\operatorname{Pr}\left(\operatorname{KLD}\left(\widehat{\boldsymbol{P}}_j^{(g)}||\boldsymbol{Q}\right)>{\it{KLD}}_{\textrm{max}}-\Delta_k\Big|\sum_{i=1}^{|\mathcal{N}|}\widehat{a}_{i, j}^{(g)}\xi_{i}^{(g)}\neq 0\right)\notag\\
	&=\! \prod_{i\mid \widehat{a}_{i, j}^{(g)}=1}\!{\left(1-p_i\right)}+\left(1-\!\prod_{i\mid \widehat{a}_{i, j}^{(g)}=1}{\left(1-p_i\right)}\right)
	\!\times\operatorname{Pr}\left(\sum_{h=1}^{|\mathcal{Z}|}\frac{\sum_{i=1}^{|\mathcal{N}|}\widehat{a}_{i, j}^{(g)}\xi_{i}^{(g)}\boldsymbol{y}_{i}(h)}{\sum_{i=1}^{|\mathcal{N}|}\widehat{a}_{i, j}^{(g)}\xi_{i}^{(g)}d_{i}}\operatorname{log}\frac{\sum_{i=1}^{|\mathcal{N}|}\widehat{a}_{i, j}^{(g)}\xi_{i}^{(g)}\boldsymbol{y}_{i}(h)}{\boldsymbol{Q}(h)\widehat{a}_{i, j}^{(g)}\xi_{i}^{(g)}d_{i}}\Big|\sum_{i=1}^{|\mathcal{N}|}\widehat{a}_{i, j}^{(g)}\xi_{i}^{(g)}\neq 0\right).\!\!
\end{align}

For the case where $\sum_{i=1}^{|\mathcal{N}|}\widehat{a}_{i, j}^{(g)}\xi_{i}^{(g)}\neq 0$, let $\bm{R}_j^{(g)}(h)=
\dfrac{\sum_{i=1}^{|\mathcal{N}|}\widehat{a}_{i, j}^{(g)}\xi_{i}^{(g)}\boldsymbol{y}_{i}(h)}{\sum_{i=1}^{|\mathcal{N}|}\widehat{a}_{i, j}^{(g)}\xi_{i}^{(g)}d_{i}}$, and define $N_j^{(g)}=\sum_{i=1}^{|\mathcal{N}|}\widehat{a}_{i, j}^{(g)}$ as the number of clients associated with ES $s_j$. Then, we have the following scenarios.

When $N_j^{(g)}=1$, we have $\bm{R}_j^{(g)}(h)=\dfrac {\bm{y}_1(h)\xi_1^{(g)}}{d_1\xi_1^{(g)}}=\dfrac{\bm{y}_1(h)}{d_1}.$

When $N_j^{(g)}=2$, we have $\bm{R}_j^{(g)}(h)=\dfrac {\bm{y}_1(h)\xi^{(g)}_1+\bm{y}_2(h)\xi^{(g)}_2}{d_1\xi^{(g)}_1+d_2\xi^{(g)}_2}=\dfrac{\bm{y}_1(h)\xi^{(g)}_1+\bm{y}_1(h)\dfrac{d_2}{d_1}\xi^{(g)}_2}{d_1\xi^{(g)}_1+d_2\xi^{(g)}_2}+\dfrac{\left(\bm{y}_2(h)-\bm{y}_1(h)\dfrac{d_2}{d_1}\right)\xi^{(g)}_2}{d_1\xi^{(g)}_1+d_2\xi^{(g)}_2}=\dfrac{\bm{y}_1(h)}{d_1}+\dfrac{d_2\left(\dfrac{\bm{y}_2(h)}{d_2}-\dfrac{\bm{y}_1(h)}{d_1}\right)\xi^{(g)}_2}{d_1\xi^{(g)}_1+d_2\xi^{(g)}_2}.$

When $N_j^{(g)}=3$, we have $\bm{R}_j^{(g)}(h)=\dfrac {\bm{y}_1(h)\xi^{(g)}_1+\bm{y}_2(h)\xi^{(g)}_2+\bm{y}_3(h)\xi^{(g)}_3}{d_1\xi^{(g)}_1+d_2\xi^{(g)}_2+d_3\xi^{(g)}_3}=\dfrac{\bm{y}_1(h)\xi^{(g)}_1+\bm{y}_1(h)\dfrac{d_2}{d_1}\xi^{(g)}_2+\bm{y}_1(h)\dfrac{d_3}{d_1}\xi^{(g)}_3}{d_1\xi^{(g)}_1+d_2\xi^{(g)}_2+d_3\xi^{(g)}_3}+\dfrac{\left(\bm{y}_2(h)-\bm{y}_1(h)\dfrac{d_2}{d_1}\right)\xi^{(g)}_2}{d_1\xi^{(g)}_1+d_2\xi^{(g)}_2+d_3\xi^{(g)}_3}+\dfrac{\left(\bm{y}_3(h)-\bm{y}_1(h)\dfrac{d_3}{d_1}\right)\xi^{(g)}_3}{d_1\xi^{(g)}_1+d_2\xi^{(g)}_2+d_3\xi^{(g)}_3}=\dfrac{\bm{y}_1(h)}{d_1}+\dfrac{d_2\left(\dfrac{\bm{y}_2(h)}{d_2}-\dfrac{\bm{y}_1(h)}{d_1}\right)\xi^{(g)}_2}{d_1\xi^{(g)}_1+d_2\xi^{(g)}_2+d_3\xi^{(g)}_3}+\dfrac{d_3\left(\dfrac{\bm{y}_3(h)}{d_3}-\dfrac{\bm{y}_1(h)}{d_1}\right)\xi^{(g)}_3}{d_1\xi^{(g)}_1+d_3\xi^{(g)}_2+d_3\xi^{(g)}_3}.$

Therefore, when $N_j^{(g)}$ takes a general positive integer, we can get	
\begin{align}
	\bm{R}_j^{(g)}(h) &= \dfrac {\bm{y}_1(h)\xi^{(g)}_1 + \bm{y}_2(h)\xi^{(g)}_2 + \ldots + \bm{y}_{N_j^{(g)}}(h)\xi^{(g)}_{N_j^{(g)}}}{d_1\xi^{(g)}_1 + d_2\xi^{(g)}_2 + \ldots +  d_{N_j^{(g)}}\xi^{(g)}_{N_j^{(g)}}} = \dfrac{\bm{y}_1(h)}{d_1} + \dfrac{d_2\left(\dfrac{\bm{y}_2(h)}{d_2} - \dfrac{\bm{y}_1(h)}{d_1}\right)\xi^{(g)}_2}{d_1\xi^{(g)}_1 + d_2\xi^{(g)}_2 + \ldots +d_{N_j^{(g)}}\xi^{(g)}_{N_j^{(g)}}} \notag\\
	&+\ldots+\dfrac{d_{N_j^{(g)}}\left(\dfrac{\bm{y}_{N_j^{(g)}}(h)}{d_{N_j^{(g)}}} - \dfrac{\bm{y}_1(h)}{d_1}\right)\xi^{(g)}_{N_j^{(g)}}}{d_1\xi^{(g)}_1 + d_2\xi^{(g)}_2 + \ldots + d_{N_j^{(g)}}\xi^{(g)}_{N_j^{(g)}}}.
\end{align}
Note that when $\dfrac{\bm{y}_1(h)}{d_1}=\underset{i\mid \widehat{a}_{i, j}^{(g)}=1}{\max}\left\{\dfrac{\bm{y}_i(h)}{d_i}\right\}$, all but the first term $\dfrac{\bm{y}_1(h)}{d_1}$ in the above equation are less than 0, i.e., $\bm{R}_j^{(g)}(h)=\dfrac{\bm{y}_1(h)}{d_1}+\dfrac{d_2\left(\dfrac{\bm{y}_2(h)}{d_2}-\dfrac{\bm{y}_1(h)}{d_1}\right)\xi^{(g)}_2}{d_1\xi^{(g)}_1+d_2\xi^{(g)}_2+\ldots+d_{N_j^{(g)}}\xi^{(g)}_{N_j^{(g)}}}+\ldots+\dfrac{d_{N_j^{(g)}}\left(\dfrac{\bm{y}_{N_j^{(g)}}(h)}{d_{N_j^{(g)}}}-\dfrac{\bm{y}_1(h)}{d_1}\right)\xi^{(g)}_{N_j^{(g)}}}{d_1\xi^{(g)}_1+d_2\xi^{(g)}_2+\ldots+d_{N_j^{(g)}}\xi^{(g)}_{N_j^{(g)}}}\le \dfrac{\bm{y}_1(h)}{d_1}$. Similarly, when $\dfrac{\bm{y}_1(h)}{d_1}=\underset{i\mid \widehat{a}_{i, j}^{(g)}=1}{\min}\left\{\dfrac{\bm{y}_i(h)}{d_i}\right\}$, $\bm{R}_j^{(g)}(h)=\dfrac{\bm{y}_1(h)}{d_1}+\dfrac{d_2\left(\dfrac{\bm{y}_2(h)}{d_2}-\dfrac{\bm{y}_1(h)}{d_1}\right)\xi^{(g)}_2}{d_1\xi^{(g)}_1+d_2\xi^{(g)}_2+\ldots+d_{N_j^{(g)}}\xi^{(g)}_{N_j^{(g)}}}+\ldots+\dfrac{d_{N_j^{(g)}}\left(\dfrac{\bm{y}_{N_j^{(g)}}(h)}{d_{N_j^{(g)}}}-\dfrac{\bm{y}_1(h)}{d_1}\right)\xi^{(g)}_{N_j^{(g)}}}{d_1\xi^{(g)}_1+d_2\xi^{(g)}_2+\ldots+d_{N_j^{(g)}}\xi^{(g)}_{N_j^{(g)}}}\ge \dfrac{\bm{y}_1(h)}{d_1}$.

Accordingly, we can easily derive that $\underset{i\mid \widehat{a}_{i, j}^{(g)}=1}{\min}\left\{\dfrac{\bm{y}_i(h)}{d_i}\right\}\le\bm{R}_j^{(g)}(h)\le\underset{i\mid \widehat{a}_{i, j}^{(g)}=1}{\max}\left\{\dfrac{\bm{y}_i(h)}{d_i}\right\}$.

To facilitate the analysis, we construct a function $\operatorname{U}\left( \cdot\right)$ on $\bm{R}_j^{(g)}(h)$ such that $\operatorname{U}\left(\bm{R}_j^{(g)}(h)\right)=\bm{R}_j^{(g)}(h)\operatorname{log}\left(\dfrac{\bm{R}_j^{(g)}(h)}{\bm{Q}(h)}\right)$. This function is evidently convex in the domain $(0,+\infty)$,
and it attains a local minimum at $\bm{R}_j^{(g)}(h)=\dfrac{\bm{Q}(h)}{e}$. Based on the relationship between the local minimum point $\dfrac{\bm{Q}(h)}{e}$ and the values of $\underset{i\mid \widehat{a}_{i, j}^{(g)}=1}{\min}\left\{\dfrac{\bm{y}_i(h)}{d_i}\right\}$ and $\underset{i\mid \widehat{a}_{i, j}^{(g)}=1}{\max}\left\{\dfrac{\bm{y}_i(h)}{d_i}\right\}$, we can determine the upper bound of $\operatorname{U}\left(\bm{R}_j^{(g)}(h)\right)$ when $\bm{R}_j^{(g)}(h)$ lies in  $\left[\underset{i\mid \widehat{a}_{i, j}^{(g)}=1}{\min}\left\{\dfrac{\bm{y}_i(h)}{d_i}\right\},\underset{i\mid \widehat{a}_{i, j}^{(g)}=1}{\max}\left\{\dfrac{\bm{y}_i(h)}{d_i}\right\}\right]$.

Specially, it can be categorized into the following three cases:
{\emph{{\romannumeral1})}} when $\dfrac{\bm{Q}(h)}{e}\ge \underset{i\mid \widehat{a}_{i, j}^{(g)}=1}{\max}\left\{\dfrac{\bm{y}_i(h)}{d_i}\right\}$, we have $\operatorname{U}\left(\bm{R}_j^{(g)}(h)\right)\le \operatorname{U}\left(\underset{i\mid \widehat{a}_{i, j}^{(g)}=1}{\min}\left\{\dfrac{\bm{y}_i(h)}{d_i}\right\}\right)$;
{\emph{{\romannumeral2})}} when $\dfrac{\bm{Q}(h)}{e}\le \underset{i\mid \widehat{a}_{i, j}^{(g)}=1}{\min}\left\{\dfrac{\bm{y}_i(h)}{d_i}\right\}$, we have $\operatorname{U}\left(\bm{R}_j^{(g)}(h)\right)\le \operatorname{U}\left(\underset{i\mid \widehat{a}_{i, j}^{(g)}=1}{\max}\left\{\dfrac{\bm{y}_i(h)}{d_i}\right\}\right)$;
{\emph{{\romannumeral3})}} when $\underset{i\mid \widehat{a}_{i, j}^{(g)}=1}{\min}\left\{\dfrac{\bm{y}_i(h)}{d_i}\right\}\le\dfrac{\bm{Q}(h)}{e}\le \underset{i\mid \widehat{a}_{i, j}^{(g)}=1}{\max}\left\{\dfrac{\bm{y}_i(h)}{d_i}\right\}$, we have $\operatorname{U}\left(\bm{R}_j^{(g)}(h)\right)\le \max{\left\{\operatorname{U}\left(\underset{i\mid \widehat{a}_{i, j}^{(g)}=1}{\min}\left\{\dfrac{\bm{y}_i(h)}{d_i}\right\}\right),\operatorname{U}\left(\underset{i\mid \widehat{a}_{i, j}^{(g)}=1}{\max}\left\{\dfrac{\bm{y}_i(h)}{d_i}\right\}\right) \right\}}$.

Let us define a function	
\begin{equation}
	\bm{G}_{j}^{(g)}(h)=
	\begin{cases}
		\operatorname{U}\left(\underset{i\mid \widehat{a}_{i, j}^{(g)}=1}{\min}\left\{\dfrac{\bm{y}_i(h)}{d_i}\right\}\right)&,\dfrac{\bm{Q}(h)}{e}\ge \underset{i\mid \widehat{a}_{i, j}^{(g)}=1}{\max}\left\{\dfrac{\bm{y}_i(h)}{d_i}\right\} \\ 
		\operatorname{U}\left(\underset{i\mid \widehat{a}_{i, j}^{(g)}=1}{\max}\left\{\dfrac{\bm{y}_i(h)}{d_i}\right\}\right)&,\dfrac{\bm{Q}(h)}{e}\le \underset{i\mid \widehat{a}_{i, j}^{(g)}=1}{\min}\left\{\dfrac{\bm{y}_i(h)}{d_i}\right\} \\
		\max{\left\{\operatorname{U}\left(\underset{i\mid \widehat{a}_{i, j}^{(g)}=1}{\min}\left\{\dfrac{\bm{y}_i(h)}{d_i}\right\}\right),\operatorname{U}\left(\underset{i\mid \widehat{a}_{i, j}^{(g)}=1}{\max}\left\{\dfrac{\bm{y}_i(h)}{d_i}\right\}\right) \right\}}&,{\text{otherwize.}}
	\end{cases}
\end{equation}
as the upper bound of $\operatorname{U}\left(\bm{R}_j^{(g)}(h)\right)$. We  can then obtain the upper bound of $\operatorname{Pr}\left(\operatorname{KLD}\left(\widehat{\boldsymbol{P}}_j^{(g)}||\boldsymbol{Q}\right)>{\it{KLD}}_{\textrm{max}}-\Delta_k\mid\sum_{i=1}^{|\mathcal{N}|}\widehat{a}_{i, j}^{(g)}\xi_{i}^{(g)}\neq 0\right)$ according to the Markov inequality\footnote{It is worth noting that the above derivation relies on the use of Markov’s inequality to upper-bound the violation probability. Although such a bound may be conservative in practice, it provides a tractable closed-form surrogate that guarantees the feasibility of the probabilistic constraint in \eqref{p1_11a}. This theoretical guarantee motivates the adoption of the Markov and union bounds in our formulation, even at the cost of potential looseness in the resulting constraints.} as follows:
\begin{align}
	&\operatorname{Pr}\left(\operatorname{KLD}\left(\widehat{\boldsymbol{P}}_j^{(g)}||\boldsymbol{Q}\right)>{\it{KLD}}_{\textrm{max}}	-\Delta_k \Big|\sum_{i=1}^{|\mathcal{N}|}\widehat{a}_{i, j}^{(g)}\xi_{i}^{(g)}\neq0\right)
	\le\dfrac{\mathbf{E}\left[\sum_{h=1}^{|\mathcal{Z}|}\operatorname{U}\left(\bm{R}_j^{(g)}(h)\right)\Big| \sum_{i=1}^{|\mathcal{N}|}\widehat{a}_{i, j}^{(g)}\xi_{i}^{(g)}\neq0\right]}{{\it{KLD}}_{\textrm{max}}	-\Delta_k}\notag\\
	&=\sum_{h=1}^{|\mathcal{Z}|}\dfrac{\mathbf{E}\left[\operatorname{U}\left(\bm{R}_j^{(g)}(h)\right)\Big| \sum_{i=1}^{|\mathcal{N}|}\widehat{a}_{i, j}^{(g)}\xi_{i}^{(g)}\neq0\right]}{{\it{KLD}}_{\textrm{max}}	-\Delta_k}\le \sum_{h=1}^{|\mathcal{Z}|}\dfrac{\bm{G}_{j}^{(g)}(h)}{{\it{KLD}}_{\textrm{max}}-\Delta_k},
\end{align}
where $\mathbf{E}\left[\cdot \right]$ is the expectation operator. Finally,  we can get the upper bound of $\operatorname{Pr}\left(\operatorname{KLD}\left(\widehat{\boldsymbol{P}}_j^{(g)}||\boldsymbol{Q}\right)>{\it{KLD}}_{\textrm{max}}	-\Delta_k\right)$ as follows: 
\begin{align}
	&\operatorname{Pr}\left(\operatorname{KLD}\left(\widehat{\boldsymbol{P}}_j^{(g)}||\boldsymbol{Q}\right)>{\it{KLD}}_{\textrm{max}}	-\Delta_k\right)=\!\prod_{i\mid \widehat{a}_{i, j}^{(g)}=1}{\left(1-p_i\right)}+\left(1-\!\prod_{i\mid \widehat{a}_{i, j}^{(g)}=1}\!{\left(1-p_i\right)}\right)\notag\\
	&\times\!\operatorname{Pr}\!\left(\operatorname{KLD}\left(\widehat{\boldsymbol{P}}_j^{(g)}||\boldsymbol{Q}\right)\!>\!{\it{KLD}}_{\textrm{max}}\!-\!\Delta_k\Big| \sum_{i=1}^{|\mathcal{N}|}\widehat{a}_{i, j}^{(g)}\xi_{i}^{(g)}\neq 0\right)\!
	\le \!\!\prod_{i\mid \widehat{a}_{i, j}^{(g)}=1}\!\!{\left(1-\!p_i\right)}\!\left(1-\!\!\prod_{i\mid \widehat{a}_{i, j}^{(g)}=1}\!\!{\left(1-\!p_i\right)}\!\right)\!\times\!\sum_{h=1}^{|\mathcal{Z}|}\!\dfrac{\bm{G}_{j}^{(g)}(h)}{{\it{KLD}}_{\textrm{max}}\!-\!\Delta_k}.\!\!\!
\end{align}	

Therefore, the constraint \eqref{p1_11a} is tightened to \eqref{p1_11f} in Sec. \ref{subsubsec:plan a}, which is
\begin{equation*}
	\prod_{i \mid \widehat{a}_{i, j}^{(g)}=1}\left(1-p_i\right)+\left(1-\prod_{i \mid \widehat{a}_{i, j}^{(g)}=1} \left(1-p_i\right)\right)\times\sum_{h=1}^{|\bm{\mathcal{Z}}|}\frac{\bm{G}_{j}^{(g)}(h)}{{\it{KLD}}_{\textrm{max}}	-\Delta_k}\le \delta, \forall s_{j}\in\mathcal{S}.
\end{equation*}
\section{Derivation of \eqref{p1_11g}}
\label{appx:derivation of 11g}
By substituting the definition of $D_j^{(g)}$ into the above \eqref{p1_11b}, we can obtain that
\begin{equation}
	\operatorname{Pr}\left(\sum_{i=1}^{|\mathcal{N}|}\widehat{a}_{i, j}^{(g)}\xi_i^{(g)} d_i<D_{\textrm {min}}+\Delta_d\right)\le \varepsilon, \forall s_{j}\in\mathcal{S}.
\end{equation}

Base on the Markov inequality, we can obtain the lower bound of the left side of the above inequality as follows:
\begin{equation}
	\operatorname{Pr}\left(\sum_{i=1}^{|\mathcal{N}|}\widehat{a}_{i, j}^{(g)}\xi_i^{(g)} d_i<D_{\textrm {min}}+\Delta_d\right)\ge 1-\frac{\mathbf{E}\left[\sum_{i=1}^{|\mathcal{N}|}\widehat{a}_{i, j}^{(g)}\xi_i^{(g)} d_i\right]}{D_{\textrm {min}}+\Delta_d}, \forall s_{j}\in\mathcal{S}.
\end{equation}

Further, according to the linearity of expectation, we can get
\begin{equation}
	\mathbf{E}\left(\sum_{i=1}^{|\mathcal{N}|}\widehat{a}_{i, j}^{(g)}\xi_i^{(g)} d_i\right)=\sum_{i=1}^{|\mathcal{N}|}\widehat{a}_{i, j}^{(g)}\mathbf{E}\left(\xi_i^{(g)}\right)d_i=\sum_{i=1}^{|\mathcal{N}|}\widehat{a}_{i, j}^{(g)}p_i d_i, \forall s_{j}\in\mathcal{S}.
\end{equation}
We can thus relax the constraint \eqref{p1_11b} to 
\begin{equation}
	1-\frac{\sum_{i=1}^{|\mathcal{N}|}\widehat{a}_{i, j}^{(g)}p_i d_i}{D_{\textrm {min}}+\Delta_d}\le \varepsilon, \forall s_{j}\in\mathcal{S},
\end{equation}
which results in \eqref{p1_11g}:
\begin{equation}
	\sum_{i=1}^{|\mathcal{N}|}\widehat{a}_{i, j}^{(g)}p_{i}d_{i}\ge\left(D_{\textrm {min}}+\Delta_d\right)\left(1-\varepsilon\right), \forall s_{j}\in\mathcal{S}\notag.
\end{equation}
\vspace{-5mm}
\section{Deatails of Operations in Alg. 3}
\label{appx:operations}
\noindent $\bullet$ \emph{\textbf{Operation 1. Addition of clients (Add)}}: 
We test a new client from the not-selected clients set (i.e., $\mathcal{N} \textbackslash{} \mathcal{N}^s$) by adding it into the selected client set and run Alg. \ref{algorithm2} to calculate $\widehat{F}^{(g)}$, denoted by $\widehat{F}^{(g)}_{\cup \left\{\bm{n}_i\right\}}$. If this operation satisfies all constraints and reduces  $\widehat{F}^{(g)}$, we update $\widehat{F}^{(g)}_{\textrm {min}}$, the corresponding selected client set $\mathcal{N}^s$ and association matrix $\widehat{\bm{A}}^{(g)}$. This  continues until we evaluate all the not-selected clients, identifying and incorporating the additions that contribute to the reduction of the cost function. If no such additional improvement is found, we retain the original set of clients (lines 4-11). 

\noindent $\bullet$ \emph{\textbf{Operation 2. Removal of clients (Remove)}}: We remove a client (i.e., move it from the long-term client set to the not-selected set) and run Alg. \ref{algorithm2} to calculate $\widehat{F}^{(g)}$, denoted by $\widehat{F}^{(g)}_{\textbackslash{} \left\{\bm{n}_i\right\}}$. If this removal satisfies all constraints and reduces $\widehat{F}^{(g)}$, we update $\widehat{F}^{(g)}_{\textrm {min}}$, $\mathcal{N}^s$ and $\widehat{\bm{A}}^{(g)}$. This continues until  checking all selected clients, seeking the removals that offer reduction in the cost function. If no such removal are found, we preserve the initial selection of clients (lines 12-19). 

\noindent $\bullet$ \emph{\textbf{Operation 3. Exchange of clients (Exchange)}}: We execute a swap by removing a currently selected client and adding another (i.e., exchanging a selected client with a not-selected one). Following this exchange, we run Alg. \ref{algorithm2} to calculate $\widehat{F}^{(g)}$, denoted by $\widehat{F}^{(g)}_{{\textbackslash{} \left\{\bm{n}_i\right\}}\cup \left\{\bm{n}_j\right\}}$. If this exchange satisfies all  constraints and decreases $\widehat{F}^{(g)}$, we update $\widehat{F}^{(g)}_{\textrm {min}}$, $\mathcal{N}^s$ and $\widehat{\bm{A}}^{(g)}$. This is carried out until examining all clients, identifying the exchanges that reduce $\widehat{F}^{(g)}$. If no such exchange is found, we retain the original set of clients (lines 20-28).

\section{Detailed Information of Experimental Testbed}
\label{appx:testbed}
All simulations were conducted on a workstation equipped with an Intel Core i9-13900K @ 3.00 GHz CPU, 32 GB memory, and an NVIDIA GeForce RTX 4080 GPU, using MATLAB R2021a and Python 3.9 with PyTorch 1.13. The decision-making modules (Plan A and Plan B) were implemented in MATLAB as single-threaded processes on the CPU, without employing GPU acceleration, parallel computing toolboxes, or commercial solvers. The model training and federated learning procedure were simulated on the above testbed using PyTorch. We note that such single-machine simulation of distributed training is a widely adopted and validated practice in both FL and HFL research \cite{ref62, ref63, ref64}, as it provides a fair and reproducible environment for comparing algorithmic performance without introducing confounding factors from heterogeneous hardware or networking infrastructures. Therefore, the adopted testbed setup is both consistent with existing literature and sufficient to demonstrate the effectiveness of the proposed framework.

\section{Datasets and Models}
\label{appx:dataset and model}
To ensure reproducibility and fair comparison with prior works, we adopt four widely used benchmark datasets, each coupled with a representative deep learning model. The details are summarized as follows:

\noindent $\bullet$ \textbf{MNIST dataset} \cite{ref39}: A standard handwritten digit recognition dataset consisting of 60,000 training and 10,000 testing grayscale images of size $28 \times 28$. We employ a standard PyTorch CNN model with 21,840 parameters.
    
\noindent $\bullet$ \textbf{Fashion-MNIST dataset} \cite{ref41}: A dataset containing 60,000 training and 10,000 testing grayscale images of 10 fashion product categories. We adopt the classical LeNet-5 model \cite{ref39}, which contains 29,034 parameters.
    
\noindent $\bullet$ \textbf{CIFAR-10 dataset} \cite{ref40}: A color image dataset with 50,000 training and 10,000 testing samples across 10 object categories. We use a PyTorch CNN model with 576,778 parameters.
    
\noindent $\bullet$ \textbf{CIFAR-100 dataset} \cite{ref40}: A more challenging extension of CIFAR-10, containing 100 classes with 600 images each. For this dataset, we employ the ResNet-18 model \cite{ref42}, which has 11,220,196 parameters.

\noindent $\bullet$ \textbf{Aerial Landscape Images (ALI) dataset} \cite{ref65, ref66}: A dataset constructed from the publicly available AID and NWPU-Resisc45 datasets, containing 15 aerial scene categories (e.g., rivers, mountains, forests, farmland, etc.) under diverse aerial conditions. To evaluate model performance, we employ a lightweight VGG network \cite{ref67} with 60,074 parameters.

These datasets cover a broad spectrum of task complexities, from simple grayscale digits to high-resolution colored images, while the selected models range from lightweight CNNs to deeper architectures (e.g., ResNet18). This diverse setup provides a comprehensive evaluation of the proposed framework under different learning scenarios.

\section{Extended Evaluations I}
\label{appx:evaluation I}
\begin{figure}[htbp]

	\centering
	\includegraphics[trim=1.5cm 0.5cm 3.5cm 0.7cm, clip, width=0.47\columnwidth]{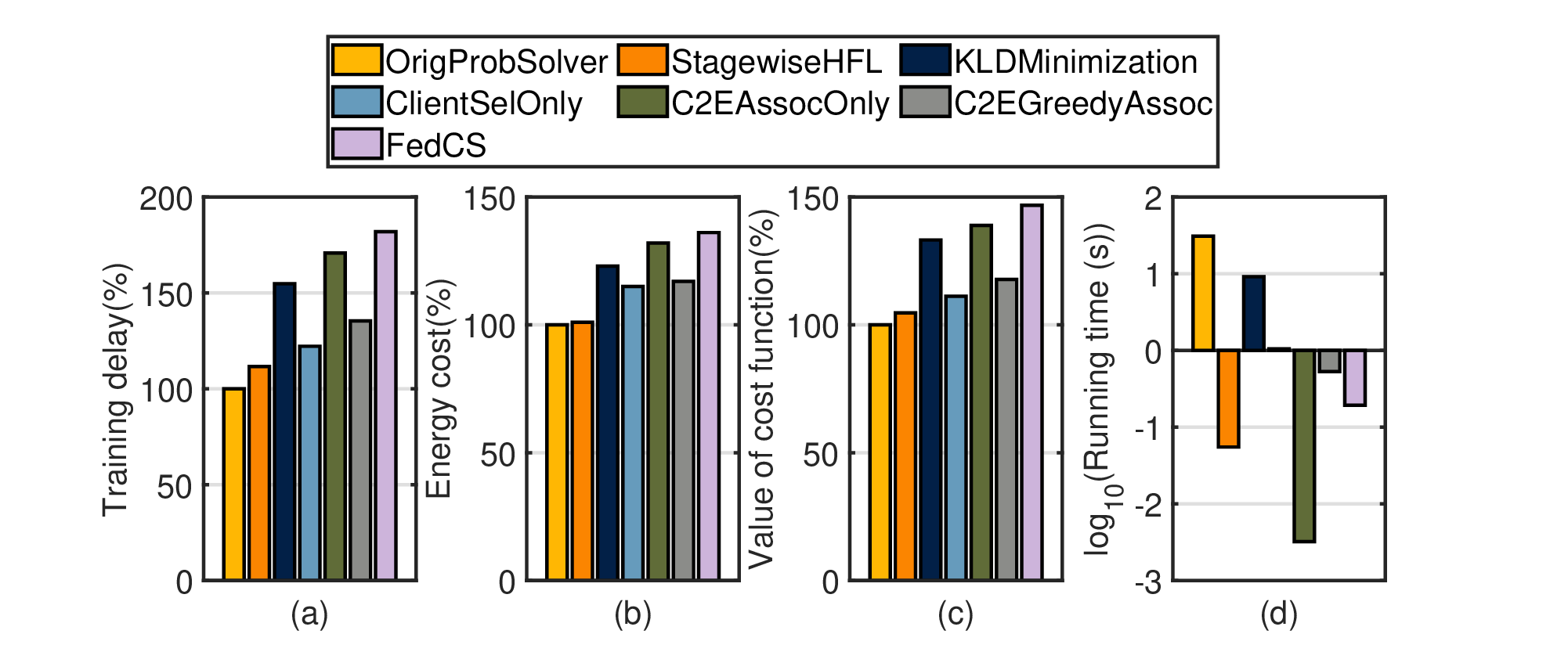}
	\caption{Performance on training delay, energy consumption, overall cost, and running time under varying channel conditions and clients’ computational capabilities.}
	\label{fig:performance with multiple plan a}

\end{figure}

To better capture realistic and dynamic scenarios, we set the channel gain between clients and ESs to fluctuate within $[10^{-9}, 10^{-8}]$, and client CPU frequencies to vary within $[1, 10]$ GHz in each global iteration (unlike the fixed settings in Sec. \ref{sec:evaluations}). Under such dynamics, multiple executions of Plan A are necessary to accurately estimate clients' online probabilities. In practice, we set Plan A to run once every 10 global rounds, thereby adapting to environmental variations. As shown in Fig. \ref{fig:performance with multiple plan a}, StagewiseHFL achieves over 90\% test accuracy on MNIST while incurring only modest increases of 13.2\%, 3.0\%, and 6.0\% in training delay, energy consumption, and overall cost, respectively, compared with OrigProbSolver, while still outperforming other baselines. For decision-making overhead, StagewiseHFL requires an average of only 51ms, demonstrating strong effectiveness and time efficiency even in highly dynamic environments. This efficiency is largely due to the fact that additional executions of Plan A can be performed in parallel with the training process, ensuring that repeated runs do not introduce noticeable real-time decision-making delay.
Moreover, the frequency of triggering Plan A can be flexibly adjusted to balance decision accuracy and runtime efficiency. For example, Plan A may be reactivated when: \textit{i)} unacceptable system costs (e.g., excessive training delay or energy consumption) are observed, \textit{ii)} channel conditions or clients' computing capabilities change significantly, or \textit{iii)} a predefined number of global rounds has been completed.
Although a more comprehensive exploration of these adaptive strategies is left for future work, the above results confirm that parallel execution of Plan A enables StagewiseHFL to remain scalable and practical for real-world deployments.

\section{Extended Evaluations II}
\label{appx:evaluation II}
\begin{figure}[H]
    
	\centering
	
	\subfigure[Cost vs. $D_{\textrm {min}}$]{
    \label{subfig:cost vs dmin for approximation}
		\includegraphics[trim=0cm 0cm 1.0cm 0.5cm, clip, width=0.47\columnwidth]{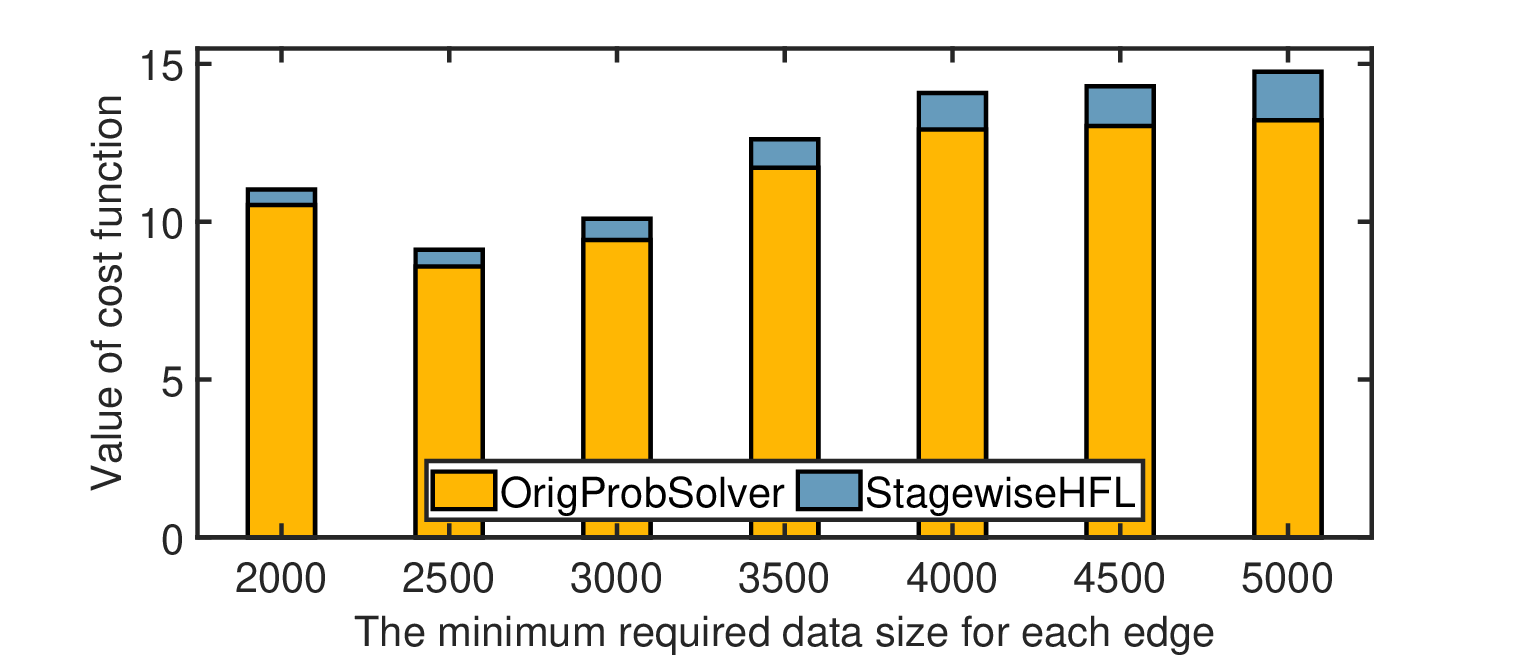}
	}
	\hspace{-10pt}
	\subfigure[Cost vs. ${\it{KLD}}_{\textrm{max}}$]{
    \label{subfig:cost vs kldmax for approximation}
		\includegraphics[trim=0cm 0cm 1.0cm 0.5cm, clip, width=0.47\columnwidth]{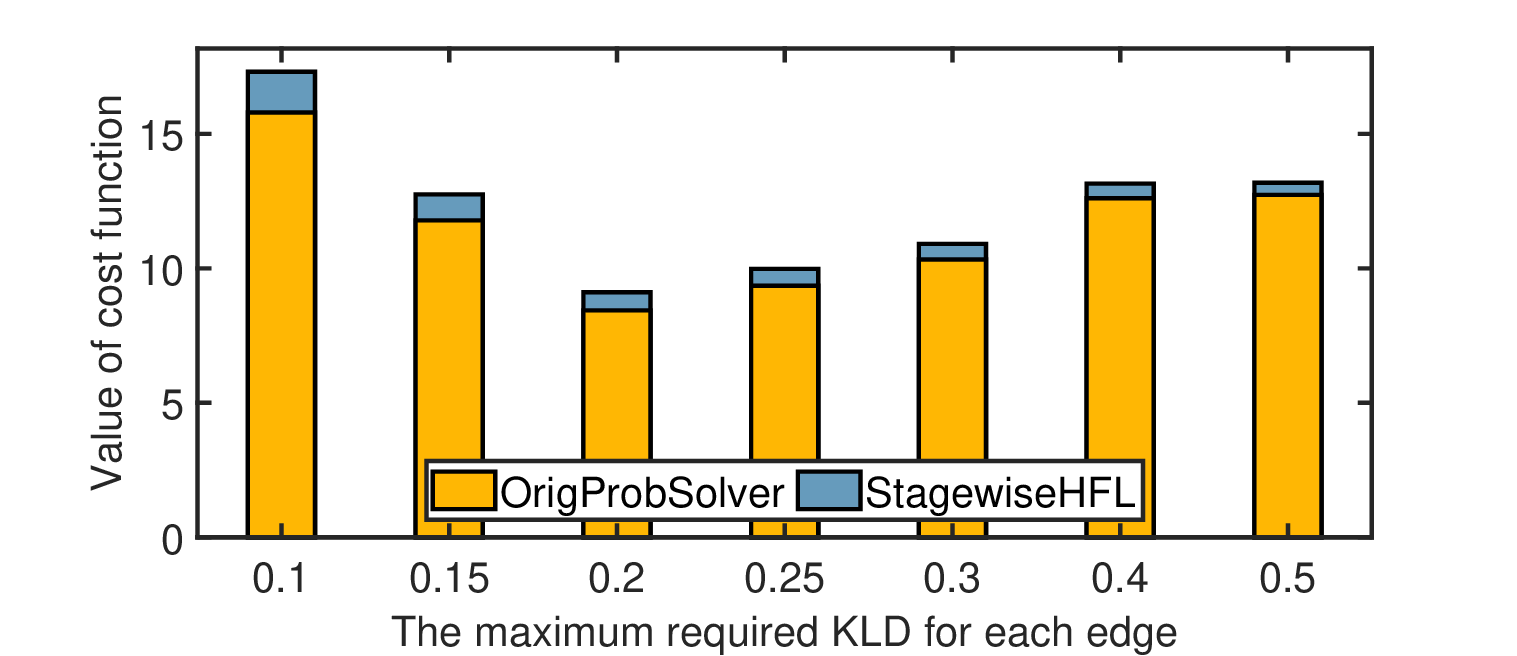}
	}
	\caption{Performance on the overall system cost for OrigProbSolver and StagewiseHFL.}
    \label{fig:performance approximation}
\end{figure}

To verify the reliability of the designed constraints \eqref{p1_11a} and \eqref{p1_11b} in Plan A, we conducted experiments under various values of $D_{\textrm{min}}$ and ${\it{KLD}}_{\textrm{max}}$, evaluating both the cost-saving performance of StagewiseHFL and its approximation to OrigProbSolver. As shown in Fig. \ref{fig:performance approximation}, in most cases, we only have slight gaps in terms of value of cost function, in comparison with OrigProbSolver.  However, when Plan A selects a large number of long-term clients, the performance of StagewiseHFL may degrade. For instance, when $D_{\textrm{min}} = 5000$ and ${\it{KLD}}_{\textrm{max}} = 0.1$, the overall cost of StagewiseHFL is 11.6\% higher in terms of time and 9.6\% higher in terms of energy compared with OrigProbSolver. This degradation is mainly due to the reduced diversity of training data, which requires more global rounds to reach the target accuracy, thereby increasing both time and energy consumption. Nevertheless, with appropriate choices of $D_{\textrm{min}}$ and ${\it{KLD}}_{\textrm{max}}$, StagewiseHFL maintains reliable performance in practice. Moreover, the observed cost gaps remain acceptable, particularly considering that OrigProbSolver involves heavy decision-making overhead (see Fig. \ref{fig:performace on running time}), making it impractical for dynamic real-world networks. From another view, these small gaps can verify the approximation of constraints \eqref{p1_11a}) and \eqref{p1_11b} against their originals, \eqref{p0_9a}) and \eqref{p0_9b}, respectively.
	
It is worth noting that although the use of Markov inequality and the union bound introduces a certain degree of constraint relaxation during the transformation, the results still show that StagewiseHFL achieves substantial cost reduction and maintains robustness across different settings, and further substantiates the theoretical soundness and practical efficacy of our proposed constraints \eqref{p1_11f} and \eqref{p1_11g}.

\section{Extended Evaluations III}
\label{appx:evaluation III}
To further validate the practical effectiveness of our proposed constraints \eqref{p1_11f} and \eqref{p1_11g}, we conducted extensive Monte Carlo simulations to quantitatively demonstrate the relationship between the theoretical thresholds ($\delta$, $\epsilon$) and their corresponding empirical violation rates ($\widehat{\delta}$,  $\widehat{\epsilon}$), which are summarized in Tables \ref{tab:delta hat vs delta} and \ref{tab:epsilon hat vs epsilon} below.

\begin{table}[h]
	\begin{minipage}{0.48\textwidth}
	\small
	\centering
	\caption{	
		Empirical violation rate $\widehat{\delta}$ vs. theoretical threshold $\delta$}
        \label{tab:delta hat vs delta}
	\setlength{\tabcolsep}{7.5pt}
	\renewcommand{\arraystretch}{1.2} 
	\begin{tabular}{|c|c|@{\hskip 3pt}|c|c|c|c|c|}
		\hline
		\rowcolor{verylightgray}
		\multicolumn{2}{|c|@{\hskip 3pt}|}{\multirow{2}{*}{$\widehat{\delta}$}} & \multicolumn{5}{c|}{$\it{KLD}_\textrm{max}$} \\ \cline{3-7}  
		\multicolumn{2}{|c|@{\hskip 3pt}|}{~} & 0.1   & 0.2   & 0.3   & 0.4   & 0.5 \\ \hline
		\hline
		\rowcolor{verylightgray} & 0.15 & 0.31 & 0.29 & 0.26 & 0.24 & 0.23 \\ 
		\cline{2-7}
		& 0.20 & 0.36 & 0.32 & 0.31 & 0.32 & 0.29 \\
		\hhline{|~|------|}
		\rowcolor{verylightgray} & 0.25 & 0.37 & 0.34 & 0.31 & 0.32 & 0.28 \\ 
		\hhline{|~|------|}
		& 0.30 & 0.38 & 0.33 & 0.30 & 0.28 & 0.29 \\
		\hhline{|~|------|}
		\rowcolor{verylightgray} \multirow{-5}{*}{~~$\delta$~~} & 0.35 & 0.41 & 0.37 & 0.35 & 0.34 & 0.33 \\ \hline
	\end{tabular}
	\end{minipage}
	\hfill
	\begin{minipage}{0.48\textwidth}
	\small
	\centering
	\caption{	
		Empirical violation rate $\widehat{\epsilon}$ vs. theoretical threshold $\epsilon$}
        \label{tab:epsilon hat vs epsilon}
	\setlength{\tabcolsep}{7pt}
	\renewcommand{\arraystretch}{1.2} 
	\begin{tabular}{|c|c|@{\hskip 3pt}|c|c|c|c|c|}
		\hline
		\rowcolor{verylightgray}
		\multicolumn{2}{|c|@{\hskip 3pt}|}{\multirow{2}{*}{$\widehat{\epsilon}$}} & \multicolumn{5}{c|}{$D_\textrm{min}$} \\ \cline{3-7}  
		\multicolumn{2}{|c|@{\hskip 3pt}|}{~} & 2000   & 2500   & 3000   & 4000   & 5000 \\ \hline
		\hline
		\rowcolor{verylightgray} & 0.15 & 0.27 & 0.26 & 0.22 & 0.23 & 0.21 \\ 
		\cline{2-7}
		& 0.20 & 0.32 & 0.30 & 0.26 & 0.24 & 0.23 \\
		\hhline{|~|------|}
		\rowcolor{verylightgray} & 0.25 & 0.33 & 0.31 & 0.28 & 0.26 & 0.25 \\ 
		\hhline{|~|------|}
		& 0.30 & 0.35 & 0.32 & 0.27 & 0.28 & 0.26 \\
		\hhline{|~|------|}
		\rowcolor{verylightgray} \multirow{-5}{*}{~~$\epsilon$~~} & 0.35 & 0.38 & 0.34 & 0.33 & 0.29 & 0.27 \\ \hline
	\end{tabular}
\end{minipage}
\end{table}

As shown in Tables \ref{tab:delta hat vs delta} and \ref{tab:epsilon hat vs epsilon}, the empirical violation rates $\widehat{\delta}$ and $\widehat{\epsilon}$ are generally close to their theoretical thresholds $\delta$ and $\epsilon$, respectively. Although in certain configurations the empirical values are noticeably higher—for example, $\widehat{\delta}=0.31$ when ${\it{KLD}}_\textrm{max}=0.1$ and $\delta=0.15$—the deviations remain bounded and fall within an acceptable range for practical applications. This is consistent with the conservative nature of the Markov and union bounds, and is further influenced by the introduction of two constants $\Delta_{k}>0$ and $\Delta_{d}>0$, which were designed to tune the satisfaction probability of the original constraints \eqref{p0_9a} and \eqref{p0_9b} and to mitigate the risk of violating KLD and data size thresholds during practical training. The results confirm that our method offers a balanced and robust trade-off between theoretical rigor and practical feasibility.

\section{Extended Evaluations IV}
\label{appx:evaluation IV}
To further validate the effect of ${\it{KLD}}_{\textrm{max}}$ on model accuracy, , building upon the observations in Sec. \ref{subsubsec:analysis on T and L} we extend our experiments to two more challenging datasets: CIFAR-100 (trained with ResNet-18) and Tiny-ImageNet (trained with ResNet-34), where Tiny-ImageNet is a downsampled variant of the ImageNet dataset, consisting of 200 classes, each with 500 training images and 50 validation images. All images are resized to 64×64 pixels. Compared with CIFAR-100, Tiny-ImageNet is more complex both in terms of class granularity and visual diversity. In the HFL setup with 100 clients, CIFAR-100 is partitioned such that each client owns 10–30 classes, while in Tiny-ImageNet each client is assigned 20–60 classes. We compare the performance of OrigProbSolver and StagewiseHFL under different values of ${\it{KLD}}_{\textrm{max}}$. For each setting, we report the final test accuracy within 100 global iteration. 

As shown in Tables \ref{tab:acc vs kldmax on cifar100} and \ref{tab:acc vs kldmax on tiny-imagenet}, imposing more restrictive ${\it{KLD}}_{\textrm{max}}$ values (i.e., enforcing lower data heterogeneity across ESs) leads to higher test accuracy for both methods. For example, on CIFAR-100, reducing ${\it{KLD}}{\textrm{max}}$ from 1.0 to 0.3 improves OrigProbSolver's accuracy from 35.34\% to 55.12\%, indicating that stricter distributional constraints effectively enhance model performance by mitigating data heterogeneity. While OrigProbSolver consistently achieves slightly higher accuracy than StagewiseHFL, the latter remains highly competitive, with only a marginal gap, and offers substantially better scalability. These results reaffirm the trend observed on MNIST, demonstrating that controlling KLD positively impacts model accuracy and that this effect extends to larger and more complex datasets.
\begin{table}[h]
	\begin{minipage}{0.48\textwidth}
		\small
		\centering
		\caption{	
			Test accuracy vs. $\it{KLD}_\textrm{max}$ on CIFAR-100 }
            \label{tab:acc vs kldmax on cifar100} 
		\setlength{\tabcolsep}{3pt}
		\renewcommand{\arraystretch}{1.2} 
		\begin{tabular}{|c|c|@{\hskip 3pt}|c|c|c|c|c|}
			\hline
			\rowcolor{verylightgray}
			\multicolumn{2}{|c|@{\hskip 3pt}|}{\multirow{2}{*}{Test accuracy (\%)}} & \multicolumn{5}{c|}{$\it{KLD}_\textrm{max}$} \\ \cline{3-7}  
			\multicolumn{2}{|c|@{\hskip 3pt}|}{~} &0.3 & 0.4 & 0.5 & 0.7 & 1.0 \\ \hline
			\hline
			\rowcolor{verylightgray} & OrigProbSolver & 55.12 & 47.77 & 43.27 & 36.49 & 35.34 \\ 
			\cline{2-7}
			\multirow{-2}{*}{Method}& StagewiseHFL & 54.37 & 46.43 & 43.66 & 36.38 & 35.41 \\
			\hline
		\end{tabular}
	\end{minipage}
	\hfill
	\begin{minipage}{0.48\textwidth}
		\small
		\centering
		\caption{	
			Test accuracy vs. $\it{KLD}_\textrm{max}$ on Tiny-ImageNet }
            \label{tab:acc vs kldmax on tiny-imagenet}
		\setlength{\tabcolsep}{3pt}
		\renewcommand{\arraystretch}{1.2} 
		\begin{tabular}{|c|c|@{\hskip 3pt}|c|c|c|c|c|}
			\hline
			\rowcolor{verylightgray}
			\multicolumn{2}{|c|@{\hskip 3pt}|}{\multirow{2}{*}{Test accuracy (\%)}} & \multicolumn{5}{c|}{$\it{KLD}_\textrm{max}$} \\ \cline{3-7}  
			\multicolumn{2}{|c|@{\hskip 3pt}|}{~} &0.3 & 0.5 & 0.7 & 1.0 & 1.5 \\ \hline
			\hline
			\rowcolor{verylightgray} & OrigProbSolver & 45.21 & 41.42 & 36.26 & 32.49 & 31.34 \\ 
			\cline{2-7}
			\multirow{-2}{*}{Method}& StagewiseHFL & 43.64 & 39.94 & 35.79 & 32.38 & 31.41 \\
			\hline
		\end{tabular}
	\end{minipage}
\end{table}
\section{Extended Evaluations V}
\label{appx:evaluation V}
\begin{figure}[H]

	\centering
	
	\subfigure[Agriculture]{
    \label{subfig:agriculture}
		\includegraphics[trim=0cm 0cm 0cm 0cm, clip, width=0.15\columnwidth]{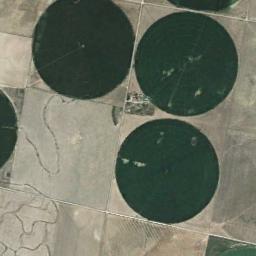}
	}
	\hspace{-2pt}
	\subfigure[Airport]{
    \label{subfig:airport}
		\includegraphics[trim=0cm 0cm 0cm 0cm, clip, width=0.15\columnwidth]{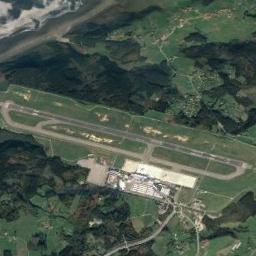}
	}
	\hspace{-2pt}
	\subfigure[Beach]{
    \label{subfig:beach}
		\includegraphics[trim=0cm 0cm 0cm 0cm, clip, width=0.15\columnwidth]{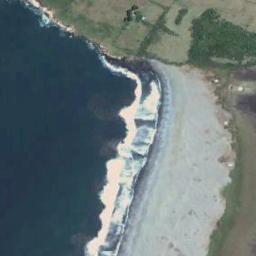}
	}
	\hspace{-2pt}
	\subfigure[City]{
    \label{subfig:city}
	\includegraphics[trim=0cm 0cm 0cm 0cm, clip, width=0.15\columnwidth]{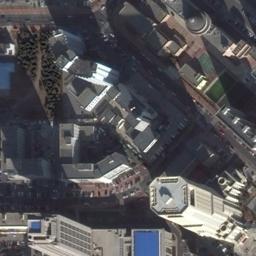}
	}
	\hspace{-2pt}
	\subfigure[Desert]{
    \label{subfig:desert}
	\includegraphics[trim=0cm 0cm 0cm 0cm, clip, width=0.15\columnwidth]{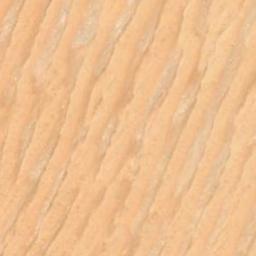}
	}
	\hspace{-2pt}
	\subfigure[Forest]{
    \label{subfig:forest}
	\includegraphics[trim=0cm 0cm 0cm 0cm, clip, width=0.15\columnwidth]{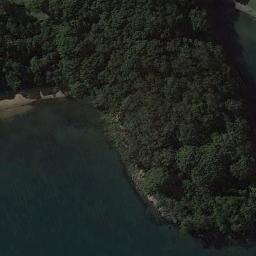}
	}
	\caption{Different categories of images in ALI}
    \label{fig:ali}
\end{figure}

To further bridge the gap between simulation-based experiments and real-world deployment, we conduct supplementary evaluations on the ALI dataset, from which we select 10 representative categories that closely emulate realistic UAV deployment scenarios (see Fig. \ref{fig:ali}).

\begin{figure}[H]
	\centering
	\begin{minipage}[t]{0.37\columnwidth}
    \label{fig:acc on ali}
		\centering
		\includegraphics[trim=0cm 0cm 1.0cm 0.5cm, clip, width=\textwidth]{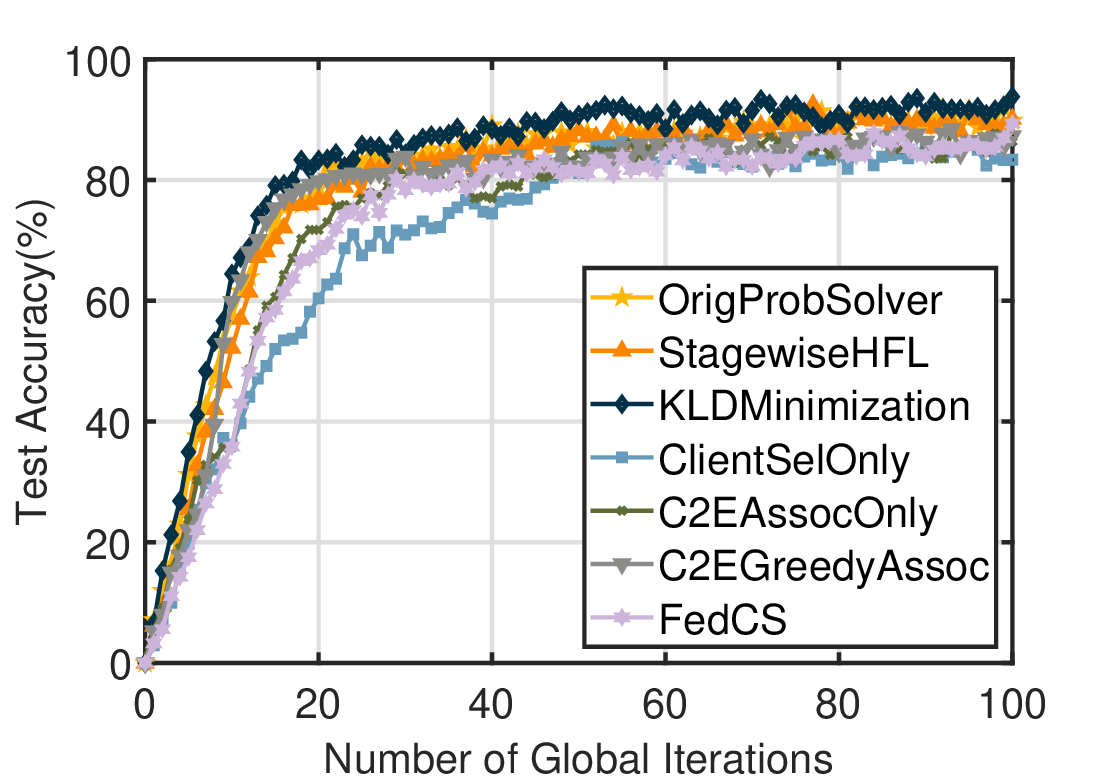}
		\caption{Performance comparison in terms of test accuracy on ALI}
		\label{fig:ali_acc}
	\end{minipage}
	\hfill
	\begin{minipage}[t]{0.57\columnwidth}
    \label{fig:performance on ali}
		\centering
		\includegraphics[trim=1.5cm 0.5cm 3.0cm 0.5cm, clip, width=\textwidth]{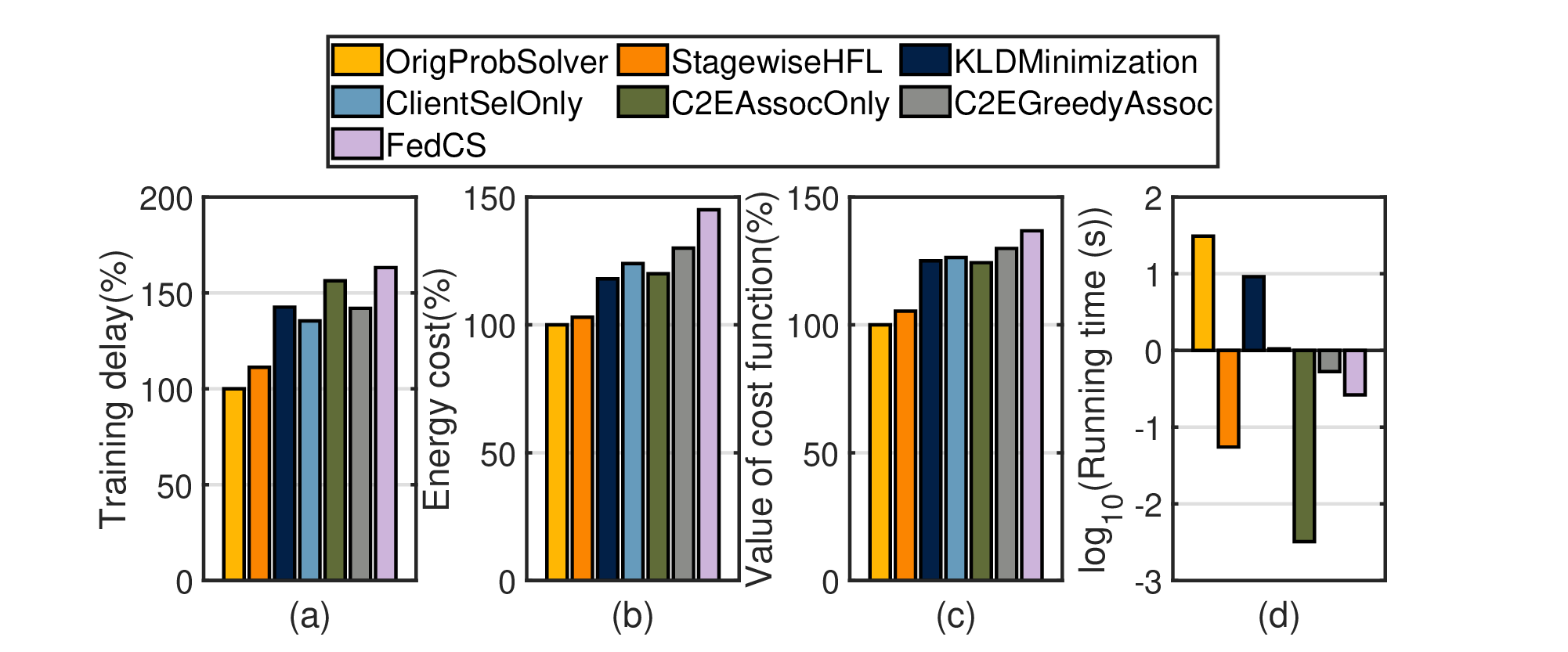}
		\caption{Performance on training delay, energy consumption, overall cost, and running time on ALI}
		\label{fig:ali_performance}
	\end{minipage}
\end{figure}

As illustrated in Fig. \ref{fig:ali_acc}, after 100 global iterations on ALI, StagewiseHFL achieves higher final test accuracy than ClientSelOnly, C2EAssocOnly, C2EGreedyAssoc, and FedCS by 2.68\%, 2.26\%, 1.48\%, and 3.29\%, respectively, while remaining only 1.12\% and 1.59\% lower than OrigProbSolver and KLDMinimization. Regarding cost efficiency on the ALI dataset (Fig. \ref{fig:ali_performance}(c)), to reach over 85\% test accuracy, StagewiseHFL incurs merely a 5.3\% loss compared to OrigProbSolver, yet outperforms KLDMinimization, ClientSelOnly, C2EAssocOnly, C2EGreedyAssoc, and FedCS by 19.6\%, 21.0\%, 18.9\%, 24.6\%, and 31.5\%, respectively. Importantly, our method achieves an average inference time of less than 100 ms (Fig. \ref{fig:ali_performance}(d)), which is an order of magnitude faster than OrigProbSolver, highlighting its suitability for real-time deployment. Overall, these results consistently demonstrate that StagewiseHFL delivers superior performance in model accuracy, cost efficiency, and decision-making speed, even under large-scale and more challenging scenarios.

\end{document}